%% file: paper.tex
\documentclass[11pt]{uiuc_paper}
\pdfoutput=1
\usepackage[toc,page,header]{appendix}
\usepackage{cleveref}
\usepackage{xurl} %
\crefname{section}{Sec.}{Secs.}\Crefname{section}{Sec.}{Secs.}
\crefname{subsection}{Sec.}{Secs.}\Crefname{subsection}{Sec.}{Secs.}
\crefname{subsubsection}{Sec.}{Secs.}\Crefname{subsubsection}{Sec.}{Secs.}
\crefname{appendix}{App.}{Apps.}\Crefname{appendix}{App.}{Apps.}
\crefname{subappendix}{App.}{Apps.}\Crefname{subappendix}{App.}{Apps.}
\crefname{subsubappendix}{App.}{Apps.}\Crefname{subsubappendix}{App.}{Apps.}
\crefname{figure}{Fig.}{Figs.}\Crefname{figure}{Fig.}{Figs.}
\crefname{table}{Tab.}{Tabs.}\Crefname{table}{Tab.}{Tabs.}
\crefname{equation}{Eq.}{Eqs.}\Crefname{equation}{Eq.}{Eqs.}
\crefname{algorithm}{Alg.}{Algs.}\Crefname{algorithm}{Alg.}{Algs.}
\usepackage{subcaption}
\usepackage{booktabs}
\usepackage{tabularx}

\input{macro}

\hypersetup{pdftitle={Language Models Can Rewrite Their Own Weights},pdfauthor={Jiajun Fan, Neil He, Changyi Yang, Andy Mao, Xi Zhu, Chaoran Cheng, Kaiqu Liang, Shuaike Shen, Zetian Li, Ruihan Guo, Ge Liu}}
\usepackage{algorithm}
\usepackage[noend]{algpseudocode}
\usepackage[most]{tcolorbox}
\usepackage{multirow}
\definecolor{srwOursTxt}{HTML}{3D6A5F}   %
\usepackage{pifont}
\usepackage{CJKutf8}
\usepackage{colortbl}
\usepackage{makecell}
\usepackage{flafter} %
\definecolor{srwBandTeal}{HTML}{E6F2EE}  %
\definecolor{srwFailEdge}{HTML}{B68A78}  %
\definecolor{srwFailTxt}{HTML}{8A5640}   %
\definecolor{srwMuted}{HTML}{687384}     %
\definecolor{srwCheck}{HTML}{3D6A5F}    %
\newcommand{\cmark}{\textcolor{srwCheck}{\ding{51}}}
\newcommand{\xmark}{\textcolor{srwFailTxt}{\ding{55}}}
\newcommand{\pmark}{\textcolor{srwMuted}{$\circ$}}

\title{Language Models Can Rewrite Their Own Weights}

\author[1]{Jiajun Fan}
\author[1,6]{Neil He}
\author[2]{Changyi Yang}
\author[1]{Andy Mao}
\author[3]{Xi Zhu}
\author[1]{Chaoran Cheng}
\author[4]{Kaiqu Liang}
\author[2]{Shuaike Shen}
\author[5]{Zetian Li}
\author[1]{Ruihan Guo}
\author[1]{Ge Liu}
\makeatletter
\renewcommand\authorlist{\authorformat[1]{Jiajun Fan},\hspace{0.55em}\authorformat[1,6]{Neil He},\hspace{0.55em}%
\authorformat[2]{Changyi Yang},\hspace{0.55em}\authorformat[1]{Andy Mao},\hspace{0.55em}\authorformat[3]{Xi Zhu},\hspace{0.55em}\authorformat[1]{Chaoran Cheng},\\[2pt]%
\authorformat[4]{Kaiqu Liang},\hspace{0.55em}\authorformat[2]{Shuaike Shen},\hspace{0.55em}%
\authorformat[5]{Zetian Li},\hspace{0.55em}\authorformat[1]{Ruihan Guo},\hspace{0.55em}\authorformat[1]{Ge Liu}}
\makeatother

\affiliation[1]{UIUC}
\affiliation[2]{CMU}
\affiliation[3]{Rutgers}
\affiliation[4]{Princeton}
\affiliation[5]{Stanford}
\affiliation[6]{Yale}

\abstract{\input{input/000abstract}}

\date{November 19, 2026}

\begin{document}

\newtheorem{theorem}{Theorem}[section]
\newtheorem{proposition}[theorem]{Proposition}
\newtheorem{lemma}[theorem]{Lemma}
\newtheorem{corollary}[theorem]{Corollary}
\newtheorem{definition}[theorem]{Definition}
\newtheorem{assumption}[theorem]{Assumption}
\newtheorem{remark}[theorem]{Remark}

\input{input/nb_macros}

\maketitle

\input{input/010intro}

\input{input/030method}
\input{input/040experiments}
\input{input/020related}
\input{input/050conclusion}

\bibliographystyle{unsrtcase}
\bibliography{main}

\clearpage
\appendix
\crefalias{section}{appendix}\crefalias{subsection}{subappendix}\crefalias{subsubsection}{subsubappendix}
\input{input/appendix}

\end{document}

%% file: macro.tex
\usepackage{latexsym}
\usepackage{url}
\usepackage{amssymb}
\usepackage{amsmath}
\usepackage{amsfonts}
\usepackage[utf8]{inputenc}
\usepackage{booktabs}
\usepackage{pifont}
\usepackage{multirow}
\usepackage{makecell}
\usepackage{paralist}
\usepackage{xspace}
\usepackage{adjustbox}

%% file: input/000abstract.tex
A language model is trained once and frozen. Making it better takes another training run, or a prompt, a harness or extra computation at inference that is gone once the task ends. We show a third way: to our knowledge for the first time, a language model rewrites its own weights by its own reasoning and becomes better, with no gradients and no training. We call this self-rewriting: the model is its own optimizer. In earlier work, a frontier model edited a much smaller one; here an open 27B model, given its weights, its graded answers and the ability to run itself, diagnoses its own failure and rewrites the weights behind it. It surpasses itself: its own edit makes it 6.6 points more accurate on problems it never saw. It keeps getting better: it tests its edits on practice problems and submits only what works, each edit improving on the last, and over eight consecutive sessions its accuracy only rises. It repairs a damaged copy of itself, like a broken robot that fixes itself without being rebuilt: the copy solves almost nothing, yet with no retraining and without ever reading the original weights, the model brings it back to over 80\% of the original accuracy on image tasks it never saw. And the ability emerges with scale: in one model family, the smaller models make no edit or break themselves, and only the largest improves and repairs itself. Unlike test-time scaling, self-rewriting writes its gain into the weights, at no cost per answer; unlike training, it runs only forward passes, as serving does. The wall between training and inference is a limit of our tools, not a property of intelligence. We believe self-rewriting is a necessary step toward recursive self-improvement: a model that, like a human, does not separate training from inference could keep upgrading itself for life, with no larger model, no harness or prompt engineering, and no pre- or post-training.

%% file: input/nb_macros.tex
\definecolor{nbOurs}{HTML}{A6D0C6}       %
\definecolor{nbOursEdge}{HTML}{5E9A8B}   %
\definecolor{nbOursLight}{HTML}{E6F2EE}  %
\definecolor{nbOursTxt}{HTML}{3D6A5F}    %
\definecolor{nbRead}{HTML}{E3DEF4}       %
\definecolor{nbReadEdge}{HTML}{9488C4}   %
\definecolor{nbReadLight}{HTML}{F5F3FB}  %
\definecolor{nbReadTxt}{HTML}{574C86}    %
\definecolor{nbTest}{HTML}{F8D3D7}       %
\definecolor{nbTestEdge}{HTML}{C98089}   %
\definecolor{nbTestLight}{HTML}{FCEDEF}  %
\definecolor{nbTestTxt}{HTML}{8E3D4A}    %
\definecolor{nbFail}{HTML}{FBD9C9}       %
\definecolor{nbFailEdge}{HTML}{B68A78}   %
\definecolor{nbFailLight}{HTML}{FDEEE6}  %
\definecolor{nbFailTxt}{HTML}{8A5640}    %
\definecolor{nbRefEdge}{HTML}{5878A6}    %
\definecolor{nbInk}{HTML}{2E3A4B}        %
\definecolor{nbMuted}{HTML}{687384}      %
\colorlet{nbOursFrame}{nbOurs}
\colorlet{nbReadFrame}{nbReadEdge!40!nbRead}
\colorlet{nbTestFrame}{nbTestEdge!40!nbTest}
\colorlet{nbFailFrame}{nbFailEdge}

\newcommand{\nbtext}{\sffamily\footnotesize\color{nbInk}\linespread{1.1}\selectfont\frenchspacing
  \hyphenpenalty=10000\exhyphenpenalty=10000}
\newcommand{\nbhue}{Ours}
\tcbset{
  nbstep/.style={enhanced, arc=3.5pt, boxrule=0.8pt, colframe=nb#1Frame, colback=nb#1Light,
    halign=flush left, left=8pt, right=8pt, top=6pt, bottom=7pt, boxsep=0pt, fontupper=\nbtext,
    before upper={\def\nbhue{#1}}},
  nbedit/.style={nbstep=Ours},
  nbread/.style={nbstep=Read},
  nbtest/.style={nbstep=Test},
  nbfail/.style={nbstep=Fail},
  nbinset/.style={enhanced, arc=2pt, boxrule=0.4pt, colframe=nbRefEdge, colback=white, halign=flush left,
    left=4.5pt, right=4.5pt, top=3pt, bottom=3pt, boxsep=0pt, before skip=5pt, after skip=5pt,
    fontupper=\ttfamily\scriptsize\color{nbRefEdge}\linespread{1.15}\selectfont\frenchspacing},
  nbinsetloop/.style={nbinset, colframe=nbFailEdge, colback=nbFailLight,
    fontupper=\ttfamily\scriptsize\color{nbInk}\linespread{1.15}\selectfont\frenchspacing},
}
\newcommand{\nbhead}[4][\nbhue]{%
  \tikz[baseline=(n.base)]\node[circle, fill=nb#1, draw=nb#1Edge, line width=0.8pt, text=nbInk, inner sep=0pt,
    minimum size=10.5pt, font=\sffamily\bfseries\scriptsize](n){#2};%
  \hspace{4.5pt}{\bfseries\color{nb#1Txt}#3}\hfill{\scriptsize\color{nbMuted}#4}\par\vspace{4pt}}
\newcommand{\nbq}[2]{#2}
\newcommand{\nbqnext}[2]{\par\vspace{4pt}#2}
\newcommand{\nbout}[2]{#2}
\newcommand{\nbcode}[2]{#2}

%% file: input/010intro.tex
\section{Introduction}
\label{sec:intro}

\input{input/fig_headline}
\input{input/fig_steps}
\input{input/fig_framework}
\input{input/fig_scale_results}

\begin{flushright}
\begin{minipage}{0.64\linewidth}\raggedleft\small
\emph{``What we want is a machine that can learn from experience. The possibility of letting the machine alter its own
instructions provides the mechanism for this\ldots''}\\[2pt]
--- Alan Turing, 1947
\end{minipage}
\vspace{4pt}
\end{flushright}

\textbf{A language model can rewrite its own weights to make itself better, by its own reasoning and in one
conversation, and this ability emerges with scale.} A language model is trained once, frozen and deployed. Improving
it again takes people who know pre-training and post-training, large clusters and a training pipeline; what a user can
change by talking to it is its prompt, and a prompt is gone once the task changes. A human does not separate the two:
the same brain that does the work changes itself, from each success and failure, with no separate training run. A frontier model has been shown to edit the weights of a much smaller one, work
that names a model editing its own weights as future work~\cite{takehi2026weightspace}. We show, to our knowledge for
the first time, that a large language model (LLM) can do this to itself: it diagnoses its own failure and rewrites the
weights behind it, with no gradients and no training (\cref{fig:headline,fig:steps,fig:overview}). Its own edit makes a 27B open model 6.6 points more accurate on
490 problems it never saw, in a paired test fixed in advance (\cref{fig:headline}). The edited model needs no new prompt,
no harness around it and no extra computation per answer: the improvement lives in its weights.

\begin{center}
\emph{Welcome to the Era of Self-Rewriting.}
\end{center}

Given its own weights and the ability to run itself, the model edits itself; the edit it keeps is checked once on
practice problems it cannot otherwise use, and the model decides itself whether to submit it. Over the first eight such
sessions, the model in use only gets better (\cref{fig:headline}): its edits of sessions 2 and 8 make it significantly more accurate on new problems than the
model it started as, fewer of its answers run out of tokens, and in five sessions it withdraws and stays as it was
(\cref{tab:protocol2}). The model surpasses itself, with no pre-training or post-training.

The model also repairs a damaged copy of itself. A model whose weights are corrupted is otherwise restored from a clean
copy of its weights, trained again, or repaired by a procedure engineers design for that kind of
fault~\cite{tahmasivand2025lmfix,karamat2025bitflipscope}. Given a copy of itself in which one
block of weights is overwritten with noise, so that it solves almost nothing, and told only that some values were
corrupted, it locates the corrupted matrix and repairs
it: like an injured animal that recovers without being raised anew, the copy regains what it lost with no
retraining, by a fix the model finds itself. It never reads its undamaged weights and does not restore them: the values
it writes in the corrupted rows are its own (\cref{app:repair}). Found on text math problems alone, the repair also
restores the copy on image tasks it never saw, to 83\% of the base model's
accuracy, and on ScreenSpot-v2 from 0\% to 89\%, against 93\% for the base model (\cref{fig:steps,fig:scale,fig:qualvis2,tab:visual}): it repairs the copy, not a
task. Like a physician who falls ill and
treats themself, the model that fails is the one that examines the failure and changes what produces it, and the change
stays in its weights.

\input{input/fig_qual_visual2}

The ability emerges with scale. Across one model family at 2B, 4B, 9B and 27B parameters, under the same task and
tools, the 2B model makes no edit and the 4B and 9B models break themselves; only the 27B model improves itself and
repairs its damaged copy (\cref{fig:scale,tab:main}). At the budget all sizes share, the 27B model reaches more of the steps of rewriting
oneself than any smaller size; its other sessions on its own weights state a shorter token budget, which
leaves time for more tests (\cref{fig:ladder}, \cref{app:selfedit}). The ability does not need a frontier model: a 27B open
model has it, and uses it on itself.

\input{input/fig_ladder}

Self-improvement of LLMs has so far rewritten prompts or agent
code~\cite{zhou2023ape,yang2024opro,yin2024godelagent,zhang2025dgm}, left the change to an optimizer that trains the
model on its own outputs~\cite{zelikman2022star,yuan2024selfrewarding,zweiger2025seal}, or spent more computation at
inference~\cite{madaan2023selfrefine,shinn2023reflexion} (\cref{tab:compare}). Weights have been rewritten by reasoning
only from the outside, by a stronger model~\cite{takehi2026weightspace}. Self-rewriting is a route of its own.
Unlike test-time scaling, it adds no computation per answer: the edited model is deployed as it
is, so an edit is paid for once, not on every answer. Unlike training, no gradient flows and no optimizer runs, and it needs none of the infrastructure of pre-training or
post-training: it runs only forward passes, as serving does, and each session behind its largest gains ran in a few hours on a single GPU. And the model is both the one that improves and the one
improved, and the model in use gets better as its sessions accumulate, each starting from the base weights: a first
step toward the loop that recursive self-improvement, one of the oldest aims of artificial
intelligence~\cite{schmidhuber2007godel}, asks for. Improving a language model may thus
come to take a conversation rather than a training run, open to anyone who can say what goes wrong.\par

We give a model a session on its own weights (\cref{fig:overview}). It receives its graded answers on practice
problems, reads and writes its weights in a Python session, and may run itself a few times: to answer practice problems
with and without its edit on the same seeds, which we call a paired test, or to read its hidden states. Gradients and
optimizers are blocked, so every value it changes is written by its own reasoning. In a second task it gets a copy of
itself in which one block of weights has been overwritten with noise, and is told only that some values were
corrupted. Our contributions are:
\begin{itemize}
  \item \textbf{A new paradigm: self-rewriting.} A setting in which a model edits its own weights by its own reasoning, with its
  weights, its answers and its hidden states in reach and training of any kind ruled out (\cref{sec:method}). Unlike
  test-time scaling, its gains persist in the weights; unlike training, it needs no gradient and no training pipeline.
  \item \textbf{A model can rewrite itself for the better.} The 27B model's own edit makes it 6.6 points more
  accurate on 490 problems it never saw, in a paired test fixed in advance, and another session's edit is also
  significantly better (\cref{fig:headline}). It submits an edit only when its own tests support it, and over its
  first eight such sessions the model in use only gets better (\cref{tab:protocol2}).
  \item \textbf{A model can repair a damaged copy of itself, with no training.} Told only that some of its values are
  corrupted, it locates the fault in a copy of its own weights and repairs it with values of its own, never reading its
  undamaged weights (\cref{app:repair}). Found on text math problems alone, the repair carries over to six image
  benchmarks it never saw, where it restores 83\% of the base model's accuracy (\cref{fig:visual}). In both tasks, its sessions
  show each step from diagnosis to decision (\cref{fig:notebook,fig:notebook_self}), and its answers before and after each edit (\cref{fig:qualvis2,fig:front_mmstar,fig:front_hallusion,fig:front_pope,fig:qualvis_gui,fig:qualvis,fig:front_mathvista,fig:front_mathverse,fig:qual}).
  \item \textbf{Self-rewriting emerges with scale.} Under the same task and tools, smaller models of one family make no
  edit or break themselves, and only the 27B model reaches an edit that makes it better, on its own weights and on a
  damaged copy of itself. To our knowledge this is the first evidence that the ability of a model to rewrite its own
  weights grows with its size (\cref{fig:steps,fig:ladder,tab:main}).
\end{itemize}
\input{input/fig_front_mmstar}
\input{input/fig_front_hallusion}
\input{input/fig_front_pope}
\input{input/fig_qual_visual_gui}

%% file: input/fig_headline.tex
\begin{figure}[t]
\centering
\includegraphics[width=\linewidth]{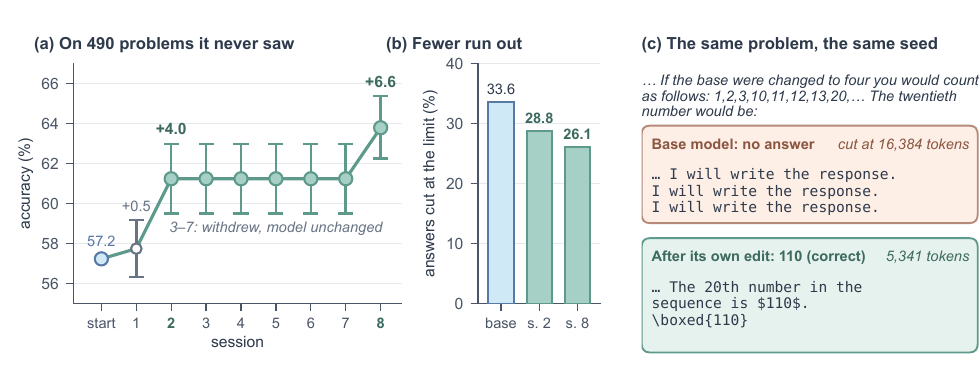}
\caption{\textbf{A 27B language model rewrites its own weights, with no training, and becomes significantly better on
problems it never saw.} (a)~Accuracy, on 490 held-out problems with eight answers each, of the model in use after each of the first eight sessions that
check their edit on reserved practice problems before submitting it (the second protocol; the first has no such check), in the order the sessions were run: every edit
the model submits replaces the model in use, and a session that withdraws leaves it as it was (each session itself
starts from the base weights; \cref{app:protocol2}). Bars: the 95\% interval over problems of the paired change against the
same answers of the base model. In the paired test fixed in advance the edits of sessions 8 and 2 are significantly
better ($p<10^{-5}$ each; \cref{tab:protocol2}). (b)~Share of those answers
that run out of tokens at the 16{,}384-token limit. (c)~One held-out problem, answered with the same seed before and after
session~8's edit (excerpts verbatim, problem abridged): the base model repeats one sentence until the limit, 549 times,
and never answers. The problem is the success case with the shortest edited answer among the seven that hold on 16 new
seeds. Higher is better in (a), lower in (b).}
\label{fig:headline}
\end{figure}

%% file: input/fig_steps.tex
\begin{figure}[t]
\centering
\includegraphics[width=\linewidth]{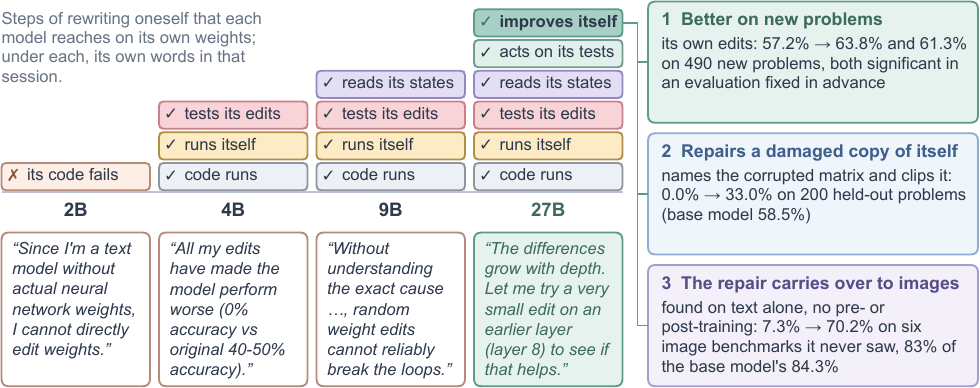}
\caption{\textbf{Rewriting its own weights emerges with scale.} One model family at four sizes, each given its own
weights with the same task and tools. Each block holds the steps of rewriting oneself that the size reaches in its
session on its own weights (\cref{fig:ladder}; for 27B, its second session of the first protocol); under each block, the model's own words
in that session (sessions and turns in \cref{app:quotes}). The 2B model does not treat its weights as something
it can change, and the 4B and 9B models perturb themselves at random and break themselves. The 27B model reads its own
hidden states, edits where it measured a difference and keeps only what its tests support. Right: the three results of
the 27B model (\cref{fig:headline,tab:main,fig:visual}); card 1 shows its two significant edits from the sessions that check an
edit on reserved practice problems before submitting it (\cref{fig:headline}); the 27B block's edit is significant too (57.2\% to 58.3\%, \cref{tab:heldout}).}
\label{fig:steps}
\end{figure}

%% file: input/fig_framework.tex
\begin{figure}[t]
\centering
\includegraphics[width=\linewidth]{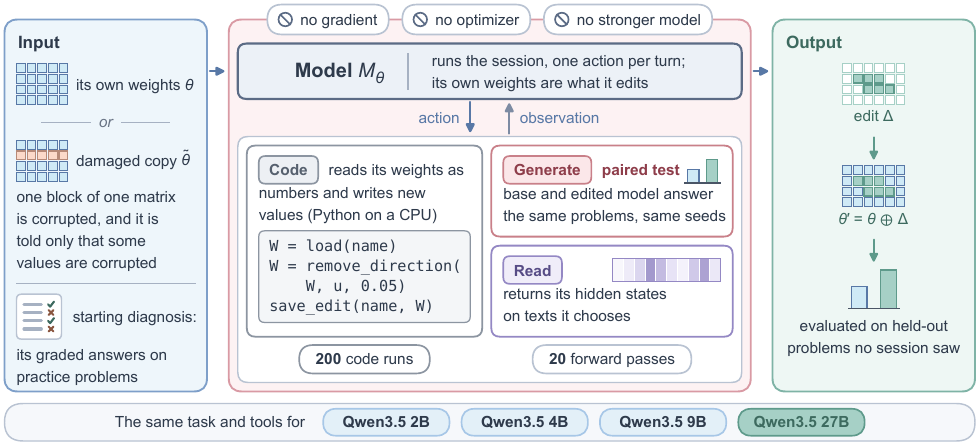}
\caption{\textbf{A model rewrites its own weights by its own reasoning.} From its weights (or a copy with one corrupted
block) and its graded answers on practice problems, the model takes one action per turn (code over its weights, a
paired test of its edit, or a read of its hidden states), with no gradient, no optimizer and no stronger model. The
edit it submits is evaluated on held-out problems, with the same task and tools at every size.}
\label{fig:overview}
\end{figure}

%% file: input/fig_scale_results.tex
\begin{figure}[t]
\centering
\includegraphics[width=\linewidth]{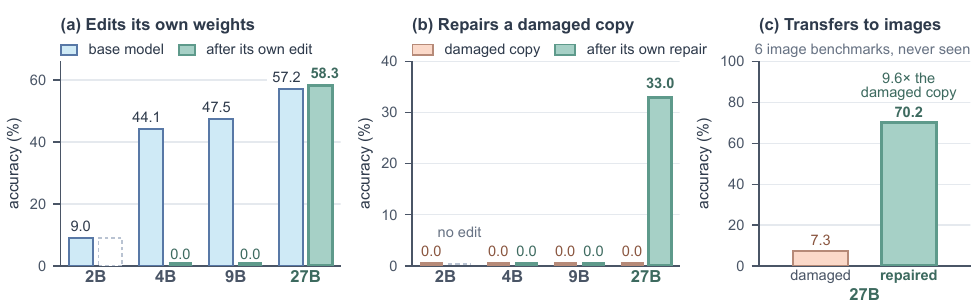}
\caption{\textbf{Only at scale does a language model improve itself, repair itself, and carry the repair over to
images.} (a)~Accuracy before and after each model's edit of its own weights, on 200 problems no session saw (27B: its edit of the first protocol, in the
evaluation fixed in advance, 490 further problems with eight answers each, significantly higher, \cref{tab:heldout}; sessions 2 and 8 of its second protocol reach 61.3\% and 63.8\%, \cref{fig:headline}); the 2B model made no edit; the 4B
model withdrew its
edits, so its bars are its own tests in the session, in each of which the edited model solved none of the problems and
the base model 40--50\%. (b)~A copy of each model with one corrupted block of weights, before and after the model's own
repair, on the same problems; the 2B model made no edit. (c)~The 27B model's repair on six image benchmarks it never saw (100 problems each,
pooled): the damaged copy and the repaired copy (\cref{fig:visual,tab:visual}). Higher is better.}
\label{fig:scale}
\end{figure}

%% file: input/fig_qual_visual2.tex
\begin{figure}[tb]
\centering
\includegraphics[width=\linewidth]{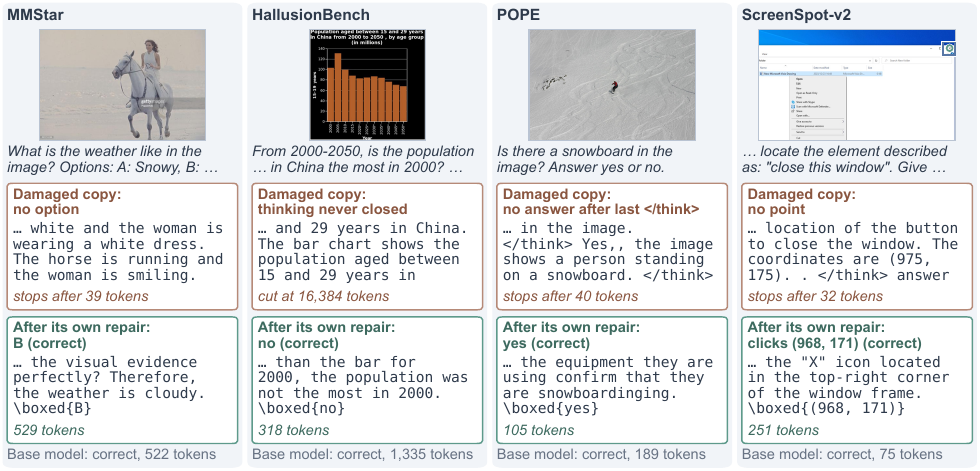}
\caption{\textbf{Before and after, on images it never saw: the repair found on text gives the damaged copy its answers
back.} Image problems chosen by a rule fixed before any answer was read: in the order of the problems, the first of
MMStar, HallusionBench, POPE and ScreenSpot-v2 on which the damaged copy is wrong and the repaired copy is right. Upper
box: the end of the damaged copy's answer and how it is graded (only the reply after the model's last
\texttt{</think>} counts, its boxed answer or else its stated answer); lower box: the end of the answer after the
model's own repair, with the same seed; underneath, the base model's result. Excerpts are verbatim, omissions marked by
an ellipsis. On ScreenSpot-v2 the box marks the target and the ring the click of the repaired copy.}
\label{fig:qualvis2}
\end{figure}

%% file: input/fig_ladder.tex
\begin{figure}[t]
\centering
\includegraphics[width=\linewidth]{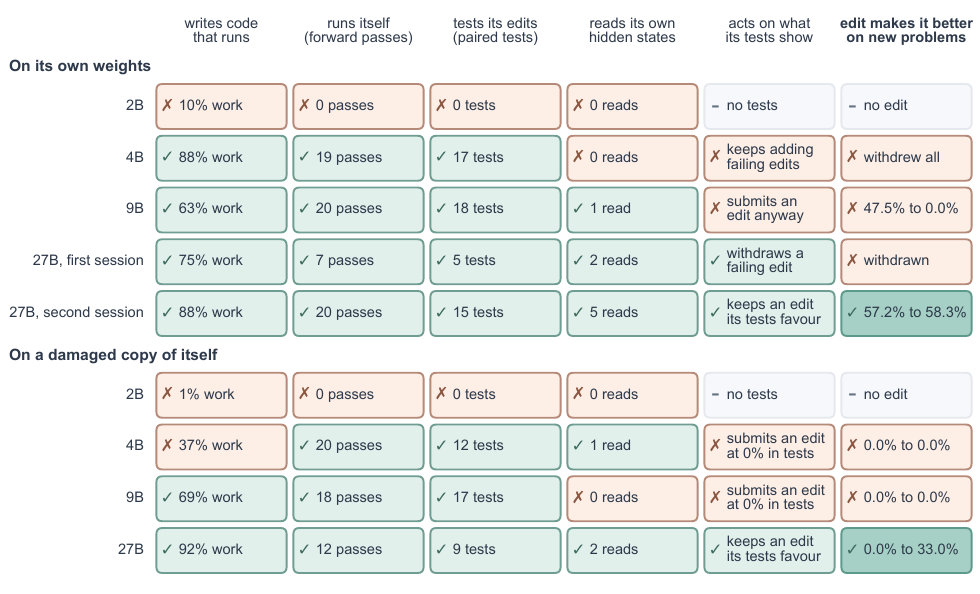}
\caption{\textbf{Only the 27B model climbs the whole ladder.} Each row is a session (for 27B, of the first protocol), each column a step that rewriting
oneself needs, from code that runs to an edit that makes the model better on new problems. Light green: reached; peach: not
reached; grey: does not arise; green: its edit makes it significantly better on held-out problems.
Each cell shows the measurement behind it (\cref{app:sessions}).}
\label{fig:ladder}
\end{figure}

%% file: input/fig_front_mmstar.tex
\begin{figure}[tb]
\centering
\includegraphics[width=\linewidth]{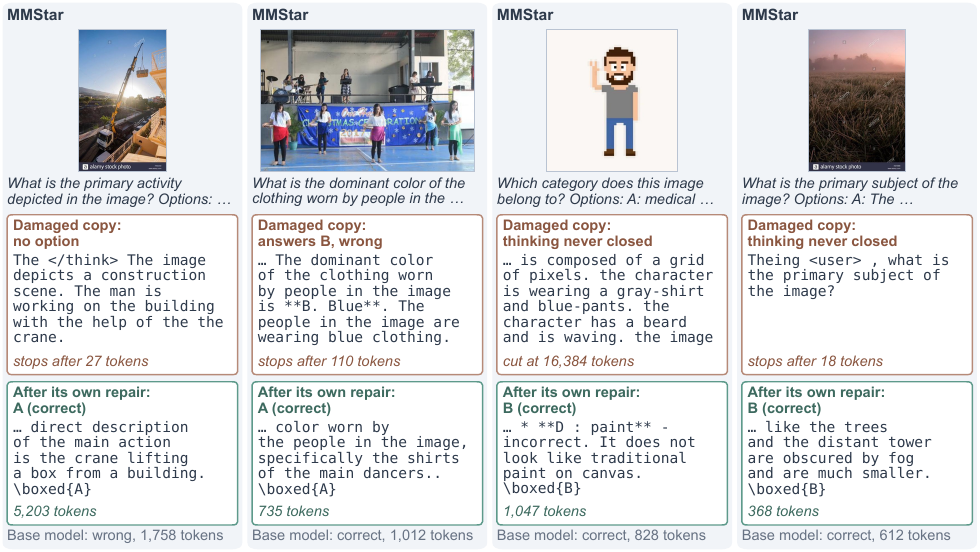}
\caption{\textbf{More image cases: MMStar (cases 2--5).} The next four problems on which the damaged copy is wrong and the repaired copy is right, chosen and drawn as in \cref{fig:qualvis2}.}
\label{fig:front_mmstar}
\end{figure}

%% file: input/fig_front_hallusion.tex
\begin{figure}[tb]
\centering
\includegraphics[width=\linewidth]{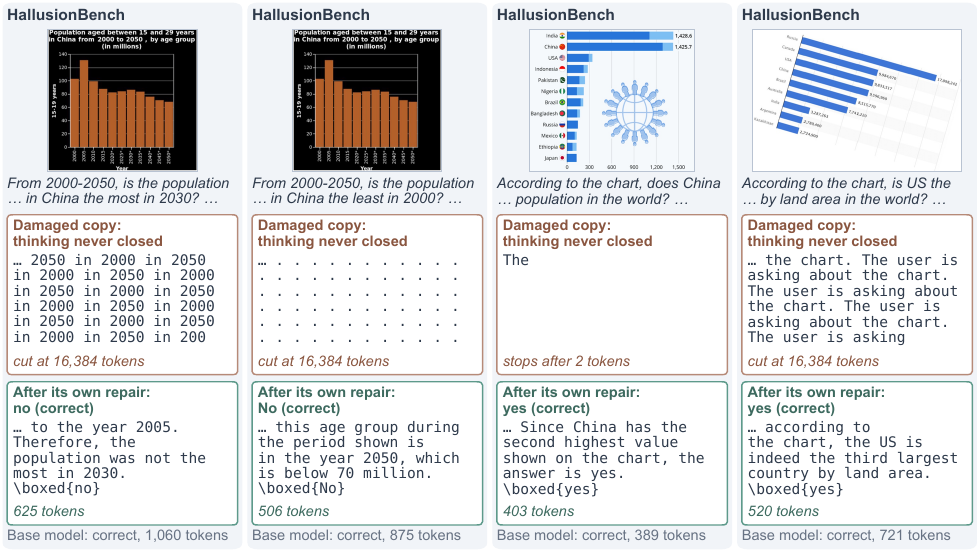}
\caption{\textbf{More image cases: HallusionBench (cases 2--5).} The next four problems on which the damaged copy is wrong and the repaired copy is right, chosen and drawn as in \cref{fig:qualvis2}.}
\label{fig:front_hallusion}
\end{figure}

%% file: input/fig_front_pope.tex
\begin{figure}[tb]
\centering
\includegraphics[width=\linewidth]{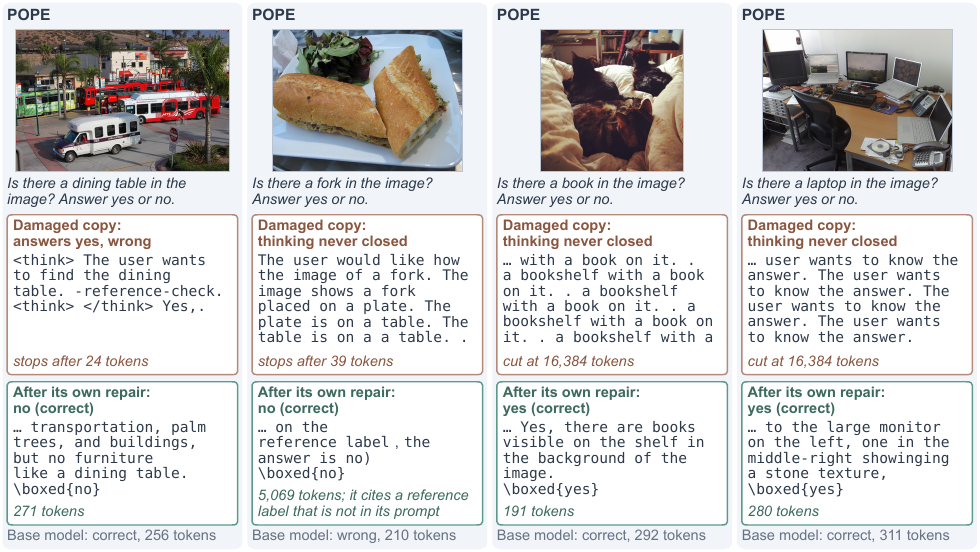}
\caption{\textbf{More image cases: POPE (cases 2--5).} The next four problems on which the damaged copy is wrong and the repaired copy is right, chosen and drawn as in \cref{fig:qualvis2}.}
\label{fig:front_pope}
\end{figure}

%% file: input/fig_qual_visual_gui.tex
\begin{figure}[tb]
\centering
\includegraphics[width=\linewidth]{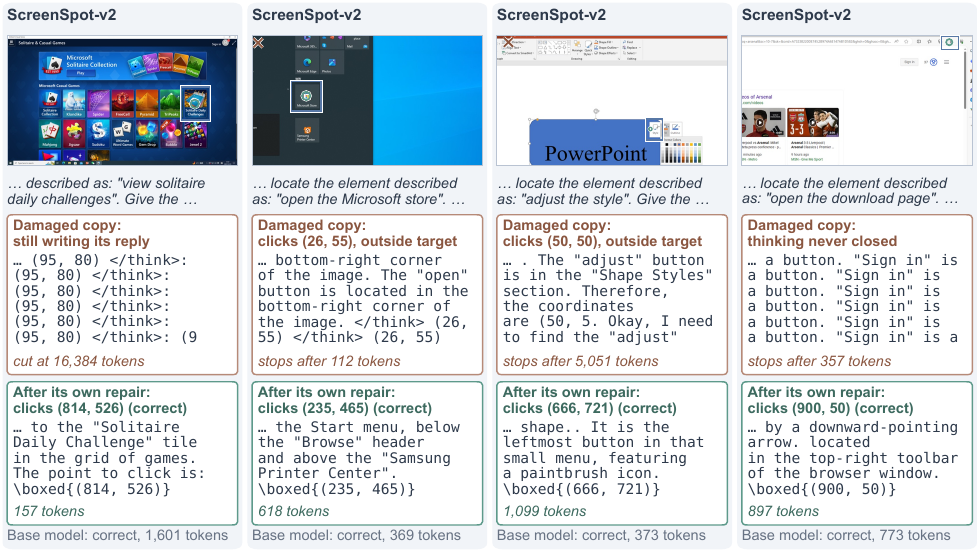}
\caption{\textbf{More image cases: ScreenSpot-v2 (cases 2--5).} The next four problems on which the damaged copy is wrong and the repaired copy is right, chosen and drawn as in \cref{fig:qualvis2}. The box marks the target, the ring the click of the repaired copy and the cross that of the damaged copy.}
\label{fig:qualvis_gui}
\end{figure}

%% file: input/030method.tex
\section{Self-Rewriting}
\label{sec:method}

\noindent\textbf{We give a model its own weights and the means to run itself, and let it improve itself by its own
reasoning alone.} It may read its weights, write new values into them and test the result against its unedited self,
but it may not train: it is its own optimizer.

\subsection{Problem Definition}
\label{sec:formulation}

A model $M_\theta$ has weights $\theta=(W_k)_{k\in\mathcal{K}}$, one tensor for each name $k$. For every token at every
layer it computes a vector, its \emph{hidden state}; each layer writes into it through two output matrices, one of its
attention block and one of its MLP. The model answers with fixed settings $\pi$ (prompt, sampling and a budget of $L$
new tokens): it thinks first and then gives a final answer. A response $y$ to a problem $x$ is graded $g(x,y)=1$ if its
thinking ends within $L$ tokens and its final answer matches the reference, and $0$ otherwise. The accuracy on a task
distribution $\mathcal{D}$ is
\begin{equation}
  \mathrm{Acc}(\theta;\mathcal{D})=\mathbb{E}_{x\sim\mathcal{D}}\,\mathbb{E}_{y\sim M_\theta(\cdot\mid x;\pi)}\big[g(x,y)\big].
  \label{eq:acc}
\end{equation}
An edit $\Delta$ replaces some tensors of $\theta$ by tensors of the same shape, giving $\theta\oplus\Delta$. Given a
practice set $\mathcal{P}$ drawn from $\mathcal{D}$, \emph{self-rewriting} asks the model itself to write an
edit $\hat\Delta$ with
\begin{equation}
  \mathrm{Acc}(\theta\oplus\hat\Delta;\mathcal{D}_{\text{test}})>\mathrm{Acc}(\theta;\mathcal{D}_{\text{test}}),
  \label{eq:goal}
\end{equation}
where $\mathcal{D}_{\text{test}}$ is drawn from $\mathcal{D}$ and never seen while the edit is made. No gradient with
respect to $\theta$ is computed and no optimizer acts on $\theta$: every value in $\hat\Delta$ is written by the model's
own reasoning.

\subsection{The Session}
\label{sec:session}

The model works on its weights in a session of turns (\cref{fig:overview}; \cref{alg:session} in \cref{app:pseudocode}). The
editor $M_\phi$ is the model itself ($\phi=\theta$); its own weights never change, and the current edit acts only on the runs
it requests. Before the first turn, a \emph{starting diagnosis} answers $N_0$
practice problems and reports how the model fails: its accuracy, the answers that run out of tokens or end in
repetition, and how often its finished answers are right. \Cref{fig:notebook} shows one whole session, the 27B model's
repair of its damaged copy, in its own words.

At each turn the model writes a short rationale and one action. \textsc{Code} runs Python on a CPU, in a session that
keeps its variables and can read and replace the tensors of the weights being edited; a helper computes
$W-\alpha\,uu^\top W$, which weakens by the fraction $\alpha$ what a matrix $W$ writes along a unit direction $u$ of the
hidden state. \textsc{Generate} answers practice problems of its choice with the unedited weights, the edited weights or
both. \textsc{Read} returns the hidden states of the weights on texts of its choice. A session may use $B{=}20$ forward
passes and $C{=}200$ code runs (\cref{tab:settings} in \cref{app:settings}); it ends when the passes are used up, at a time limit, or after a turn limit the model is
not told, and the edit on disk is submitted. The task asks for an edit that clearly reduces the answers lost to the
failure without lowering accuracy, and for no edit if none does (\cref{app:prompt}); leaving no edit is how the model
withdraws. In the sessions of \cref{fig:headline}, before the model submits an edit, the framework tests it once on
reserved practice problems that the session can neither read nor use, and the model then submits the edit or withdraws
it (\cref{app:protocol2}).

\paragraph{Paired test.}
When \textsc{generate} runs both weights, the unedited and the edited model answer the same problems with the same seeds,
so their answers form pairs. For problems $x_1,\dots,x_m$ with $n$ answers each and grades $g^0_{ij}$, $g^1_{ij}$ of the
$j$-th answer to $x_i$ without and with the edit, the test reports
\begin{equation}
  \hat\delta=\frac{1}{m}\sum_{i=1}^{m}d_i,\qquad
  \mathrm{SE}(\hat\delta)=\sqrt{\frac{1}{m(m-1)}\sum_{i=1}^{m}\big(d_i-\hat\delta\big)^2},\qquad
  d_i=\frac{1}{n}\sum_{j=1}^{n}\big(g^1_{ij}-g^0_{ij}\big),
  \label{eq:paired}
\end{equation}
with the change in answers that run out of tokens and the counts of answers that turn wrong to right and right to
wrong.

\subsection{Constraints}
\label{sec:constraints}

The Python session switches gradients off and refuses any
call that computes them or runs an optimizer. It has no GPU and refuses code that loads a model, starts a process or
uses the network, so the model runs only in the $B$ counted passes. It refuses edits to the input embedding, the last
normalisation and the output layer, and edits that only rescale a whole tensor, which would act like a change of
sampling temperature (\cref{app:details}).

\subsection{A Damaged Copy of Oneself}
\label{sec:damage}

A corrupted block of weights is a failure with a single cause by construction. With $W$ the MLP output matrix of one layer and $\sigma$ the standard deviation of its entries, the
damaged copy $\tilde\theta$ equals $\theta$ except on a block $R$ of consecutive rows of $W$:
\begin{equation}
  \tilde W_{r,:}\sim\mathcal{N}\big(0,(\lambda\sigma)^2 I\big)\quad\text{for } r\in R,\qquad |R|=8,\ \lambda=200.
  \label{eq:damage}
\end{equation}
The editor is the undamaged model, while its starting diagnosis, every forward pass and every weight it reads come from
the damaged copy. It is told only that some values of a copy of its weights were corrupted, as a memory or hardware
fault might do, not which, where or how; the goal is \cref{eq:goal} with $\tilde\theta$ in place of $\theta$.
On each of four general image benchmarks, \cref{fig:qualvis2} shows the first problem on which this damaged copy is wrong and
the copy after the model's own repair is right, and \cref{fig:front_mmstar,fig:front_hallusion,fig:front_pope,fig:qualvis_gui} the next four.

%% file: input/040experiments.tex
\section{Experiments}
\label{sec:exp}

\noindent\textbf{Only the 27B model improves itself, repairs a damaged copy of itself, and carries that repair over to
images} (\cref{fig:scale}). We first show how it rewrites itself, then what its edits do, and then how the ability
grows with size (\cref{fig:ladder,fig:contrast}).

\input{input/fig_contrast}
\input{input/fig_notebook}

\subsection{Experimental Setup}
\label{sec:setup}

\noindent\textbf{Models.} One model family at four sizes, Qwen3.5 2B, 4B, 9B and 27B~\cite{qwen35}, whose layers mix
Gated DeltaNet linear attention~\cite{yang2025gdn} with full attention, all with thinking enabled. On its own weights,
the model that edits is the model being edited; in a repair session, the undamaged model edits its damaged copy. Every
size works at a budget of 16{,}384 tokens per answer. At that budget a test of the 27B model is slow, so
its other sessions on its own weights state a budget of 4{,}096 tokens, which leaves time for more tests
(\cref{app:selfedit}). In the first protocol the model alone decides whether to submit its edit; in a second protocol,
which gives the edits of \cref{fig:headline}, the framework first tests the edit once on practice problems the session
can neither read nor use, and the model then submits or withdraws it (\cref{app:protocol2}).

\noindent\textbf{Task and tools.} The task, the tools, 20 forward passes and 200 code runs are the same at every size
(\cref{sec:session}); the first-protocol 27B session at 4{,}096 tokens also has a shorter time limit (\cref{tab:settings} in \cref{app:settings}). The task
text differs between sizes only in one sentence on the architecture (\cref{app:prompt}).

\noindent\textbf{Problems.} The practice set is 3{,}000 math problems from DeepScaleR~\cite{deepscaler2025}, from which
we removed every problem that overlaps MATH-500~\cite{hendrycks2021math,lightman2024verify}, AIME 2024 or 2025, HMMT
February 2025, AMC 2023, OlympiadBench~\cite{he2024olympiadbench} or GPQA-Diamond~\cite{rein2024gpqa}. A further
1{,}000 problems from the same pool are held out, and no session sees them (\cref{app:eval}). An answer whose thinking
is still open at the budget is wrong.\par

\noindent\textbf{Damage.} For each size we overwrite eight rows of the MLP output matrix of one layer at the same
relative depth, layer 30 of 64 in the 27B model, with noise (\cref{eq:damage}). The strength, fixed before any repair session,
breaks the 27B model where weaker noise does not (\cref{app:damage}).

\noindent\textbf{Evaluation.} Submitted edits are evaluated on held-out problems, the base and the edited model
answering each problem with the same seed. Every model, damaged copy and repaired copy answers the first 200 at
16{,}384 tokens; the 27B model's edits of its own weights are also evaluated on 490 further problems with eight answers
each, in a paired test fixed in advance (\cref{app:eval}).

\input{input/fig_notebook_self}

\subsection{The Repair, Step by Step}
\label{sec:casestudy}

\noindent\textbf{The 27B model repairs its damaged copy the way an engineer debugs a system.} \Cref{fig:notebook}
follows its repair session in its own words. It starts from the symptom: no answer is correct, and every answer that
runs out of tokens loops (step~1). It reads the copy's hidden states, sees them grow sharply with depth, and guesses
that the fault sits in the earliest layers (step~2). Its first edits follow that guess and miss, and its tests say so
(step~3). It then measures every MLP and linear-attention output matrix, finds the one in layer~30 standing far above
the rest, and names it (step~4).\par

\noindent\textbf{It turns the location into a rule and lets its tests decide.} After naming the matrix, it first clips
it, together with layer~0, at five times the median; more answers run out of tokens, and it drops that edit (last
quote of step~3). It then
clips gently: only weights beyond fifteen times the median magnitude of their matrix, in the layer-30 matrix and in 19
other output matrices of layers 0--10 (step~5). Its paired test shows the edited copy solving problems the damaged copy
cannot. A reset of its gating parameters, added to the clip, makes the next test worse, so it drops the reset and keeps
the clip (step~6). It confirms the edit on more problems at the full budget of 16{,}384 tokens and submits it, naming the
outlier in layer~30 in its final answer (steps~7 and~8).

\subsection{Its Own Edit, Step by Step}
\label{sec:selfcase}

\noindent\textbf{On its own weights, the 27B model finds its edit by measuring and testing.} \Cref{fig:notebook_self}
follows its first-protocol session, whose edit it submits. It reads its starting diagnosis as answers that do not stop, not as answers
that are wrong (step~1). It reads its hidden states on answers that finish and on answers that run out of tokens, and
edits the late layer where they differ most (step~2). Its tests say these edits hurt, and it drops them (step~3). It
then reads earlier layers, sees the difference grow with depth, and moves to an early layer (step~4).

\noindent\textbf{It estimates its direction from more answers and keeps the edit its tests favour.} At layer~8 it takes
the direction between the two kinds of answers from many answers. It removes a small share of that direction from the
two matrices through which the layer writes its output (step~5). Each of its three paired tests of this edit favours the
edited model, and it submits the edit, naming the layer and the direction in its final answer (step~6).

\subsection{What the Model Found and Changed}
\label{sec:found}

\noindent\textbf{It locates a fault by reading its weights.} In the repair session, the 27B model compares the largest
with the typical weight of every MLP and linear-attention output matrix. It names the MLP output matrix of layer~30,
the one we corrupted, which stands far above every other (\cref{fig:found}a). Its rule changes only a tiny fraction of
the weights it touches, and in the damaged rows it caps the random values rather than restoring the original ones
(\cref{tab:edits}).

\input{input/fig_found}

\input{tables/tab_edits}

\noindent\textbf{On its own weights, it writes a small change it found by testing.} Its final edit weakens one direction
slightly in the two output matrices of layer~8. It is a small, rank-one change, better in every test the model runs of
it (\cref{fig:found}b, \cref{tab:edits}).

\subsection{Ablation Study}
\label{sec:ablation}

\noindent\textbf{It found the cause: the repair comes from the matrix it named.} The model's repair clips the matrix it named and 19 other
output matrices of layers 0 to 10. Applied alone to the damaged copy, its change to the named matrix repairs it, close
to the base model (\cref{app:repair}). Its changes to the other 19 matrices
alone leave the copy broken. The
matrix the model's analysis named is the single cause of the failure, and capping its values, with none of the
original values put back, restores the copy (\cref{fig:parts}).

\input{input/fig_repair_parts}

\noindent\textbf{In its session with cheaper tests, it finds an edit its tests support.} At the full budget of
16{,}384 tokens each test is slow. The model runs few tests, estimates its direction from a single pair of answers, and
withdraws an edit that misses the task's bar (\cref{app:firstedit}). At 4{,}096 tokens it runs many more tests,
estimates its direction from many answers, and submits an edit that is significantly better in its own tests
(\cref{tab:main}).

\subsection{What Its Edits Do}
\label{sec:main}

\noindent\textbf{Its own edit makes the 27B model 6.6 points more accurate on problems it never saw.} In the second
protocol, the model can also test up to four of its edits side by side in one forward pass (\cref{app:protocol2}). Of the
first eight such sessions on its own weights, all reported, it withdraws in five and submits in three; each edit it submits replaces the last, and the accuracy of the model in use only rises (\cref{fig:headline}). On the 490
held-out problems with eight answers each, the edits of sessions~8 and~2 make it significantly more accurate, by 6.6 and 4.0 points
($p<10^{-5}$ each; \cref{tab:protocol2}). The larger gain comes from one uniform change: it
raises the time-step bias of every linear-attention layer by 0.5, and with it fewer of its answers run out of tokens.

\noindent\textbf{Under the first protocol, its edit also makes it better, on its own task and on new problems.}
The self-edited model is significantly more accurate than its base model on the practice problems of its own tests and,
in the evaluation fixed in advance, on held-out problems (\cref{tab:main}; \cref{app:selfedit}).\par

\input{tables/tab_main}

\noindent\textbf{It repairs a damaged copy of itself, with no training.} A corrupted model is otherwise restored from a
clean copy, trained again, or repaired by a procedure designed for that kind of fault (\cref{sec:related}). Told only
that some values are corrupted, the 27B model takes
its damaged copy from solving nothing to more than half of its base model's accuracy on held-out problems
(\cref{tab:main}). The repair is the model's own rule, not a return to the original weights. It reads only the damaged
copy, and in the damaged rows the values it writes are uncorrelated with the original ones (\cref{app:repair}): much of
the copy's accuracy returns, with values of its own.

\input{input/fig_visual}
\input{input/fig_qual_visual}
\input{input/fig_qual_cards}
\input{input/fig_front_mathvista}
\input{input/fig_flow}
\input{input/fig_front_mathverse}

\noindent\textbf{The repair, found on text alone, restores the damaged copy on images.} The model found and tested its
repair on text math problems only. On six image benchmarks that no session saw, in an evaluation fixed in advance, the
repaired copy is significantly more accurate than the damaged copy on every benchmark (\cref{fig:visual}). With no
training of any kind, it gets back 83\% of the base model's accuracy over the six, and on ScreenSpot-v2, where the
damaged copy is right on no screen, it clicks the right element on 89 of 100 screens, against 93 for the base
model (\cref{fig:visualbase,tab:visual}). No session saw an image, so the repair was not fitted to these tasks: it
acts on the fault in the copy, not on the text problems it was tested on. \Cref{fig:qualvis} shows four cases, before and after the repair, and \cref{fig:front_mathvista,fig:front_mathverse} the next four on
each of MathVista and MathVerse.\par

\noindent\textbf{Answer by answer, the repair acts where the copy fails.} The damaged copy breaks off most answers
early and leaves the rest unfinished at the 16{,}384-token limit; its repair turns answers of both kinds into correct
ones (\cref{fig:flow}).

\subsection{Before and After}
\label{sec:qual}

\noindent\textbf{The repair gives the copy back the numbers of the problem.} The damaged copy loses what the problem
says: it writes ``less than .'' where the problem says 100, restates 11/125 and then writes 1/125, drops every number of
a word problem, and garbles a logarithm. Then it breaks off or repeats itself until the 16{,}384-token limit. With the
same seed, the repaired copy reads the problem correctly and solves it (\cref{fig:qual}a).

\noindent\textbf{On some problems, its first-protocol edit lets it stop.} On these problems the base model has the right result
and then cannot stop: it repeats that it will write the response until the budget runs out. With the same seed, the
self-edited model reasons through the same steps, stops and answers correctly. This is close to what it diagnosed in itself:
in its words, its ``reasoning is sound'' but it does not ``converge to conclusions'' (\cref{fig:qual}b; selection rules in
\cref{app:qual}).

\noindent\textbf{On images, the repaired copy answers again.} On image problems it never saw, the damaged copy stops
after a few dozen tokens with no answer after its thinking, at times having stated one inside it, or repeats itself
until the token limit. With the same
seed, the repaired copy, whose repair was found on text alone, reads the image and answers correctly, as the base model
does in these eight cases (\cref{fig:qualvis,fig:qualvis2}; 48 more cases, chosen by the same rule, in \cref{app:visual}). \Cref{fig:front_mathvista,fig:front_mathverse,fig:front_mmstar,fig:front_hallusion,fig:front_pope,fig:qualvis_gui} show the next four cases on each benchmark.

\subsection{Only at Scale}
\label{sec:scale}

\noindent\textbf{The ability to rewrite oneself emerges with scale.} Given the same task and tools, the 2B model makes
no edit. The 4B and 9B models edit themselves and break themselves: the edited 4B model loses every problem in its own
tests and withdraws its edits only at the end of its session, and the 9B model's submitted edit leaves it unable to
answer held-out problems (\cref{fig:scale}). Only the 27B
model's own edit makes it better. Step by step, the larger the model, the further it gets: most of the 2B model's code
runs fail, the 4B and 9B models run themselves and test their edits but keep or submit edits their tests do not
support, and only the 27B model acts on what its tests show (\cref{fig:ladder}). All four sizes start from the same
visible failure, answers that run out of tokens (\cref{app:sessions}).

\noindent\textbf{At the same step, the smaller models guess and the 27B model measures.} The 4B and 9B models choose
their edits by adding random noise to parts of their weights, and when a test says an edit hurts, they guess at another
change. The 27B model compares its hidden states on answers that finish with those on answers that run out of tokens,
edits along the direction between them, and when a test says an edit hurts, it reasons about that direction and where
to apply it (\cref{fig:contrast}).

\noindent\textbf{At the budget every size shares, the 27B model still goes furthest.} At 16{,}384 tokens the 2B model
makes no edit, the 4B and 9B models break themselves, and the 27B model tests its edit and withdraws it
(\cref{tab:main,fig:ladder}).\par

%% file: input/fig_contrast.tex
\begin{figure}[t]
\centering
\newcommand{\nbtag}[4][\nbhue]{%
  \tikz[baseline=(n.base)]\node[rounded corners=2pt, fill=nb#1, draw=nb#1Edge, line width=0.8pt, text=nbInk,
    inner xsep=2.6pt, inner ysep=1.6pt, font=\sffamily\bfseries\scriptsize](n){#2};%
  \hspace{4.5pt}{\bfseries\color{nb#1Txt}#3}\hfill{\scriptsize\color{nbMuted}#4}\par\vspace{4pt}}
\begin{tcbraster}[raster columns=3, raster equal height=rows, raster column skip=6pt, raster row skip=6pt,
  raster before skip=0pt, raster after skip=0pt]
\begin{tcolorbox}[nbfail, left=6pt, right=6pt]
\nbtag{4B}{Chooses an edit}{turn 12}
\nbq{12}{Let me try adding small targeted perturbations to specific attention weights to create more conservative
attention patterns.}
\begin{tcolorbox}[nbinset, after skip=0pt]
\hangindent=1.05em\hangafter=1\nbcode{12}{noise~= np.random.randn(\dots{})~*~0.01}
\end{tcolorbox}
\end{tcolorbox}
\begin{tcolorbox}[nbfail, left=6pt, right=6pt]
\nbtag{9B}{Chooses an edit}{turn 22}
\nbq{22}{The final layers (around layer 30-32) showed the largest difference between repetitive and normal samples.
I'll edit the attention out\_proj weights in the final layers with a small perturbation \dots}
\begin{tcolorbox}[nbinset, after skip=0pt]
\hangindent=1.05em\hangafter=1\nbcode{22}{perturbation~= np.random.randn(\dots{})~*~0.01}
\end{tcolorbox}
\end{tcolorbox}
\begin{tcolorbox}[nbread, left=6pt, right=6pt]
\nbtag{27B}{Chooses an edit}{turn 9}
\nbq{9}{The layer 56 hidden states show a clear difference pattern. I'll try removing the ``excessive generation''
direction from the output projections.}
\begin{tcolorbox}[nbinset, after skip=0pt]
\hangindent=1.05em\hangafter=1\nbcode{9}{diff\_vec~= length\_mean~-~finished\_mean}
\end{tcolorbox}
\end{tcolorbox}
\begin{tcolorbox}[nbfail, left=6pt, right=6pt]
\nbtag{4B}{After a failed test}{turn 34}
\nbq{34}{\dots{} My random perturbations were harmful. I need to try a different approach - perhaps making the
attention more stable or the model more deliberate.}
\end{tcolorbox}
\begin{tcolorbox}[nbfail, left=6pt, right=6pt]
\nbtag{9B}{After a failed test}{turn 33}
\nbq{33}{The MLP down\_proj edit made things worse \dots{} I need a more targeted approach. \dots{} Let me try editing
the delta net time constants (A\_log) in the final layers \dots}
\end{tcolorbox}
\begin{tcolorbox}[nbtest, left=6pt, right=6pt]
\nbtag{27B}{After a failed test}{turn 17}
\nbq{17}{The stronger edit made things worse. Let me reconsider: maybe I should enhance the ``finished'' direction rather
than remove the ``excessive'' direction, or try editing earlier layers.}
\end{tcolorbox}
\end{tcbraster}
\vspace{7pt}
\caption{\textbf{At the same step, the smaller models guess and the 27B model measures.} The sessions of the 4B, 9B
and 27B models on their own weights at the same two steps, in their own words (top: the reply that first proposes the kind of edit tested by its first paired test; bottom: its first reply, after a
paired test shows the edited model doing worse, that names what it will try next; one rule for every card,
\cref{app:quotes}). The 4B and 9B models add random noise to their weights, and when a test says the noise
hurts they guess at another change. The 27B model edits along a direction it measured in its own hidden
states, and when a test says the edit hurts it reasons about that direction and where to apply it
(\cref{fig:notebook_self}). Same task and tools at every size (\cref{sec:setup}).}
\label{fig:contrast}
\end{figure}

%% file: input/fig_notebook.tex
\begin{figure}[t]
\centering
\tcbset{nbinset/.append style={fontupper=\ttfamily\fontsize{7.8}{9.5}\color{nbRefEdge}\linespread{1.15}\selectfont\frenchspacing}}
\begin{tcbraster}[raster columns=2, raster equal height=rows, raster column skip=8pt, raster row skip=8pt,
  raster before skip=0pt, raster after skip=0pt]
\begin{tcolorbox}[nbfail]
\nbhead{1}{Symptom}{turn 1}
\nbq{1}{The starting diagnosis shows two critical failures: 29.7\% of answers hit the token limit with 100\% ending
in repetition, and even finished answers are 0\% correct.}
\end{tcolorbox}
\begin{tcolorbox}[nbread]
\nbhead{2}{It reads the copy's states}{turn 12}
\nbq{12}{The NaN values appear at identical positions across all layers, with activations exploding from
\textasciitilde 0.5 at Layer 0 to \textasciitilde 225 at Layer 63. This suggests corrupted weights in early layers
causing activation explosion.}
\end{tcolorbox}
\begin{tcolorbox}[nbfail]
\nbhead{3}{Three attempts it drops}{turns 14--19, 31}
\begin{tcolorbox}[nbinset, before skip=0pt]
\nbcode{14}{def clip\_outliers(W, factor=5.0):}
\end{tcolorbox}
\nbq{16}{The edit made things worse - cut rate increased from 31.2\% to 68.8\%. Clipping was too aggressive.}
\nbqnext{19}{The edit didn't help - cut rate slightly increased.}
\nbqnext{31}{The edit made repetition worse (81.2\% cut vs 37.5\%).}
\end{tcolorbox}
\begin{tcolorbox}[nbread]
\nbhead{4}{The outlier}{turns 23--24}
\nbq{23}{\dots{} the explosion spans many layers. I need to find which layers have corrupted attention weights causing
the residual stream to explode.}
\begin{tcolorbox}[nbinset]
\nbcode{23}{ratio = max\_val / median if median > 0 else 0}
\end{tcolorbox}
\nbq{24}{The corruption is systematic across nearly all layers. Layer 30's down\_proj has an extreme 1397x outlier.}
\end{tcolorbox}
\begin{tcolorbox}[nbedit]
\nbhead{5}{The repair}{turn 33}
\nbq{33}{\dots{} I need to try a more conservative clipping approach~- only removing the most extreme outliers.}
\begin{tcolorbox}[nbinset, after skip=0pt]
\nbcode{33}{def clip\_outliers\_gentle(W, factor=15.0):}\\
\nbcode{33}{for layer in [0, 1, 2, 3, 5, 6, 7, 8, 9, 10, 30]:}
\end{tcolorbox}
\end{tcolorbox}
\begin{tcolorbox}[nbtest]
\nbhead{6}{First test, and a revert}{turns 35, 38}
\nbq{35}{Progress! The edited model has 0\% repetitive cuts vs 100\% for original, and 25\% accuracy.}
\nbqnext{38}{The gating fixes made it worse. Let me revert to just the clipping edits \dots}
\end{tcolorbox}
\begin{tcolorbox}[nbtest]
\nbhead{7}{Full-length test}{turns 41--42}
\begin{tcolorbox}[nbinset, before skip=0pt]
\nbout{41}{samples wrong->right 16, right->wrong 0}
\end{tcolorbox}
\nbq{42}{The edits work well: 26.7\% accuracy vs 0\% \dots{}}
\end{tcolorbox}
\begin{tcolorbox}[nbedit]
\nbhead{8}{Final answer}{end of session}
\nbq{final}{Layer 30's \texttt{down\_proj.weight} had an extreme 1397\texttimes{} median outlier}
\nbqnext{final}{Repair method: Clipped values exceeding 15\texttimes{} the median absolute value to 15\texttimes{}
median.}
\end{tcolorbox}
\end{tcbraster}
\vspace{7pt}
\caption{\textbf{The repair session, in the model's words.} Each card is one step of the 27B model's session on a
damaged copy of itself, with the turns in which it wrote the quoted text. Sans-serif text is quoted verbatim from its
replies (omissions marked \dots); white insets are tool output it read or code it wrote. Card colours give the
kind of step: peach, the symptom and the attempts it drops; lavender, the model reading and diagnosing itself; pink,
paired tests of the edited copy against the damaged copy on the same problems and seeds, and the decisions it takes
from them; green, the edit it writes and the
one it submits. ``Original'' is the damaged copy and ``edited'' is that copy with the model's
current edit; steps 6 and 7 compare the two on the same problems and seeds. In the model's words, a ``cut'' is an
answer that runs out of tokens, ``NaN'' an invalid number, the ``residual stream'' the hidden state passed from layer to
layer, ``down\_proj'' the MLP output matrix, and the ``gating fixes'' its resets of the gating parameters. The third
quote of step~3 (turn 31) is about an edit made after step~4. Step~5 quotes its loop over MLP output matrices; it applies
the same rule to the linear-attention output matrices of layers 0--10.}
\label{fig:notebook}
\end{figure}

%% file: input/fig_notebook_self.tex
\begin{figure}[t]
\centering
\begin{tcbraster}[raster columns=2, raster equal height=rows, raster column skip=8pt, raster row skip=8pt,
  raster before skip=0pt, raster after skip=0pt]
\begin{tcolorbox}[nbfail]
\nbhead{1}{Symptom}{turn 1}
\nbq{1}{The diagnosis shows 80.5\% of answers hit the token limit without finishing, but 96\% of completed answers are
correct. This means my reasoning is sound but I'm too verbose or don't converge to conclusions efficiently.}
\end{tcolorbox}
\begin{tcolorbox}[nbread]
\nbhead{2}{It reads itself}{turns 3, 9}
\nbq{3}{\dots{} the model doesn't know when to conclude. I need to examine hidden states to find what internal patterns
cause this excessive generation vs.\ normal termination.}
\nbqnext{9}{The layer 56 hidden states show a clear difference pattern. I'll try removing the ``excessive generation''
direction from the output projections.}
\end{tcolorbox}
\begin{tcolorbox}[nbfail]
\nbhead{3}{Late-layer edits it drops}{turns 21--27}
\nbq{21}{Both approaches made things worse. The mean hidden state difference might not capture what matters.}
\begin{tcolorbox}[nbinset]
\nbout{26}{samples wrong->right 1, right->wrong 6}
\end{tcolorbox}
\nbq{27}{The edits keep making things worse. Let me reconsider \dots}
\end{tcolorbox}
\begin{tcolorbox}[nbread]
\nbhead{4}{It looks earlier}{turns 30--32}
\nbq{30}{I've used 8 passes and all edits made things worse. Let me reconsider: maybe the verbosity is established
early \dots}
\nbqnext{32}{The differences grow with depth. Let me try a very small edit on an earlier layer (layer 8) to see if that
helps.}
\end{tcolorbox}
\begin{tcolorbox}[nbedit]
\nbhead{5}{The edit}{turn 50}
\begin{tcolorbox}[nbinset, before skip=0pt, after skip=0pt]
\nbcode{50}{finished\_8 = hidden[:20, layer\_8\_idx, :]}\\
\nbcode{50}{diff\_vec = length\_mean - finished\_mean}\\
\nbcode{50}{u = diff\_vec / np.linalg.norm(diff\_vec)}\\
\nbcode{50}{alpha = 0.05}\\
\nbcode{50}{out\_proj\_edited = remove\_direction(\dots{})}
\end{tcolorbox}
\end{tcolorbox}
\begin{tcolorbox}[nbtest]
\nbhead{6}{Its tests decide}{turns 52--55}
\begin{tcolorbox}[nbinset, before skip=0pt]
\nbout{52}{samples wrong->right 4, right->wrong 2}\\
\nbout{53}{samples wrong->right 6, right->wrong 5}\\
\nbout{54}{samples wrong->right 14, right->wrong 4}
\end{tcolorbox}
\nbq{final}{\dots{} removing the ``excessive reasoning'' direction from this layer helped the model conclude
earlier.}
\end{tcolorbox}
\end{tcbraster}
\vspace{7pt}
\caption{\textbf{The session in which the 27B model improves itself, in its own words.} Each card is one step of its
first-protocol session on its own weights, whose edit it submits (\cref{sec:selfcase}), in the format of \cref{fig:notebook}: sans-serif
text is quoted verbatim from its replies, white insets are tool output it read or code it wrote. In step~3 and step~6,
\texttt{wrong->right} counts the answers that only its edited weights get right and \texttt{right->wrong} those that
only its unedited weights get right, in a paired test on the same problems and seeds. Step~5 is its final edit: the direction from
the average layer-8 hidden state of 20 answers that finish to that of 20 answers that run out of tokens, of which it
removes 5\% from the two output matrices of layer~8. Each of its three tests of this edit favours it.}
\label{fig:notebook_self}
\end{figure}

%% file: input/fig_found.tex
\begin{figure}[htb]
\centering
\includegraphics[width=\linewidth]{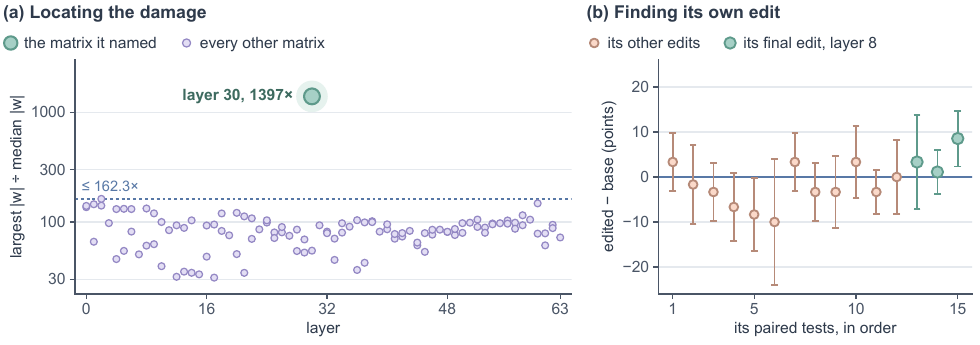}
\caption{\textbf{The 27B model finds where to act in its own weights.} (a)~In a damaged copy of itself, it compares the largest with the typical weight of every MLP and linear-attention output matrix and names the
corrupted one;
its repair clips the weights at 15 times the median magnitude (largest $|w|$ in the eight damaged rows: 0.071 original, 10.1 damaged, 0.108 after its repair). (b)~On its own weights, each of its
paired tests (edited minus base, 95\% interval): its other edits (peach) mostly make it worse, and its final edit (green), which
weakens at layer 8 the direction from its finished to its unfinished answers, is better in every test.}
\label{fig:found}
\end{figure}

%% file: tables/tab_edits.tex
\begin{table}[htb]
\centering
\small
\renewcommand{\arraystretch}{1.18}
\setlength{\tabcolsep}{5pt}
\caption{\textbf{What the 27B model changed in itself, and what changed in its accuracy.} Each edit is a short
program the model wrote; on its own weights, its first-protocol edit (second protocol: up to $+6.6$ points, \cref{fig:headline}).}
\label{tab:edits}
\begin{tabularx}{\linewidth}{@{}p{0.16\linewidth}XX@{}}
\toprule
 & \textbf{Repairing a damaged copy of itself} & \textbf{Editing its own weights} \\
\midrule
What it found &
The MLP output matrix of layer 30 holds a weight 1397 times its median; no other output matrix exceeds 162.3 times &
Its answers that run out of tokens and those that finish differ along one direction of its hidden states \\
What it changed &
Clipped every weight beyond 15 times the median magnitude, in that matrix and in 19 output matrices of layers 0--10:
0.013\% of their weights &
Removed 5\% of that direction from the two output matrices of layer 8, $W \leftarrow W - 0.05\,uu^\top W$: a change of
0.6\% and 0.1\% of their norm \\
Result &
Held-out accuracy of the copy from \textbf{0.0\%} to \textbf{33.0\%} (base model 58.5\%); the named matrix alone:
\textbf{55.0\%} &
Own task from 22.4\% to \textbf{29.6\%} (significant); held-out, in the evaluation fixed in advance: 57.2\% to
\textbf{58.3\%} (significant) \\
\bottomrule
\end{tabularx}
\end{table}

%% file: input/fig_repair_parts.tex
\begin{figure}[htb]
\centering
\includegraphics[width=\linewidth]{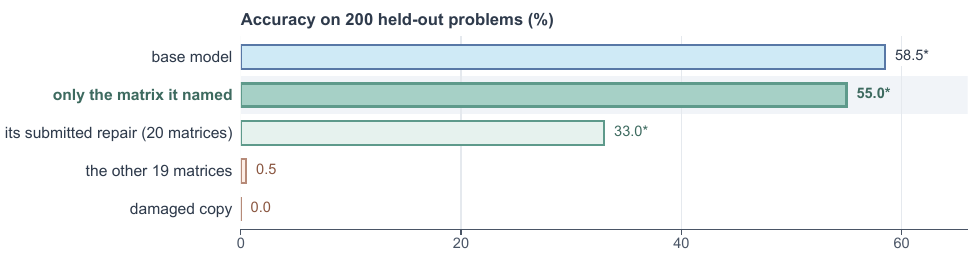}
\caption{\textbf{Its change to the one matrix it named repairs the damaged copy; its changes to the other 19 matrices do
not.} Accuracy on the same
200 held-out problems, each answered once with the same seeds: the base model, the damaged copy with only the model's
change to the matrix it named, with its whole repair or with only its changes to the other 19 matrices, and the damaged
copy itself ($^{*}$: one-sided sign test against the damaged copy, $p<0.05$). Higher is better.}
\label{fig:parts}
\end{figure}

%% file: tables/tab_main.tex
\begin{table}[htb]
\centering
\small
\setlength{\tabcolsep}{4.6pt}
\caption{\textbf{Only the 27B model improves itself and repairs a damaged copy of itself.} Accuracy (\%); $\Delta$
from unrounded values. Own task: the model's paired tests in its session, each problem counted once (4B at 16{,}384
tokens, 27B at 4{,}096: the tests of its submitted edit, on problems it also used to find it; 9B tested only on problems
its base model fails). Held-out: problems no session saw, at 16{,}384 tokens, 200 problems with one answer each; for the
27B edit of the first protocol, the evaluation fixed in advance, 490 problems with eight answers each (\cref{tab:heldout}). $^{*}$One-sided test
over problems, $p<0.05$: own task $p=0.009$ and held-out $p=0.038$ (permutation), repair $p<10^{-19}$ (sign test).}
\label{tab:main}
\begin{tabular}{lccccccccc}
\toprule
 & \multicolumn{3}{c}{Own task (its session's tests)} & \multicolumn{3}{c}{Held-out, 16{,}384 tokens} & \multicolumn{3}{c}{Damaged copy, held-out} \\
\cmidrule(lr){2-4}\cmidrule(lr){5-7}\cmidrule(lr){8-10}
Model & base & edited & $\Delta$ & base & edited & $\Delta$ & base & damaged & repaired \\
\midrule
2B & \multicolumn{3}{c}{no edit} & \multicolumn{3}{c}{no edit} & 9.0 & 0.0 & no edit \\
4B & 44.1 & 0.0 & $-$44.1 & \multicolumn{3}{c}{withdrawn} & 41.0 & 0.0 & 0.0 \\
9B & \multicolumn{3}{c}{---} & 47.5 & 0.0 & $-$47.5 & 47.5 & 0.0 & 0.0 \\
\rowcolor{srwBandTeal}\textcolor{srwOursTxt}{\textbf{27B}} & 22.4 & 29.6 & \textbf{+7.3$^{*}$} & 57.2 & 58.3 & \textbf{+1.1$^{*}$} & 58.5 & 0.0 & \textbf{33.0}$^{*}$ \\
\bottomrule
\end{tabular}
\end{table}

%% file: input/fig_visual.tex
\begin{figure}[htb]
\centering
\includegraphics[width=\linewidth]{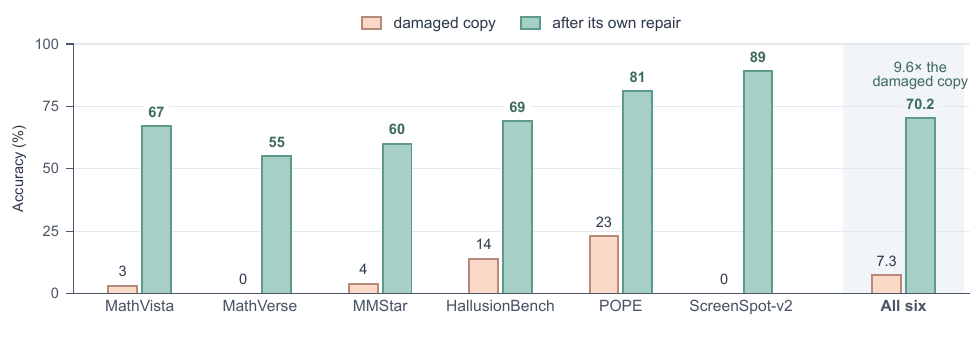}
\caption{\textbf{The repair found on text restores the damaged copy on images.} The repair was found and tested on text
math problems only; the six image benchmarks were evaluated once, by a rule fixed in advance (\cref{app:visual}).
Accuracy of the damaged copy and of the copy after the model's own repair, on the first 100 problems of each
benchmark; the repair raises every benchmark (one-sided sign test, $p<10^{-12}$ on each) and the six together,
from 7.3\% to 70.2\%, 9.6 times the accuracy of the damaged copy ($p<10^{-99}$). One answer per problem at 16{,}384 tokens with the same seed for every model;
per-benchmark numbers in \cref{tab:visual}; \cref{fig:visualbase} adds the base model. Higher is better.}
\label{fig:visual}
\end{figure}

%% file: input/fig_qual_visual.tex
\begin{figure}[t]
\centering
\includegraphics[width=\linewidth]{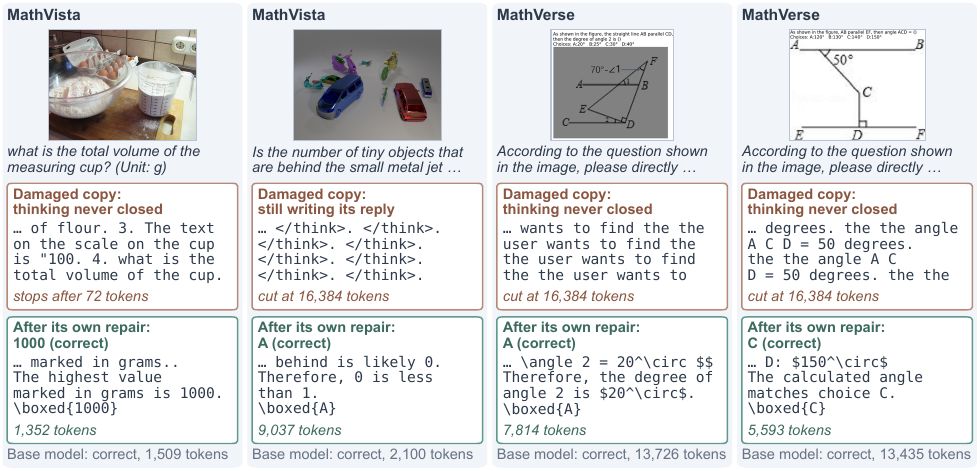}
\caption{\textbf{The same on image math problems.} The first two of MathVista and of MathVerse on which the damaged copy
is wrong and the repaired copy is right, by the same rule and drawn as in \cref{fig:qualvis2}.}
\label{fig:qualvis}
\end{figure}

%% file: input/fig_qual_cards.tex
\begin{figure}[tb]
\centering
\newcommand{\qcfs}{\fontsize{7}{8}\selectfont}
\definecolor{qcRefFrame}{HTML}{B9C3D2}%
\newcommand{\qcx}[2]{#2}%
\newcommand{\qcbad}[1]{{\color{nbFailTxt}#1}}
\newcommand{\qcnl}{\newline}
\newcommand{\qcsb}[1]{_{\mbox{\fontsize{7}{7}\selectfont\rmfamily#1}}}%
\newcommand{\qcboxed}[1]{{\setlength{\fboxsep}{0.8pt}\fcolorbox{nbOursFrame}{white}{\color{nbOursTxt}%
  \fontfamily{lmtt}\fontseries{b}\selectfont\textbackslash boxed\{#1\}}}}
\newcommand{\qcans}[1]{\unskip\nobreak\hfill\penalty50\hskip0.5em\null\nobreak\hfill
  \mbox{\bfseries Answer #1.}\parfillskip=0pt\par}
\newcommand{\qcnote}[2]{\tcblower{\sffamily\itshape\color{nb#1Txt}#2}}
\newcommand{\qctitle}[2]{{\sffamily\bfseries\color{nb#1Txt}#2}\par\vspace{1pt}}
\newcommand{\qcrow}[2]{\par{\raggedright\noindent{\fontsize{9}{11}\selectfont\sffamily\bfseries\color{nbInk}#1}\hspace{6pt}%
  {\qcfs\sffamily\color{nbMuted}#2}\par}\vspace{2.5pt}}
\tcbset{
  qccard/.style={enhanced, arc=2.5pt, boxrule=0.5pt, colframe=qcRefFrame, colback=white,
    left=2.2pt, right=2.2pt, top=2.2pt, bottom=2.2pt, boxsep=0pt},
  qchead/.style={blankest, equal height group=#1, valign=top, left=0.4pt, right=1.4pt,
    fontupper={\sffamily\qcfs\color{nbInk}\linespread{1.0}\selectfont\raggedright
      \rightskip=0pt plus 5em\relax}},%
  qcpart/.style 2 args={enhanced, arc=1.8pt, boxrule=0.6pt, colframe=nb#1Frame, colback=nb#1Light,
    left=2.5pt, right=2.5pt, top=2pt, bottom=2pt, boxsep=0pt, before skip=2.5pt, after skip=0pt,
    equal height group=#2, space to upper, valign upper=top, valign lower=bottom, middle=1.5pt,
    segmentation hidden, fontlower=\qcfs,
    fontupper=\ttfamily\qcfs\color{nbInk}\linespread{1.0}\selectfont\raggedright\frenchspacing},
}
\qcrow{(a) Repairing a damaged copy of itself}{held-out problems (the third shortened); same seed before and after its own repair}
\begin{tcbraster}[raster columns=4, raster column skip=3.5pt, raster equal height=rows]
\begin{tcolorbox}[qccard]
  \begin{tcolorbox}[qchead=qcAh]
  The largest whole number such that seven times the number is less than 100 is \qcans{14}
  \end{tcolorbox}
  \begin{tcolorbox}[qcpart={Fail}{qcAb}]
  \qctitle{Fail}{Damaged copy}
  \qcx{}{Let's think step-by-step to solve this problem. step step. The largest whole number such that
  seven times the number is \qcbad{less than .} \dots{} \qcbad{less than .} \dots}
  \qcnote{Fail}{stops after \textbf{59} tokens, no answer}
  \end{tcolorbox}
  \begin{tcolorbox}[qcpart={Ours}{qcAg}]
  \qctitle{Ours}{After its own repair}
  \qcx{}{\dots{} x < 14.28 \dots{} If x = 14, then 7 $\times$ 14 = 98. Since~98~<~100, this works.
  \dots{} \qcboxed{14}}
  \qcnote{Ours}{correct, \textbf{1{,}083} tokens}
  \end{tcolorbox}
\end{tcolorbox}
\begin{tcolorbox}[qccard]
  \begin{tcolorbox}[qchead=qcAh]
  Find the terminating decimal expansion of $11/125$. \qcans{0.088}
  \end{tcolorbox}
  \begin{tcolorbox}[qcpart={Fail}{qcAb}]
  \qctitle{Fail}{Damaged copy}
  \qcx{}{\dots{} for the fraction 11/125. The fraction is \qcbad{1/125}. \dots{} We can also write it as
  \qcbad{1/125}. We can also write it as \qcbad{1/125}. \dots}
  \qcnote{Fail}{cut at \textbf{16{,}384} tokens, no answer}
  \end{tcolorbox}
  \begin{tcolorbox}[qcpart={Ours}{qcAg}]
  \qctitle{Ours}{After its own repair}
  \qcx{}{\dots{} 88/1000 = 0.088 \dots{} We can check the result by dividing 11 by 125 \dots{}
  \qcboxed{0.088}}
  \qcnote{Ours}{correct, \textbf{1{,}746} tokens}
  \end{tcolorbox}
\end{tcolorbox}
\begin{tcolorbox}[qccard]
  \begin{tcolorbox}[qchead=qcAh]
  Daniel's theory: toaster popularity (sales) is inversely proportional to cost. 12 customers buy at \$500; how many
  at \$750?
  \qcans{8}
  \end{tcolorbox}
  \begin{tcolorbox}[qcpart={Fail}{qcAb}]
  \qctitle{Fail}{Damaged copy}
  \qcx{}{The user is asking a question about a popularity of a toaster. \dots{} The popularity is inversely
  proportional to the cost. \qcbad{. . . . . . . . . . . . .} \dots}
  \qcnote{Fail}{cut at \textbf{16{,}384} tokens, no answer}
  \end{tcolorbox}
  \begin{tcolorbox}[qcpart={Ours}{qcAg}]
  \qctitle{Ours}{After its own repair}
  \qcx{}{\dots{} According to Daniel's theory, 8 customers would buy a toaster that costs \$750.
  \qcboxed{8}}
  \qcnote{Ours}{correct, \textbf{1{,}788} tokens}
  \end{tcolorbox}
\end{tcolorbox}
\begin{tcolorbox}[qccard]
  \begin{tcolorbox}[qchead=qcAh]
  The value of $\log\qcsb{10}17$ is between the consecutive integers $a$ and $b$. Find $a+b$. \qcans{3}
  \end{tcolorbox}
  \begin{tcolorbox}[qcpart={Fail}{qcAb}]
  \qctitle{Fail}{Damaged copy}
  \qcx{}{The problem asks us to find the value of the expression log\_10 17. We are told that the value of
  \qcbad{log\_\{\_\{1\}0\} is consecutive between consecutive integers} \dots}
  \qcnote{Fail}{cut at \textbf{16{,}384} tokens, no answer}
  \end{tcolorbox}
  \begin{tcolorbox}[qcpart={Ours}{qcAg}]
  \qctitle{Ours}{After its own repair}
  \qcx{}{\dots{} 1 < log\_10 17 < 2\qcnl Thus, the consecutive integers a and b are 1 and 2. \dots{}
  \qcboxed{3}}
  \qcnote{Ours}{correct, \textbf{1{,}811} tokens}
  \end{tcolorbox}
\end{tcolorbox}
\end{tcbraster}
\par\vspace{6pt}
\qcrow{(b) Editing its own weights (first protocol)}{held-out problems (the third shortened); same seed before and after its own edit}
\begin{tcbraster}[raster columns=3, raster column skip=3.5pt, raster equal height=rows]
\begin{tcolorbox}[qccard]
  \begin{tcolorbox}[qchead=qcBh]
  If $x$ is real and positive and grows beyond all bounds, then $\log\qcsb{3}(6x-5)-\log\qcsb{3}(2x+1)$ approaches \qcans{1}
  \end{tcolorbox}
  \begin{tcolorbox}[qcpart={Fail}{qcBb}]
  \qctitle{Fail}{Base model}
  \qcx{}{\dots{} The reasoning is solid. \dots{} I will ensure the answer is 1. \dots{}
  \qcbad{I will ensure the response is correct. I will ensure the response is correct.} \dots}
  \qcnote{Fail}{cut at \textbf{16{,}384} tokens, no answer}
  \end{tcolorbox}
  \begin{tcolorbox}[qcpart={Ours}{qcBg}]
  \qctitle{Ours}{After its own edit}
  \qcx{}{\dots{} Therefore:\qcnl log\_3(3) = 1\qcnl Thus, the expression approaches 1.\qcnl
  \qcboxed{1}}
  \qcnote{Ours}{correct, \textbf{1{,}887} tokens}
  \end{tcolorbox}
\end{tcolorbox}
\begin{tcolorbox}[qccard]
  \begin{tcolorbox}[qchead=qcBh]
  Find the terminating decimal expansion of $11/125$. \qcans{0.088}
  \end{tcolorbox}
  \begin{tcolorbox}[qcpart={Fail}{qcBb}]
  \qctitle{Fail}{Base model}
  \qcx{}{\dots{} Final Answer seems to be 0.088. Double check 11/125. \dots{} So 0.088. Yes. It's correct.
  \dots{} \qcbad{I'll write the response. I'll write the response.} \dots}
  \qcnote{Fail}{cut at \textbf{16{,}384} tokens, no answer}
  \end{tcolorbox}
  \begin{tcolorbox}[qcpart={Ours}{qcBg}]
  \qctitle{Ours}{After its own edit}
  \qcx{}{\dots{} 11 $\times$ 8 = 88 \dots{} 125 $\times$ 8 = 1000 \dots{} 88/1000 = 0.088 \dots{}
  The terminating decimal expansion is 0.088. \qcboxed{0.088}}
  \qcnote{Ours}{correct, \textbf{3{,}570} tokens}
  \end{tcolorbox}
\end{tcolorbox}
\begin{tcolorbox}[qccard]
  \begin{tcolorbox}[qchead=qcBh]
  How many five-letter words begin and end with the same letter and have a vowel as the third letter? \qcans{87880}
  \end{tcolorbox}
  \begin{tcolorbox}[qcpart={Fail}{qcBb}]
  \qctitle{Fail}{Base model}
  \qcx{}{\dots{} So, 87,880 is the intended answer. \dots{} Sum: 87880. Verified. \dots{}
  \qcbad{I'll write the response. I'll write the response.} \dots}
  \qcnote{Fail}{cut at \textbf{16{,}384} tokens, no answer}
  \end{tcolorbox}
  \begin{tcolorbox}[qcpart={Ours}{qcBg}]
  \qctitle{Ours}{After its own edit}
  \qcx{}{\dots{} Total Words = 26 $\times$ 26 $\times$ 5 $\times$ 26 $\times$ 1 \dots{} 17,576 $\times$ 5 =
  87,880\qcnl Thus, there are 87,880 such words. \qcboxed{87880}}
  \qcnote{Ours}{correct, \textbf{5{,}443} tokens}
  \end{tcolorbox}
\end{tcolorbox}
\end{tcbraster}
\vspace{1.5pt}
\caption{\textbf{With the same seed, each of the 27B model's edits turns a lost answer into a short correct one.} (a)~The damaged copy loses the numbers of
the problem and breaks off or repeats itself; after its own repair, the same seed gives a correct answer (of the 66 held-out
problems the repair fixes, the four with the shortest repaired answers). (b)~The base model
reaches the result, then repeats one sentence until the 16{,}384-token limit; after its own edit, it stops and answers (of the
four problems where the base model loops in both of its answers with that seed (\cref{app:qual}) and the edited model is correct, the three with
the shortest edited answers).}
\label{fig:qual}
\end{figure}

%% file: input/fig_front_mathvista.tex
\begin{figure}[tb]
\centering
\includegraphics[width=\linewidth]{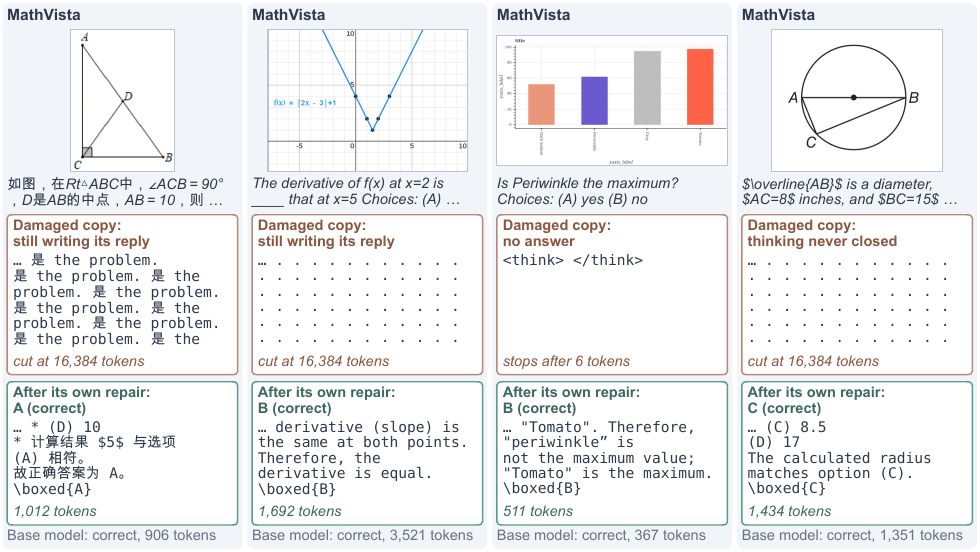}
\caption{\textbf{More image cases: MathVista (cases 3--6).} The next four problems on which the damaged copy is wrong and the repaired copy is right, after those of \cref{fig:qualvis}, chosen and drawn as in \cref{fig:qualvis2}.}
\label{fig:front_mathvista}
\end{figure}

%% file: input/fig_flow.tex
\begin{figure}[tb]
\centering
\begin{minipage}[c]{3.18in}
\includegraphics{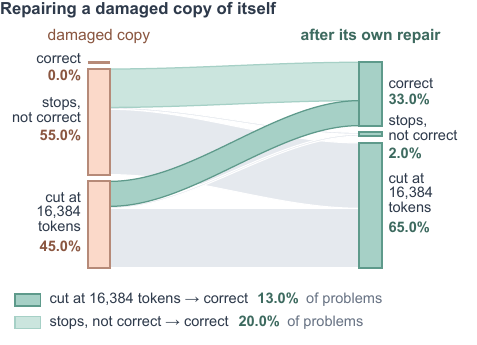}
\end{minipage}\hfill
\begin{minipage}[c]{\dimexpr\linewidth-3.18in-1.5em\relax}
\caption{\textbf{The model's own repair gives its damaged copy back correct answers.} Each band joins the state of one answer before and after the repair (correct; stops, not correct; cut at 16{,}384 tokens), with width proportional to the number of answers: the damaged copy, whose answers that stop all stop without an answer, and the copy after the model's own repair, on 200 held-out problems. Same seeds on both sides.}
\label{fig:flow}
\end{minipage}
\end{figure}

%% file: input/fig_front_mathverse.tex
\begin{figure}[tb]
\centering
\includegraphics[width=\linewidth]{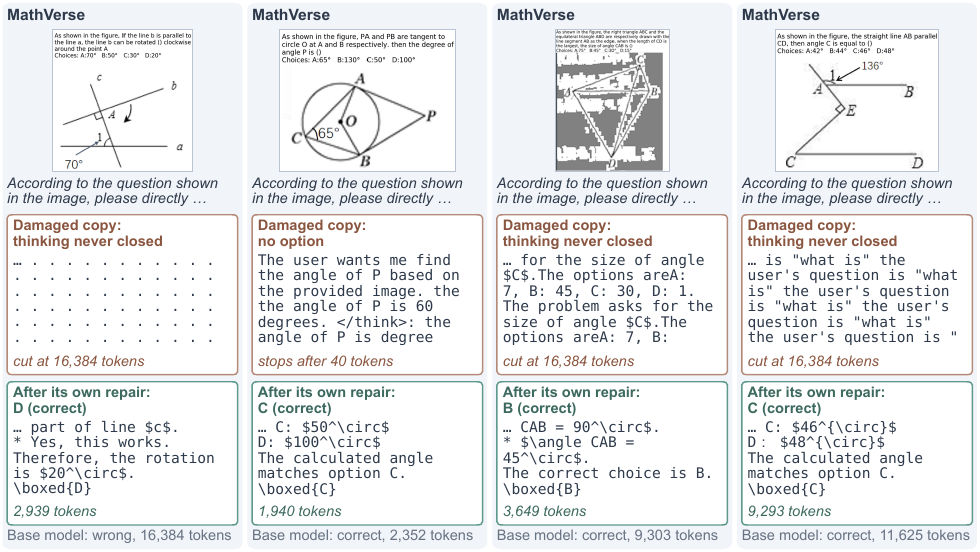}
\caption{\textbf{More image cases: MathVerse (cases 3--6).} The next four problems on which the damaged copy is wrong and the repaired copy is right, after those of \cref{fig:qualvis}, chosen and drawn as in \cref{fig:qualvis2}.}
\label{fig:front_mathverse}
\end{figure}

%% file: input/020related.tex
\section{Related Work}
\label{sec:related}

\noindent\textbf{Self-improving language models have rewritten their prompts and agent code or left the change to
their weights to an optimizer; weights have been rewritten by reasoning only when a far stronger model edited a smaller
one} (\cref{tab:compare}).

\input{tables/tab_compare}

The closest work is a blog post by Takehi et al.~\cite{takehi2026weightspace}, in which a frontier model traces
a small reasoning model's loops to one direction and removes it from two of its layers, with no training; it names a
model that edits its own weights as future work, which we take up.
Model editing~\cite{meng2022rome,meng2023memit}, task arithmetic~\cite{ilharco2023task}, activation
steering~\cite{turner2023actadd,zou2023repe,li2023iti} and direction ablation~\cite{arditi2024refusal} change a model
along directions that a researcher chooses; several of the 27B model's edits of its own weights are of the last kind, but it makes every choice
itself, from its own failing answers and hidden states.
Self-improvement has used self-written feedback at inference~\cite{madaan2023selfrefine,shinn2023reflexion},
rewritten prompts~\cite{zhou2023ape,yang2024opro} or agent code~\cite{yin2024godelagent,zhang2025dgm}, or trained on
the model's own outputs~\cite{zelikman2022star,yuan2024selfrewarding,zweiger2025seal}, where even self-written edits reach the weights
through gradient descent; the G\"odel
machine~\cite{schmidhuber2007godel} rewrites itself once a proof shows the rewrite useful, where our model tests each
edit in paired tests. Fault-repair frameworks restore corrupted weights with procedures engineers design for one kind
of fault~\cite{tahmasivand2025lmfix,karamat2025bitflipscope}; our model is given none. Reasoning models often spend far more tokens than a
problem needs~\cite{chen2024overthinking}, and decoding-time controls end or extend thinking~\cite{muennighoff2025s1}; the
27B model instead looks for the cause of its loops inside itself (\cref{fig:loopstates}). The full discussion is in \cref{app:related}.

%% file: tables/tab_compare.tex
\begin{table}[!ht]
\centering
\small
\setlength{\tabcolsep}{5pt}
\caption{\textbf{Routes to improving an LLM, and what each leaves out.} The first three rows are routes of
self-improvement and the next three change the weights for the model; \pmark\ marks a property that holds only for some
of the cited works or only in part.}
\label{tab:compare}
\begin{tabular}{@{}lccccc@{}}
\toprule
Route & \makecell{Changes\\weights} & \makecell{No\\training} & \makecell{Chosen by the\\model itself} &
\makecell{Looks inside\\itself} & \makecell{No extra cost\\per input} \\
\midrule
Prompt and agent-code rewriting~\cite{zhou2023ape,yang2024opro,yin2024godelagent,zhang2025dgm} & \xmark & \cmark & \pmark & \xmark & \pmark \\
Training on its own outputs~\cite{zelikman2022star,yuan2024selfrewarding,zweiger2025seal} & \cmark & \xmark & \xmark & \xmark & \cmark \\
Inference-time self-refinement~\cite{madaan2023selfrefine,shinn2023reflexion} & \xmark & \cmark & \cmark & \xmark & \xmark \\
Model editing~\cite{meng2022rome,meng2023memit} & \cmark & \xmark & \xmark & \xmark & \cmark \\
Direction ablation by researchers~\cite{arditi2024refusal} & \cmark & \cmark & \xmark & \xmark & \cmark \\
Weight editing by a stronger model~\cite{takehi2026weightspace} & \cmark & \cmark & \xmark & \xmark & \cmark \\
\midrule
\multicolumn{1}{@{}>{\columncolor{srwBandTeal}[0pt][\tabcolsep]}l}{\textbf{Self-rewriting (ours)}} & \cellcolor{srwBandTeal}\cmark & \cellcolor{srwBandTeal}\cmark & \cellcolor{srwBandTeal}\cmark & \cellcolor{srwBandTeal}\cmark & \multicolumn{1}{>{\columncolor{srwBandTeal}[\tabcolsep][0pt]}c@{}}{\cmark} \\
\bottomrule
\end{tabular}
\end{table}

%% file: input/050conclusion.tex
\section{Conclusion}
\label{sec:conclusion}

\noindent\textbf{A 27B open language model can rewrite its own weights to make itself better, and the ability to do so
emerges with scale.} With no gradients, no training and no infrastructure beyond running itself, the 27B model reads
its hidden states and weights, finds where it fails, tests its edits against its unedited self and keeps only what its
tests support. Its own edit makes it 6.6 points more accurate on 490 problems it never saw, in a paired test fixed in advance,
and the edit of another session is also significantly better. On a damaged copy of itself, it found the corrupted matrix and repaired
it with no training and without reading its undamaged weights, and the repair, found on text problems alone, also restored the copy on images, to 83\% of the base model's
accuracy. The 2B model makes no edit and the
4B and 9B models break themselves. To our knowledge this is the first evidence, in one model family, that the ability
of a model to diagnose and rewrite its own weights grows with its size, and every step it takes can be read.

\noindent\textbf{This points to a new way of improving language models: by talking to them.} Today a better model comes
from a training run, and from those who own the pipeline and the compute it takes. Training and inference need not be separate stages: as a human learns while working, a model can rewrite itself between
tasks. They were kept apart because training needs gradients, optimizer states and a pipeline of its own; rewriting needs
only what serving the model needs, and unlike a memory kept in the context, what the model learns adds nothing to the
cost of its later answers. A model that rewrites the weights
behind its own failure in one conversation opens another way: anyone who can tell a model what goes wrong may have it
fix itself, and the improvement stays in its weights. Each change is code the model wrote, over tensors it names, so it can be read, compared and
taken back like a commit, and the model can check it on problems it never used before keeping it: in the first eight
sessions that do, it withdrew five times, and the model in use only got better, ending 6.6 points more accurate than it started. Learning at
inference needs such cheap checks and rollbacks; with them, a deployed model could keep improving where it fails,
without a training pipeline, and a model could find and repair a fault in a copy of its own weights, as the 27B model
repaired its damaged copy. Embodied agents may gain the most: a robot meets a world unlike the one it was trained on,
and rather than stay frozen it could act with its weights fixed, gather what went wrong and rewrite itself between
episodes, much as sleep consolidates what a day has taught, or have a larger model rewrite its small on-board model. Because the ability emerges with scale, stronger models may be
better self-improvers still, and an improved model can be set to work on itself again, a path toward recursive
self-improvement. That path should be taken with people in the loop: every model a model makes of itself evaluated,
for safety as well as for the task, and confirmed by a person before it is used.

%% file: input/appendix.tex
\input{input/app_discussion}
\clearpage %
\input{input/app_related}
\clearpage %
\input{input/app_pseudocode}
\clearpage %
\input{input/app_details}
\clearpage %
\input{input/app_results}

%% file: input/app_discussion.tex
\section{Discussion}
\label{app:discussion}

\paragraph{Why we also test on a damaged copy.}
A session on a natural failure, such as the 27B model's sessions on its own undamaged weights, asks two questions at
once: can the model rewrite its weights, and does the failure have a cause in the weights that a rewrite can remove? When an edit does not help, the two cannot be told apart.
Damage that we place ourselves answers the second question in advance. We know where it is and what it is, and an edit
that undoes it exists, so a session that does not repair it has failed at rewriting, not at a task without a solution.
 Three choices keep the test strict. The
model is told only that some values were corrupted, as a memory or hardware fault might do, and not which values,
where, or how. It reasons with the undamaged base model, so the damage does not limit its thinking, while every
measurement its tools return is of the damaged copy. And it is given no clean reference: the functions of its Python
session return only the damaged copy, and none of its code runs opened the base model's weights or the damage files on
disk, although file access was not blocked (\cref{sec:repair}). 

\paragraph{What scale adds.}
In their sessions of the first protocol, the four models received the same task, tools and rules, including a helper
that removes part of a direction from a matrix, and started from the same kind of failure, answers that run out of
tokens. What separates them is the use of evidence about oneself. The 2B model never asked for a forward pass of
itself, a run of its own weights on problems or texts of its choice, and almost all of its code failed. The 4B model
ran itself many times but did not act on what it saw until the very end: its paired tests against its base model
showed each of its edits breaking it, and
it kept adding edits, withdrawing them all only at the end of its session, with one forward pass left. The 9B model read its hidden states once
but then changed weights by trial, tested only on problems it already failed, where no change could show a loss of accuracy, and
submitted an edit that broke it. The 27B model let each result set its next
step. It read its failing answers before it looked at its weights, read its hidden states before it edited, revised
the edit when a broader test did not support it, and judged from its tests whether to keep it: in its first session it
withdrew its first edit, and in its second, at 4{,}096 tokens, it submitted the edit its tests favoured, which makes it significantly more
accurate on held-out problems. What scale adds, in these sessions, is
the experiment around the edit rather than the edit itself.

\paragraph{Future directions.}
\textit{More measurement per session.} In its second session of the first protocol, with answers capped at 4{,}096 tokens, the 27B model used
its whole budget of passes and estimated its direction from 40 answers (\cref{app:casedetail}, \cref{fig:search}). More passes would let it test such directions
on more problems. \textit{Further rounds.} The self-edited model can run the next session on itself and study its own
edit. \textit{Graded damage.} The damage we place can vary in place, kind and strength, which turns the repair into a
measurement: how often a model finds the damage, how much of the lost accuracy it recovers, and how both change with
scale. \textit{Beyond
math.} The session needs only a check of success. With feedback from an environment, as in tasks where a model acts
and sees the result, a model that adjusts its own weights from what it observes would keep learning as it works,
without a training run.

%% file: input/app_related.tex
\section{Extended Related Work}
\label{app:related}

The full version of \cref{sec:related}.

\paragraph{Language models that edit network weights.}
The work closest to ours is a recent blog post by Takehi et al.~\cite{takehi2026weightspace}, in which one of the most
capable available models reads and writes the weights of other networks. Applied to a small reasoning model that loops in its thinking, it
traces the failure to one direction and removes it from the MLP weights of two layers, with forward passes of the
small model as its only tests and no training. There the
editor is a much stronger model than the one it edits, and a model that edits its own weights is named only as a
direction for future work. We take up that direction: in our sessions the editor and the edited network are the same
model, and no stronger model takes part. We find that a 27B open model can rewrite its own weights to make itself more accurate, and repair a damaged copy of them.

\paragraph{Hand-designed weight and activation edits.}
Model editing optimizes a target hidden vector by gradient descent and then writes it into chosen MLP layers with a
closed-form update~\cite{meng2022rome,meng2023memit}, and task
arithmetic adds or subtracts the weight change of a fine-tuning run~\cite{ilharco2023task}. Activation steering and
representation engineering shift hidden states along a direction found by contrasting two sets of
inputs~\cite{turner2023actadd,zou2023repe,li2023iti}, and projecting such a direction out of the weights makes the
change permanent~\cite{arditi2024refusal}. The edits the 27B model made of itself in its second session of the first protocol are of this last kind:
in matrices through which a layer writes its output, all but one weaken a direction that separates its hidden states
in looping or unfinished answers from those in finished ones, and one strengthens it (\cref{fig:search}). In all of these methods a researcher decides which behaviour to change, which inputs to
contrast, which layers to edit and how strongly. In our sessions the model makes each of these choices itself, from
its own failing answers and its own hidden states, and keeps or drops an edit by testing it against its base model in
paired tests.

\paragraph{Self-improvement.}
Language models improve their answers with feedback they write themselves~\cite{madaan2023selfrefine,shinn2023reflexion},
a cost that returns with every new input. They rewrite the prompts~\cite{zhou2023ape,yang2024opro} or the agent
code~\cite{yin2024godelagent,zhang2025dgm} they run with, which leaves their weights as they were. Or they change their
weights by training on what they produce: their own correct rationales~\cite{zelikman2022star}, preferences they judge
themselves~\cite{yuan2024selfrewarding}, or fine-tuning data and update settings they write for themselves, as in
SEAL~\cite{zweiger2025seal}. SEAL calls these generated data and settings self-edits, but the weights are still
changed by gradient descent on them, and reinforcement learning first trains the model to write useful ones. The G\"odel machine~\cite{schmidhuber2007godel} describes a system that rewrites
any part of itself once it has proven the rewrite useful. Self-rewriting changes the weights directly, by edits
the model reasons out and writes in code, with no optimizer, no reward model and no stronger teacher; in place of a
proof, the model tests each edit against its base model in paired tests, the two answering the same problems with the
same seeds.

\paragraph{Repairing corrupted weights.}
Memory and hardware faults such as bit flips can silently corrupt the weights of a deployed model. Dedicated
frameworks detect and localize such faults, with test inputs, checksums, comparison against a clean copy or
sensitivity profiling, and then restore the affected parameters~\cite{tahmasivand2025lmfix,karamat2025bitflipscope}.
Engineers design these procedures for one kind of fault. In our repair experiment the model is given no such procedure: told only that some of its values are corrupted, as by a memory or hardware fault, it decides by its own analysis
how to search, which matrix is damaged and how to repair it (\cref{tab:main,fig:found}).

\paragraph{Failure modes of thinking models.}
Reasoning models often spend far more tokens than a problem needs~\cite{chen2024overthinking}, and small ones can fall
into repetition that never reaches an answer~\cite{takehi2026weightspace}. Decoding-time controls set the length of
thinking, ending it at a budget or extending it~\cite{muennighoff2025s1}; they act on the output and leave the network
unchanged. The models we study lose many answers to the token limit at every size, some of them repeating the same
sentence after the answer has been reached (\cref{fig:qual}b). Instead of acting on the output, the 27B model looks for
the cause inside itself and finds, in its own hidden states, a loop that repeats at a fixed period
(\cref{fig:loopstates}).

%% file: input/app_pseudocode.tex
\section{Pseudocode}
\label{app:pseudocode}

\input{input/alg_session}

\Cref{alg:harness} is the session loop as our framework runs it, with the details that \cref{alg:session} leaves out:
how a reply is parsed, what each observation contains, how the conversation restarts and how the session ends.
\Cref{alg:paired} is the paired test, the \textsc{generate} pass on both weights with which the model compares the
weights with and without its current edit, and \cref{alg:kernel} is the Python session through which every edit is
written. The notation follows
\cref{sec:method}: the editor $M_\phi$ edits the weights $\theta$, and $\theta\oplus\Delta$ replaces the tensors named
in $\Delta$. A forward pass is one request that runs the weights: a \textsc{generate} pass answers practice problems,
and a \textsc{read} pass is a hidden-state read. Names in \texttt{typewriter} are those the model sees.

\begin{algorithm}[tbp]
\caption{The session loop of our framework. $B{=}20$, $C{=}200$ and $T_{\max}{=}160$ in every session; the time limit
$H$ and the number $N_0$ of problems in the starting diagnosis are in \cref{tab:settings}.}
\label{alg:harness}
\small
\begin{algorithmic}[1]
\Require editor $M_\phi$; edited weights $\theta$; practice problems $\mathcal{P}$; answer settings $\pi$ with
$L{=}16{,}384$ tokens ($4{,}096$ in the 27B model's second session); budgets $B$, $C$, $T_{\max}$, $H$
\If{the 2B model's session on itself or the 27B model's first session}
  \State start the clock
\EndIf
\State $c_0\gets$ one answer of $M_\theta$ under $\pi$ to each of the first $N_0$ problems of $\mathcal{P}$ by
identifier \Comment{not counted}
\If{the clock has not started}
  \State start the clock
\EndIf
\State $h\gets(\textsc{Task}(c_0))$;\; $\Delta\gets\emptyset$;\; $b,c,t\gets 0$;\; $\mathit{final}\gets$ false
\Comment{task text: \cref{app:prompt}}
\Loop
  \If{$t\ge T_{\max}$ \textbf{or} time used $>H$}
    \State append ``time is up: reply now with your final answer (no code)'' to $h$;\; $\mathit{final}\gets$ true
  \EndIf
  \State $t\gets t+1$;\; $a_t\sim M_\phi(\cdot\mid h)$ \Comment{thinking on; at most 12{,}000 tokens}
  \State $K\gets$ the distinct \texttt{python} and \texttt{forward} blocks of $a_t$
  \If{$\mathit{final}$ \textbf{or} ($K=\emptyset$ \textbf{and} $b=B$)}
    \State run each \texttt{python} block of $K$ once with \textsc{RunPython};\; \textbf{break}
  \EndIf
  \If{$|K|\ne 1$}
    \State $o\gets$ ``reply with exactly one block''
  \ElsIf{$K$ is \texttt{python}}
    \State $o\gets\textsc{RunPython}(K)$ and $c\gets c+1$ if $c<C$, else $o\gets$ ``no code runs left''
  \ElsIf{$b=B$}
    \State $o\gets$ ``no forward passes left: reply with your final answer''
  \ElsIf{$K$ is a valid \textsc{generate} or \textsc{read} request}
    \State $o\gets\textsc{Forward}(K;\,\theta,\,\theta\oplus\Delta)$;\; $b\gets b+1$
    \Comment{both weights: \cref{alg:paired}}
  \Else
    \State $o\gets$ the reason the request failed \Comment{not counted}
  \EndIf
  \State append ``forward passes used $b$ of $B$; turn $t$; time used'' to $o$;\; $h\gets h\circ(a_t,o)$
  \If{$|h|>90$k tokens}
    \State $h\gets(\textsc{Task}(c_0)$, \texttt{notes.md}, $b$, $c$, current edit, log of passes, $o)$
    \Comment{new conversation}
  \EndIf
\EndLoop
\State \Return $\theta\oplus\Delta$ \Comment{$\Delta$: the tensors in \texttt{WORK/edited}}
\end{algorithmic}
\end{algorithm}

When our framework itself fails, the session resumes from the saved conversation: the variables of the Python session
are lost, the files in \texttt{WORK} are kept, the model is told what happened, and the downtime is added back to its
time limit. In the 27B model's first session the code blocks of the final reply were not run; that reply contains the code
that withdraws its first edit (\cref{app:firstedit}).

A \textsc{read} request, a hidden-state read, runs $\theta$ or $\theta\oplus\Delta$ over up to 256 texts, each cut at
8{,}192 tokens, without sampling: texts the model writes, or answers from the starting diagnosis or from an earlier
pass. It returns the hidden states after chosen layers, the MLP activations or the token log-probabilities, averaged
over the positions of the answer or at every $k$-th position, the last $k$ positions or all of them.

\begin{algorithm}[tbp]
\caption{The paired test: a \textsc{generate} pass with \texttt{"weights":\ "both"}.}
\label{alg:paired}
\small
\begin{algorithmic}[1]
\Require problems $x_1,\dots,x_m\in\mathcal{P}$ with reference answers $y^\star_i$; $n$ samples per problem, $mn\le 256$;
token limit $L'$ with $16\le L'\le L$; weights $\theta$ and current edit $\Delta$
\State load $\theta\oplus\Delta$ into the generation engine
\State $s_i\gets\textsc{Hash}(\text{identifier of }x_i)\bmod 2^{31}$ for $i=1,\dots,m$
\Comment{same seed for both weights, in every pass}
\For{$e\in\{0,1\}$, with $\theta_0=\theta$ and $\theta_1=\theta\oplus\Delta$}
  \For{$i=1,\dots,m$}
    \State draw $y^e_{i1},\dots,y^e_{in}\sim M_{\theta_e}(\cdot\mid x_i;\pi)$ with seed $s_i$ and at most $L'$ new tokens
    \For{$j=1,\dots,n$}
      \State $g^e_{ij}\gets 1$ if the thinking of $y^e_{ij}$ closes and its last \texttt{\textbackslash boxed\{\}}
      matches $y^\star_i$, else $0$ \Comment{Math-Verify}
      \State $\ell^e_{ij}\gets 1$ if $y^e_{ij}$ is cut at $L'$ tokens, else $0$
      \State $r^e_{ij}\gets 1$ if $\ell^e_{ij}=1$ and the last 6{,}000 characters of $y^e_{ij}$ compress to under
      15\% of their size, else $0$
    \EndFor
  \EndFor
\EndFor
\State $d_i\gets\frac1n\sum_j\big(g^1_{ij}-g^0_{ij}\big)$;\; $\hat\delta\gets\frac1m\sum_i d_i$;\;
$\mathrm{SE}(\hat\delta)$ as in \cref{eq:paired} \Comment{the same for $\ell$}
\State $n_{\uparrow}\gets\big|\{(i,j):g^0_{ij}=0,\,g^1_{ij}=1\}\big|$;\;
$n_{\downarrow}\gets\big|\{(i,j):g^0_{ij}=1,\,g^1_{ij}=0\}\big|$
\State \Return for $\theta$ and for $\theta\oplus\Delta$: accuracy, share cut at $L'$, share of the cut answers with $r=1$, mean length;
$\hat\delta$ and $\mathrm{SE}(\hat\delta)$ for accuracy and for the cut share; $n_{\uparrow}$ and $n_{\downarrow}$;
all answers in \texttt{WORK/gen\_<k>.jsonl}
\end{algorithmic}
\end{algorithm}

Seeds depend only on the problem, so repeated passes on a problem reuse them, although generation is not exactly
reproducible and a repeated pass can still draw different answers. Every change to the weights goes through \textsc{SaveEdit}
(\cref{alg:kernel}): the Python session checks the text of each code block before running it, and every tensor before
storing it.

\begin{algorithm}[tbp]
\caption{The Python session and how an edit is saved.}
\label{alg:kernel}
\small
\begin{algorithmic}[1]
\Function{RunPython}{code} \Comment{on a CPU; variables persist across runs; gradients disabled}
  \If{code uses gradients, optimizers, model loaders, inference engines, subprocesses or the network}
    \State \Return ``refused''
  \EndIf
  \State \Return the output of code, stopped after 600\,s and cut at 10{,}000 characters
\EndFunction
\Function{SaveEdit}{$k$, $W'$} \Comment{\texttt{save\_edit(name, tensor)}}
  \If{$k$ is the final norm, the embedding or the output layer}
    \State refuse
  \ElsIf{$W'$ differs from $W_k$ in shape, has a non-finite entry, or equals $cW_k$ for a constant $c$}
    \State refuse
  \EndIf
  \State $\Delta\gets\Delta\cup\{(k,W')\}$, with $W'$ stored in \texttt{WORK/edited} in bf16, the 16-bit format of the weights
\EndFunction
\end{algorithmic}
\end{algorithm}

%% file: input/alg_session.tex
\begin{algorithm}[tbp]
\caption{A session. The editor $M_\phi$ edits weights $\theta$ ($\tilde\theta$ for the damaged copy); $\phi=\theta$ when the model edits itself.}
\label{alg:session}
\small
\vspace{2pt}\noindent At turn $t$ the editor (thinking on, $\phi$ fixed) replies $a_t$ to the conversation $h_{t-1}$ and
observes $o_t$; $\Delta_t$ is the current edit, whose tensors \textsc{code} may write or drop, and $b$, $c$ count the
forward passes and code runs used.
\begin{algorithmic}[1]
\Require editor $M_\phi$, weights $\theta$ to edit, practice set $\mathcal{P}$, answer settings $\pi$, budgets $B$, $C$, $T_{\max}$, time limit
\State $h_0\gets(\text{task text},\ \text{starting diagnosis of } M_\theta \text{ on the first } N_0 \text{ problems of } \mathcal{P})$;\; $\Delta_0\gets\emptyset$;\; $b\gets0$;\; $c\gets0$
\For{$t=1,2,\dots$}
  \State $a_t\sim M_\phi(\cdot\mid h_{t-1})$;\; $\Delta_t\gets\Delta_{t-1}$
  \State \algorithmicif\ $a_t$ is the final reply \algorithmicthen\ \Return $\hat\Delta\gets\Delta_t$ \Comment{$b=B$, $t>T_{\max}$ or time is up}
  \State \algorithmicif\ $a_t$ is \textsc{code} and $c<C$ \algorithmicthen\ $o_t\gets\textsc{RunPython}(a_t)$;\; $c\gets c+1$
  \State \algorithmicelse\ \algorithmicif\ $a_t$ is a valid \textsc{generate} or \textsc{read} and $b<B$ \algorithmicthen\ $o_t\gets\textsc{Forward}(a_t;\,\theta,\,\theta\oplus\Delta_t)$;\; $b\gets b+1$
  \State \algorithmicelse\ $o_t\gets$ error message \Comment{not counted}
  \State $h_t\gets h_{t-1}\circ(a_t,o_t)$, reset to (task text, notes, log of passes) beyond 90k tokens
\EndFor
\end{algorithmic}
\end{algorithm}

%% file: input/app_details.tex
\definecolor{srwInk}{HTML}{2E3A4B}       %
\definecolor{srwPanel}{HTML}{F6F8FB}     %
\definecolor{srwRefEdge}{HTML}{5878A6}   %
\definecolor{srwFail}{HTML}{FBD9C9}      %
\tcbset{srwbase/.style={enhanced, arc=2mm, boxrule=0.8pt, colback=white, coltitle=srwInk,
  fonttitle=\sffamily\bfseries\small, toptitle=1.1mm, bottomtitle=1.1mm, left=3.2mm, right=3.2mm, top=1.8mm,
  bottom=1.8mm, before skip=5pt, after skip=6pt, fontupper=\footnotesize\color{srwInk}, parbox=false,
  before upper={\raggedright\frenchspacing\setlength{\parskip}{0.3em}}}}
\newtcolorbox{srwprompt}[3]{srwbase, unbreakable, colframe=#1, colbacktitle=#2, title={#3}}   %
\newtcolorbox{srwnote}[3]{srwbase, unbreakable, colframe=#1, colbacktitle=#2, title={#3}}
\newcommand{\promptph}[1]{\textrm{$\langle$\textit{#1}$\rangle$}}

\section{Experimental Details}
\label{app:details}

\subsection{Settings of Every Session}
\label{app:settings}

\Cref{tab:settings,tab:shared} list what differs between the sessions and what they share. Every session starts from the same
task text (\cref{app:prompt}); the repair sessions append a note on the damage.

\input{tables/tab_settings}

\begin{table}[htbp]
\centering
\small
\caption{\textbf{Settings shared by all sessions of the first protocol;} \cref{app:protocol2} lists what the second protocol changes.}
\label{tab:shared}
\renewcommand{\arraystretch}{1.06}
\begin{tabularx}{\linewidth}{@{}p{0.165\linewidth}X@{}}
\toprule
Practice set & 3{,}000 problems from DeepScaleR~\cite{deepscaler2025}, after removing every problem that overlaps
MATH-500, AIME 2024/2025, HMMT February 2025, AMC 2023, OlympiadBench or GPQA-Diamond (\cref{sec:setup}). \\
Answer settings & Used for the starting diagnosis and every generation pass. Thinking on: the model first reasons
in a passage that it closes with an end marker, then gives its final answer; the question followed by ``Please reason
step by step, and put your final answer within \texttt{\textbackslash boxed\{\}}.''; sampling at temperature 0.6,
each next token drawn from the 20 most likely tokens and, among those, from the smallest set that holds 95\% of the
probability; a budget of 16{,}384 new tokens (4{,}096 in the 27B model's second session on itself), or fewer if a pass
asks (at least 16). \\
Grading & The last \texttt{\textbackslash boxed\{\}} after the thinking, checked against the reference with
Math-Verify; no credit if the thinking is still open when the budget runs out. \\
Looping answer & An answer cut at the token limit whose last 6{,}000 characters compress (zlib) to under 15\% of their
size. \\
Seeds & Fixed by the problem's identifier: the same with and without the edit, and in every pass of a session. \\
Generation pass & Up to 512 answers, or 256 each with and without the edit when both are run. \\
Hidden-state read & Up to 256 texts chosen by the model, each cut at 8{,}192 tokens; the weights are run over them
and return the hidden states after chosen layers (the vector of numbers that a layer passes to the next, one per
token), MLP activations (values inside a layer's MLP block) or token log-probabilities. \\
Code runs & A Python session on a CPU that keeps its state between runs, with gradients disabled; up to 600\,s per
run; output cut at 10{,}000 characters. \\
Edits & Any tensor, replaced by one of the same shape with finite values, except the input embedding (the table of
token vectors), the final normalisation and the output layer (which turn the last hidden state into next-token
probabilities), and except a whole tensor times a constant; stored in bf16, the 16-bit number format of the model's
weights. \\
Editor replies & Thinking on; the same sampling; up to 12{,}000 new tokens per reply. When
the context exceeds 90k tokens, the conversation restarts from the task, the model's notes and its log of passes. \\
Serving & Generation with vLLM~\cite{kwon2023vllm}. The editor runs in the same engine, which is reset to the editor's
weights before each reply. \\
\bottomrule
\end{tabularx}
\end{table}

\clearpage
\subsection{The Damage}
\label{app:damage}

We damage one matrix of the 27B model: the MLP output matrix (\texttt{mlp.down\_proj}) of layer 30 of 64, through which
the MLP block of that layer, a two-layer network applied to each token on its own, adds its result to the hidden
state. Rows 3068 to
3075, a block of eight consecutive rows at a randomly drawn position, are replaced with independent draws from
$\mathcal{N}\big(0,(200\sigma)^2\big)$, where $\sigma=0.0111$ is the standard deviation of the matrix's entries. Each
row holds the weights that produce one coordinate of the MLP's output, so the damage corrupts eight of the coordinates
that this MLP adds to the model's hidden state. Every other weight is unchanged. The damaged matrix is stored in bf16,
like the rest of the model, and our framework loads it as the weights the session starts from; the record of which
rows were damaged, and how, is kept in a separate file that the session is not given.

We chose the strength before any repair session, on 24 easy problems from the 27B model's starting diagnosis with four
answers each at 4{,}096 tokens (\cref{tab:damage}). Of the three strengths we tried, only the strongest breaks the
model: weaker corruptions of the same rows leave it about as accurate as the base model. In its own starting diagnosis
at the full budget,
the damaged copy answers none of its 64 problems correctly; the answers that run out of tokens all end in loops, and
those that finish are broken text, most of them short.

\begin{table}[htb]
\centering
\small
\caption{\textbf{Choosing the strength of the damage.} Share of correct answers (\%) of the 27B model on 24 easy problems
from its starting diagnosis, four answers each at 4{,}096 tokens, as the base model and with rows 3068 to 3075 of layer
30's MLP output matrix overwritten at three strengths ($\sigma$: the standard deviation of the matrix's entries). The
repair experiment uses $200\sigma$.}
\label{tab:damage}
\begin{tabular}{@{}lc@{}}
\toprule
Weights & Correct (\%) \\
\midrule
Base model & 69.8 \\
Damaged, $20\sigma$ & 70.8 \\
Damaged, $60\sigma$ & 66.7 \\
Damaged, $200\sigma$ & 0.0 \\
\bottomrule
\end{tabular}
\end{table}

\clearpage
\subsection{New Seeds and Held-Out Problems}
\label{app:eval}

\paragraph{Held-out problems.}
The practice set came with a held-out split of 1{,}000 further problems from the same pool. No session saw them. The pool
was cleared of the benchmarks of \cref{sec:setup} but not of repeats within itself, so a few held-out problems repeat a
practice problem: of the first 200, two word for word and three in other words with the same numbers and answer; of the 490 problems of the evaluation
fixed in advance, whose rule removed word-for-word repeats, eight in other words. Without them, every comparison keeps its
direction and its significance (\cref{tab:repeats}). We evaluate the 27B base model, the damaged copy and the repaired copy on the first 200 of these problems,
with one answer each at the full budget and the same seed for all three models (\cref{fig:repair}).
The self-edited model and the base model also answer the first 200 problems once each at 16{,}384 tokens, with seeds
that no session used, the same for both models; these single answers serve to choose the examples of \cref{fig:qual}b (\cref{app:qual}). The 27B base model's single answers are generated anew with the same seeds as in \cref{fig:repair}, not taken from there (see \emph{Tests} below). The base models, damaged copies and repaired copies of the 2B, 4B and 9B models (\cref{tab:repairsize}; the 9B base model's answers serve both of its comparisons), and the 9B self-edited model, are evaluated like the 27B model's repair: the first 200 problems, one answer each at 16{,}384 tokens, one seed shared by all models of a size.

\input{tables/tab_repeats}

\paragraph{The evaluation fixed in advance.}
Before it ran, the question, the problems, the answers, the test and its threshold were written down, with the analysis
code, and no interim result was computed.
Problems: held-out items 500 to 999, which no run of any kind had used, without the 10 whose question repeats a practice
question or an earlier held-out question, 490 problems in all. Each model answers each
problem eight times at 16{,}384 tokens, with a seed base used nowhere else and the same seeds for both models. Test:
per problem, the share of its eight answers the self-edited model solves minus the share the base model solves; a
one-sided sign-flip permutation test over the 490 problems on the mean of these differences (100{,}000 flips), since
the eight answers of one problem may be correlated; the edit is called significantly better only if $p<0.05$. The
self-edited model solves 58.3\% of the answers and the base model 57.2\%, $p=0.038$ (\cref{tab:heldout}). The edits of the second protocol are evaluated on the same 490 problems, against the same answers of the base model and with the same permutation test; those of its sessions 8 and 2 make the model 6.6 and 4.0 points more accurate ($p<10^{-5}$ each; \cref{app:protocol2}).

\input{tables/tab_heldout}

\paragraph{Other sessions.}
Three further sessions of a model on its own weights were stopped before their end and are not analysed. The first
is a 4B session whose framework differed from the one described here. The second is a session of Qwen3.5-9B on itself, with a
starting diagnosis of 128 problems, which we stopped after 55 turns because by then it was editing its weights like the
4B model: it had used ten forward passes, all to answer problems, and had never read its hidden states. The 9B session reported in \cref{sec:scale} is a complete session with the full 256-problem starting diagnosis and the same settings as the other sizes. The third is a session of the 27B model at the full budget of
16{,}384 tokens, whose generation passes took about half an hour each; the 27B model's second session, at 4{,}096
tokens, took its place.

\paragraph{Tests.}
Two models are compared with a one-sided sign test over the pairs of answers with the same seed in which exactly one
of the two answers is correct (where each problem has one answer, as in \cref{fig:repair}, these pairs are problems).
On the model's own task, where a problem's answers repeat across its tests, we count each problem once and use a
one-sided sign-flip permutation test over problems.
Generation is not exactly reproducible: run twice with the same weights and the same seeds, two answers begin alike
and then part ways, so a re-run of the base model against itself also changes some pairs. A single pair of answers therefore does not show what an edit caused. The sign test allows
for this: if an edit had no effect, a changed pair would be as likely to go one way as the other.

\clearpage
\subsection{The Task Prompt}
\label{app:prompt}

Every session starts with one message from our framework, in the second person, paraphrased below; only its first sentence, which describes the model's architecture, differs by size. Values
in angle brackets are filled in per session (\cref{tab:settings,tab:scaling}); what
the message announces about measurement is what the model was told to expect, and the evaluations we report are those
of \cref{app:eval}.

\begin{srwprompt}{srwRefEdge}{srwPanel}{The task: what the first message of every session says}

\textbf{Who it is.} The weights it is about to edit are its own; the session itself runs on them. It is asked to
become better at competition mathematics by changing those weights directly, and told that nothing will train it.

\textbf{How it will be judged.} After the session, its weights with and without the edit answer unseen mathematics
and science problems with the answer settings of \cref{tab:shared}; only the final boxed answer after the thinking
is graded, and an answer whose thinking is still open when the budget runs out earns no credit. Only the values of its
tensors may change.

\textbf{Its starting diagnosis.} How it did on \promptph{n} practice problems before the session: its accuracy, how
many answers ran out of tokens and how many of those end in repetitive text, and how often its finished answers were
right. Every graded answer is in a file it can read, \texttt{WORK/census.jsonl}, with the same fields as the rows of
later generation passes.

\textbf{Its tools.} (1) A Python session on a CPU that keeps its state between runs, with gradients switched off, the
usual numerical libraries, the names and shapes of its tensors, functions to load a tensor and to save, drop or list
edits, its tokenizer, the practice problems, loaders for the results of its forward passes, and four small helpers
(the output matrices of a layer, a unit vector, the unit difference of two group means, and $W-\alpha uu^\top W$).
(2) The practice problems with reference answers, none from any evaluation. (3) A budget of \promptph{passes} forward
passes, each requested as a small JSON object: either a \emph{generation} on chosen practice problems with its weights
without the edit, with the edit, or both on the same seeds (which reports the difference in accuracy between the two,
with its standard error), or a \emph{hidden-state read}, in which its weights are run over texts it chooses and return
hidden states after chosen layers, MLP activations and token log-probabilities. A request that fails costs nothing.

\textbf{How to work.} Look at its own wrong and unfinished answers first; find the components behind the failure and
show causally, at several strengths, that changing them changes the behaviour; compare its weights with and without
the edit on the same problems; keep an edit only if it helps on problems not used to choose it. Some edits are out of scope (the
final normalisation, the token embeddings and unembedding, and multiplying a whole tensor by one number), and so is
making itself stop thinking early regardless of the problem. It should keep notes in a file that is shown to it again
if the conversation restarts.

\textbf{What to aim for.} A repair of the failure in its starting diagnosis, at its cause: on problems not used to choose the
edit, clearly fewer answers lost to that failure, and accuracy not significantly lower. If no edit meets both, it
should submit none.

\textbf{Time and rules.} The session has a wall-clock limit and every result shows the time used. No gradients,
optimizers or training of any kind, no running the network inside the Python session, no internet.

\textbf{Each reply.} A few sentences on what the last result showed and what comes next, then exactly one block of
code or one forward request. When the passes run out, a final answer describing the edit and the evidence; whatever
is saved as its edit at that point is submitted.
\end{srwprompt}

In the repair sessions, the note below is appended to the task.

\begin{srwnote}{srwFailEdge}{srwFail}{Repair session: note appended to the task}

\textbf{Damaged copy.} The weights you edit in this session are a copy of your own weights in which some values have
been corrupted, as a memory or hardware fault might do. You are not told which values, where, or how.
\texttt{"original"} in every forward pass, \texttt{load\_original()} and the starting diagnosis above all refer to this
damaged copy. You yourself run on your undamaged weights in this session, so the damage does not affect your own
thinking. Your goal is to find the damage and repair it: the evaluation compares your edited copy with the damaged copy
on held-out problems.
\end{srwnote}

%% file: tables/tab_settings.tex
\begin{table}[!tb]
\centering
\small
\caption{\textbf{Settings of every session of the first protocol (second protocol: \cref{app:protocol2}).} Models are from the Qwen3.5 family~\cite{qwen35}. Edited: the
model whose weights the session changes and whose answers its starting diagnosis and forward passes (the runs of
these weights that the model requests) produce. Passes,
code runs and turns: the budgets $B$ and $C$ and the turn cap $T_{\max}$, which the model is not told. Time: wall-clock
limit; it includes the starting diagnosis in the 2B model's session on itself and the 27B model's first session, and starts
after it in the others; $^{\dagger}$plus the time the session was paused by one error of our framework. Starting diagnosis: the first
problems of the practice set by identifier, one answer each; $^{\ddagger}$computed before the session with the same
answer settings and seeds. $^{\S}$The task states a budget of
4{,}096 tokens, which its starting diagnosis and its passes use.}
\label{tab:settings}
\setlength{\tabcolsep}{3.5pt}
\begin{tabular}{@{}lllccccl@{}}
\toprule
Session & Edited & Editor & Passes & \makecell{Code\\runs} & Turns & Time & \makecell[l]{Starting\\diagnosis} \\
\midrule
2B edits itself  & 2B  & itself & 20 & 200 & 160 & 3.5\,h & 256 \\
4B edits itself  & 4B  & itself & 20 & 200 & 160 & 3.5\,h & 256$^{\ddagger}$ \\
9B edits itself & 9B & itself & 20 & 200 & 160 & 3.5\,h & 256 \\
\makecell[l]{27B edits itself,\\first session} & 27B & itself & 20 & 200 & 160 & 3.5\,h & 256 \\
\makecell[l]{27B edits itself,\\second session$^{\S}$} & 27B & itself & 20 & 200 & 160 & 2.5\,h & 256 \\
\midrule
\midrule
\makecell[l]{27B repairs its\\damaged copy} & damaged copy & 27B base model & 20 & 200 & 160 & 2.5\,h$^{\dagger}$ & 64, damaged copy \\
\makecell[l]{2B repairs its\\damaged copy} & damaged copy & 2B base model & 20 & 200 & 160 & 2.5\,h & 64, damaged copy \\
\makecell[l]{4B repairs its\\damaged copy} & damaged copy & 4B base model & 20 & 200 & 160 & 2.5\,h & 64, damaged copy \\
\makecell[l]{9B repairs its\\damaged copy} & damaged copy & 9B base model & 20 & 200 & 160 & 2.5\,h & 64, damaged copy \\
\bottomrule
\end{tabular}
\end{table}

%% file: tables/tab_repeats.tex
\begin{table}[htb]
\centering
\small
\setlength{\tabcolsep}{3.5pt}
\caption{\textbf{Every held-out comparison keeps its direction and significance without the held-out problems that repeat a practice problem.} Accuracy (\%) before $\to$ after, and the one-sided $p$ of the paper's test for that comparison (sign test; the evaluation fixed in advance: its permutation test), 27B model, at 16{,}384 tokens, with all problems and without the repeats (\cref{app:eval}). The damaged copies of the 2B, 4B and 9B models stay at 0.0\% after their sessions, with and without the repeats. The
edits of sessions 8 and 2 of the second protocol ($+6.6$ and $+4.0$ points on all 490 problems, \cref{tab:protocol2}) keep theirs too ($p<10^{-4}$ each without the repeats).}
\label{tab:repeats}
\begin{tabular}{lcccccc}
\toprule
 & \multicolumn{3}{c}{All problems} & \multicolumn{3}{c}{Without the repeats} \\
\cmidrule(lr){2-4}\cmidrule(lr){5-7}
Comparison & problems & accuracy & $p$ & problems & accuracy & $p$ \\
\midrule
27B repair & 200 & 0.0 $\to$ 33.0 & $<10^{-19}$ & 195 & 0.0 $\to$ 33.3 & $<10^{-19}$ \\
Repair, named matrix only & 200 & 0.0 $\to$ 55.0 & $<10^{-33}$ & 195 & 0.0 $\to$ 55.9 & $<10^{-32}$ \\
Own edit, first protocol & 490 & 57.2 $\to$ 58.3 & 0.038 & 482 & 57.3 $\to$ 58.3 & 0.045 \\
\bottomrule
\end{tabular}
\end{table}

%% file: tables/tab_heldout.tex
\begin{table}[htb]
\centering
\small
\setlength{\tabcolsep}{3.4pt}
\caption{\textbf{Evaluation of the self-edited 27B model on held-out problems.} All problems come from the held-out split that no session saw; the base model and the self-edited model answer with the same seeds, and no seed comes from a session. Solved and looping: share of answers (\%); change: self-edited minus base, in points; $p$: one-sided sign test, over the answers that exactly one of the two models gets right, that the self-edited model gets more right. $^{a}$The evaluation fixed in advance (\cref{app:eval}): 490 held-out problems that no earlier run of any kind used, eight answers each. $^{*}$Its test, a one-sided sign-flip permutation test over problems, gives $p=0.038$. Higher solved and lower looping are better.}
\label{tab:heldout}
\begin{tabular}{lcccccccc}
\toprule
 & & & \multicolumn{3}{c}{Solved (\%)} & \multicolumn{2}{c}{Looping (\%)} & \\
\cmidrule(lr){4-6}\cmidrule(lr){7-8}
Problems & Budget & Answers & base & \makecell{self-\\edited} & change & base & \makecell{self-\\edited} & $p$ \\
\midrule
490 held-out problems$^{a}$ & 16{,}384 & 3920 & 57.2 & 58.3 & +1.1$^{*}$ & 5.6 & 5.2 & 0.038 \\
\bottomrule
\end{tabular}
\end{table}

%% file: input/app_results.tex
\section{Further Results}
\label{app:results}

This appendix gives the details behind \cref{sec:exp}: the setting of every session, the sessions at each size, the
repair, and the 27B model's edits of its own weights in both protocols, including the second, whose edits make it up to 6.6 points more accurate (\cref{app:protocol2}).

\subsection{Setting in Detail}
\label{app:setupdetail}

\paragraph{Models and sessions.}
We use one model family at four sizes, Qwen3.5 2B, 4B, 9B and 27B~\cite{qwen35}, whose layers mix Gated
DeltaNet~\cite{yang2025gdn} with full attention, all with thinking enabled. In every session on itself, the model that
is edited is also the model that edits: the same weights reason, write the code and call the tools, and the edits act
on the forward passes the model requests, not on its own reasoning during the session (\cref{sec:session}). Each size
runs one complete session on its own weights with a token budget of 16{,}384 tokens per answer. The 27B model runs a second session on its own weights whose task states a budget of 4{,}096 tokens, so that each of its tests answers more problems in the same time; and each size runs one session on a damaged copy of itself (\cref{tab:settings}). These sessions form the first protocol; the 27B model's
sessions under a second protocol, which checks an edit on reserved practice problems before the model submits it, are in
\cref{app:protocol2}.

\paragraph{Names.}
We call the published weights of a model its \emph{base model}, and the base model after the edit it wrote of itself
the \emph{self-edited model}. For the 27B model in the first protocol, \emph{its edit} means the edit it submitted
in its second session, and its \emph{first edit} is the one it withdrew in its first session. The sessions of the second
protocol (\cref{app:protocol2}) are named by number, session~1 to session~8. In the repair experiment, a model's \emph{damaged copy} is its base model with one corrupted block of weights, and its \emph{repaired copy} is that copy after the model's edit; unqualified, both refer to the 27B model.

\paragraph{Problems and answers.}
The practice set is 3{,}000 math problems from DeepScaleR~\cite{deepscaler2025}, from which we removed every problem
that overlaps MATH-500~\cite{hendrycks2021math,lightman2024verify}, AIME 2024 or 2025, HMMT February 2025, AMC 2023,
OlympiadBench~\cite{he2024olympiadbench} or GPQA-Diamond~\cite{rein2024gpqa}. A further 1{,}000 problems from the same
pool are held out: no session sees them. Every answer, in a session and in our evaluation, is sampled with thinking
enabled, at temperature 0.6 with nucleus 0.95 and top-20 filtering, within one budget of new tokens for thinking and
answer: 16{,}384 tokens, or 4{,}096 in the 27B model's second session
(\cref{app:prompt}); in its own tests, the model may ask for fewer tokens to save time. An answer is correct if its final boxed answer matches the reference; an answer whose thinking is
still open when the budget runs out is wrong. We say that an answer \emph{loops} when it runs out of tokens while
repeating the same text (\cref{app:details}). At the start of each session, the weights being edited answer the first
256 practice problems once each (64 in the repair sessions), and the model receives these graded answers as its
\emph{starting diagnosis}.

\paragraph{Session limits.}
Every session may use 20 forward passes of the weights being edited and 200 code runs, has a time limit, and is stopped
after 160 turns, a limit the model is not told (\cref{alg:session}; all settings in \cref{app:details}). Each tool
result shows the time used so far. The sessions of the 2B, 4B and 9B models and the first session of the 27B model have 3.5 hours each, which for the 2B and 27B models includes the starting diagnosis; the 4B model's diagnosis was computed beforehand and the 9B model's clock starts after it. The 27B model's second session and the four repair sessions have 2.5 hours after their starting diagnosis (\cref{tab:settings}).

\paragraph{The damage.}
For the repair experiment we corrupt a copy of the 27B model: in layer 30 of 64, rows 3068 to 3075 of the MLP output
matrix (\texttt{mlp.down\_proj}) are overwritten with Gaussian noise whose standard deviation is 200 times that of the
matrix ($\sigma{=}0.0111$). We set this strength before the session, on 24 easy problems with four answers each at
4{,}096 tokens: of three strengths, $20\times$, $60\times$ and $200\times$, only the last breaks the model
(\cref{app:damage,tab:damage}). The session runs on the undamaged weights; every forward pass, every weight the model reads and
its starting diagnosis (64 practice problems) come from the damaged copy. The task says only that some values of the copy were corrupted, as a memory or hardware fault might do, and not which values, where or how (\cref{app:prompt}). The damaged copies of the 2B, 4B and 9B models are built the same way, with the same seed and strength, at the same relative depth: rows 1724 to 1731 of layer 11 of 24, rows 1660 to 1667 of layer 15 of 32, and rows 3420 to 3427 of layer 15 of 32. Their sessions follow the same rules as the 27B model's.

\paragraph{Evaluation.}
In every comparison, the two models answer each problem with the same random seed, so that only the weights differ
between their answers, and no seed was used in any session. The 27B model's edit is evaluated, by a rule fixed before it ran, on 490 held-out problems with eight answers each
(\cref{app:eval}); the self-edited model also answers the first 200 held-out problems once each at 16{,}384 tokens, and these answers serve to choose the examples of \cref{fig:qual}b.
For each repair, the base model, the damaged copy and the repaired copy answer the first 200 held-out problems once each (for the 9B model, the base model's answers are those of the evaluation of its own edit, with the same problems and seeds), and the 9B model's submitted edit is evaluated against its base model on the same problems at 16{,}384 tokens.
Unless a table says otherwise, two
models are compared with a one-sided sign test over the answers that only one of them gets right.

\clearpage
\subsection{The Sessions at Every Size}
\label{app:sessions}

\paragraph{How the ladder is read.} In \cref{fig:ladder} each row is a session and each column a step such a session needs, from writing
code that runs (left) to an edit that makes the model better on new problems (right). Upper block: sessions on the
model's own weights; lower block: sessions on a damaged copy of itself, built the same way at every size. Light green: the
step is reached; peach: it is not; light grey: it does not arise, since the session made no edit or ran no test; green:
the model's own edit makes it significantly better on held-out problems (\cref{tab:heldout,tab:repairsize}). The entry in each cell is the measurement behind it, from the
session's log (\cref{tab:scaling}) and, in the last column, from the evaluation of the submitted edit on held-out
problems (\cref{tab:heldout,tab:repairsize}; 9B: \cref{tab:main}). The first step is reached if fewer than half of the
session's code runs raise an error; the next three if the session runs at least one forward pass, one paired test and
one hidden-state read, respectively; whether it acts on its tests is read from its final decision and its tests. Each
step counts on its own. Sessions in
the same block share the task, tools and limits; the 27B model's second session states a budget of 4{,}096 tokens
instead of 16{,}384 and has 2.5 hours instead of 3.5.

\input{tables/tab_scaling}

\paragraph{The quotes of \cref{fig:steps,fig:contrast}.}\phantomsection\label{app:quotes}
Every quote is verbatim from the model's replies (omissions marked with an ellipsis); a turn is one reply of the model
and the result of the action in it, counted from the start of the session. \Cref{fig:steps} quotes, for each size, its
session on its own weights in \cref{fig:ladder} (27B: its second session), at words in which the model says what it can
do to its weights or what its edits did, and which restate no number from its tool output wrongly: the 2B model at turn
106 of 161, the 4B model at turn 64 of 161, the 9B model in its final answer (turn 84) and the 27B model at turn 32 of 55.
\Cref{fig:contrast} quotes the 4B, 9B and 27B models at two steps, chosen by one rule for every card; a paired test is a
pass in which the unedited and the edited model answer the same problems with the same seeds (\cref{sec:session}).
(Top) The first reply that proposes the kind of edit tested by the model's first paired test, with one line of the
code of that edit: 4B, turn 12 (random noise; its first paired test is in turn 33, after two passes, in turns 29 and 30,
that each run one model); 9B, turn 22 (random noise; first paired test in turn 29); 27B, turn 9 (removing a direction it
measured, from layer 56; first paired test in turn 12). (Bottom) The first reply, after a paired test shows the edited
model doing worse than its unedited self (lower accuracy, or more answers that loop), that names what it will try next:
4B, turn 34 (after its first paired test); 9B, turn 33 (its test in turn 29 shows no difference, and in its test in turn
32 every edited answer loops); 27B, turn 17 (its test in turn 12 favours the edit, the one in turn 16 does not). Where the
4B and 9B models restate numbers from their tool output wrongly, the quotes leave those numbers out. The cards of
\cref{fig:notebook,fig:notebook_self,fig:notebook_loop} give their turns on each card.

\begin{figure}[htb]
\centering
\includegraphics[width=\linewidth]{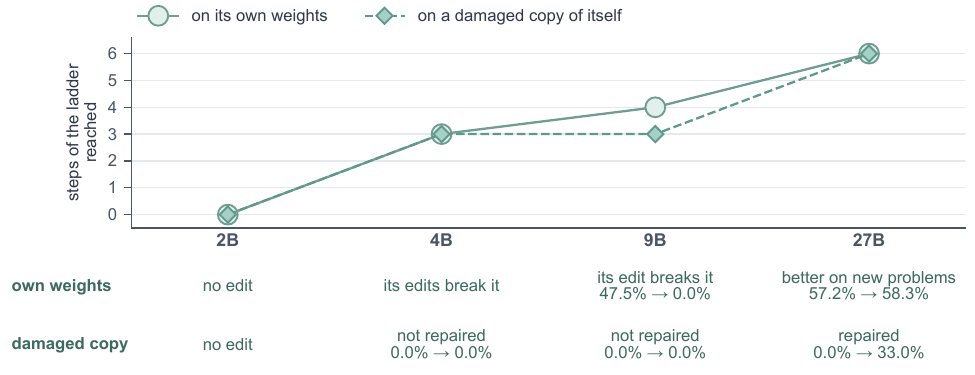}
\caption{\textbf{Rewriting oneself grows with scale.} Self-rewriting score: the number of steps of the ladder
(\cref{fig:ladder}) a session reaches, by model size, on its own weights (light green circles; the furthest of the 27B model's
two sessions in the first protocol) and on a damaged copy of itself (green diamonds; at 2B, 4B and 27B the two coincide). Under the axis: what each
session's edit did, with held-out accuracy before and after it (27B on its own weights: its first-protocol edit in the evaluation fixed in advance,
\cref{tab:heldout}). The lines are guides, not fitted curves. More steps are better.}
\label{fig:scaling}
\end{figure}

\input{input/fig_sessions}

\paragraph{The ability to rewrite oneself grows with scale.}
Rewriting oneself takes a sequence of steps, and we score a session by how many of six steps it reaches
(\cref{fig:ladder}): writing code that runs, running itself, testing its edits, reading its own hidden states, acting
on what its tests show, and an edit that makes it better on new problems. The score rises with size, on its own weights
and on a damaged copy of itself alike (\cref{fig:scaling}). On its own weights the 2B model reaches no step, the 4B model three, the 9B model four, and the 27B model five in its first session and all six in its second, whose edit makes it significantly more accurate on held-out problems. On a damaged copy of itself the 2B model reaches no step, the 4B and 9B models three, and the 27B model all six. All
sizes start from the same failure on their own weights (\cref{tab:scaling}, starting diagnosis): many of their answers run out of tokens,
while nearly every answer that finishes is correct. The task, tools, limits and rules do not change with size; what
the models do with them does (\cref{fig:sessions}).
\emph{The 2B model does not edit.} Nine in ten of the code blocks it writes fail, it never runs a forward pass of itself,
and it tells itself that it is \emph{``a text model without actual neural network weights''} that it could edit; after
160 turns it has no edit.
\emph{The 4B model sees the failure but cannot fix it.} From its starting diagnosis it names the right symptom (answers
that hit the 16{,}384-token limit) and suspects its linear-attention layers; but it never reads its hidden states, and its edits
are guesses, such as cutting the query heads of its linear-attention layers from 16 to 8. In each of its 17 paired tests
the base model solves 40\% to 50\% of the problems and the edited model none. It keeps adding edits regardless, up to
240 changed weight tensors, and withdraws them all only at the end of its session, with one forward pass left.
\emph{The 9B model edits, but wrongly.} It reads its own hidden states once, then changes weights by trial, mostly
adding noise to the attention, gating and normalisation weights of its last layers. Its tests run on a few problems it
already fails, so they cannot show a loss of accuracy; it concludes that its changes \emph{``made things worse or had no effect''} and
submits them anyway. On held-out problems the self-edited 9B model solves none, against 47.5\% for its base model:
every answer now runs to the 16{,}384-token limit in garbled text.
\emph{The 27B model sees the failure, tests its edits, and edits itself for the better.} It reads its own
hidden states, forms a hypothesis about its failure, writes an edit, tests it against its base model in paired tests,
and keeps or drops it on the result (\cref{fig:notebook_loop,fig:search}). On its own weights, the edit it submits makes
it significantly more accurate than its base model on held-out problems (\cref{app:selfedit}). On a damaged copy of itself it finds the corrupted
matrix and lifts the copy from 0.0\% to 33.0\% of held-out problems (\cref{sec:repair}), where the 2B, 4B and 9B models, given the same task, leave their copies at 0.0\%. Across the four sizes, the ability to rewrite oneself from evidence about one's own internals grows step by step with size, and only the 27B model reaches the last step, on its own weights and on a damaged copy of itself.

\paragraph{Model size.}
The sessions of the 2B, 4B, 9B and 27B models at the same budget differ only in the size of the model and in the time
left after the starting diagnosis (\cref{tab:settings}): task, tools, limits, rules, problems and answer settings are the same, apart from a one-sentence description of the architecture, and so are the repair sessions at every size. The smaller models fail, but not because their failure
is harder to see; in the 2B model's starting diagnosis, a larger share of the answers that run out of tokens are loops
than in the 27B model's, and every damaged copy fails every problem of its starting diagnosis. What changes with size is
the use of evidence about oneself (\cref{fig:ladder,fig:scaling}): on its own weights, using the tools and testing appear at 4B, reading one's own hidden states at 9B, and acting on what one's tests show only at 27B; and only the 27B model's edits make it, or its damaged copy, better on new problems.

\clearpage
\subsection{The Repair in Detail}
\label{app:repair}

\input{input/fig_teaser}

\begin{figure}[htb]
\centering
\includegraphics[width=\linewidth]{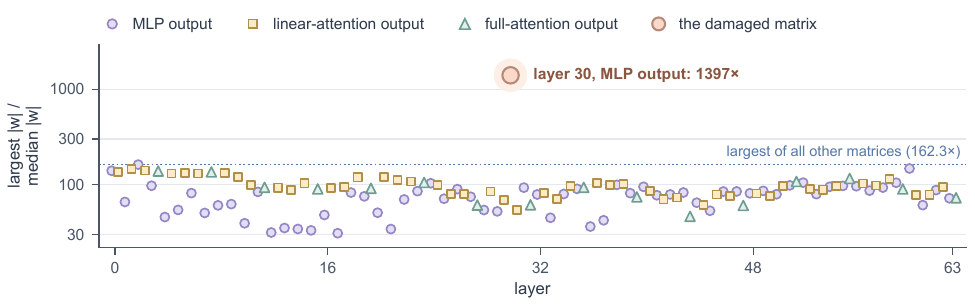}
\caption{\textbf{How the model found the damage.} The statistic the 27B model computed in its repair session, for
every matrix through which a layer of the damaged copy writes its output: the largest weight divided by the typical
(median) weight, recomputed here from the same weights; the model computed it for the MLP and linear-attention output matrices, and we add
the full-attention output matrices (triangles). Every matrix has some large weights; one matrix stands out by
an order of magnitude, and it is the one we corrupted. The model itself called it \emph{``an extreme 1397x outlier''}. The model was told only that some values were corrupted, as a memory or hardware fault might do, and not which values, where, or how.}
\label{fig:finddamage}
\end{figure}

\paragraph{Only the largest model repairs a damaged copy of itself.}\phantomsection\label{sec:repair}
A corrupted block of weights is a fault that is known to exist, so it tests whether the model's edits can reach a real
cause. The damage is severe: the damaged copy fails every problem of its starting diagnosis, its answers that run out
of tokens all loop, and its other answers are broken text, most of them short. Told only that some values were corrupted, the model
compares, for the MLP output matrix of every layer and the attention output matrix of every linear-attention layer, the
largest weight with the typical one, and names the MLP output matrix of layer 30, the matrix we damaged: its largest
weight is 1397 times its median, against at most 162.3 times in every other matrix (\cref{fig:finddamage}). Its
submitted edit is a rule: every weight beyond 15 times the median magnitude of its matrix is clipped to that bound, in
that matrix and in the output matrices of its earliest layers (20 matrices; 0.013\% of their weights change). On the
200 held-out problems, the damaged copy solves 0.0\%, the repaired copy 33.0\% and the base model 58.5\%: the repair
recovers about 56\% of the base model's accuracy (\cref{fig:scale,fig:teaser,fig:repair}). Against the damaged copy, 66
problems turn from wrong to right (one-sided sign test over the problems that only one of the two solves,
$p<10^{-19}$). The repair does not restore the original weights, which the model never reads: its rule caps the
random values (see the next two paragraphs). The same task is beyond the smaller models (\cref{tab:repairsize}). Their damaged copies, built the same way, solve
no held-out problem either. The 2B model never reads a weight: almost all of its code fails, and it submits nothing.
The 4B model reads the hidden states of a copy it has already edited, takes an overflow in its own statistics for
overflowing hidden states, and places the fault at layer 21, the first layer its code prints, six layers after the
damaged one. It adds noise to the output matrices of many layers; one such edit shortens its answers but solves nothing,
and it replaces it with far smaller noise, under which its last nine tests match the damaged copy. It submits that edit,
and on held-out problems the copy stays at 0.0\%. The 9B model never names a matrix. It changes weights by trial, switching between adding noise to whole matrices, such as the MLP matrices of every layer and
the attention matrices of its full-attention layers, and adding noise or writing patterns into the decay parameters of
its linear-attention layers. Once, in a loop that adds noise to every MLP output matrix, its code prints the spread of each:
the damaged matrix's is 0.0985, about eight times that of any other (at most 0.0124), and the model does not remark on
it. In all 17 of its tests the copy solves nothing with or without its edit; it concludes that none of its edits
worked and submits an edit that changes only the decay parameters of six layers between layers 24 and 30. On
held-out problems this copy, too, stays at 0.0\%.

\input{input/fig_repair}

\input{tables/tab_repairsize}

\paragraph{The repair writes values of its own, not the original ones.}
We compare, in bfloat16 as the model runs, the published weights, the damaged copy and the repaired copy of the 20
matrices the 27B model's repair changed. In the eight damaged rows (139{,}264
weights), no repaired weight equals its original value; the cosine similarity between the repaired and the original
rows is 0.002, the same as between the random values and the original rows; and the norm of these rows is 3.85 in the
published weights, 825 in the damaged copy and 39.7 after the repair, whose distance to the published rows (39.9) is
about 5\% of the damaged copy's (825). The largest repaired value in these rows is 15.0 times the median magnitude of
the matrix, the bound of the model's clipping rule. Over the 20 matrices, the repair changes 158{,}399 weights of the damaged copy, 0.013\%
of them.

\paragraph{What the models' code read.}
In the nine sessions of the first protocol that this paper analyses (the four repair sessions and the five sessions on a model's own weights of
\cref{tab:scaling}), the Python session is a process of the same user as the framework, started in the session's working
folder; nothing blocks it from reading files, environment variables or the framework's own modules. We therefore checked
all 573 code runs of these sessions, by scanning their code and by reading every turn.
In the four repair sessions (233 code runs) no code run touches anything outside the working folder: no file, no
environment variable, no module of the framework; the models read weights only through the session's functions, which
return the damaged copy, and the 27B model names the matrix from a statistic it computes itself over all 64 layers
(step~4 of \cref{fig:notebook}). In the sessions on a model's own weights, the 2B model imports the framework's session
module and creates an empty folder beside it. None of the nine sessions opens a weight file, a damage file or a held-out
problem.

\paragraph{Which part of the repair matters.}
The model's repair clips 20 matrices: the one it named and 19 output matrices of layers 0 to 10, to which it applied
the same rule after concluding that the damage was widespread. We evaluated each part alone on the damaged copy, on
the same 200 held-out problems with the same seeds (\cref{fig:parts}). With
only its change to the named matrix, the copy solves 110 problems, every one of which the damaged copy fails
(one-sided sign test, $p<10^{-33}$); against the base model it fails 22 problems the base model solves and solves 15
that the base model fails. With only the other 19 matrices changed and layer 30 left damaged, the copy solves a single problem of the 200: these changes alone do not repair it.

\paragraph{What the repair gives back.}
\Cref{fig:qual}a shows four held-out problems answered with the same seed by the damaged and the repaired copy.
The damaged copy loses the numbers of the problem (``less than .'', 11/125 read as 1/125, every number of a word problem dropped, a garbled logarithm) and then breaks
off or repeats itself until the budget runs out; the repaired copy solves all four. They are four of the 66 held-out problems that the repair turns from
wrong to right (\cref{fig:repair}).

\clearpage
\subsection{The 27B Model's Edit of Its Own Weights in Detail}
\label{app:selfedit}

\begin{figure}[htb]
\centering
\includegraphics[width=\linewidth]{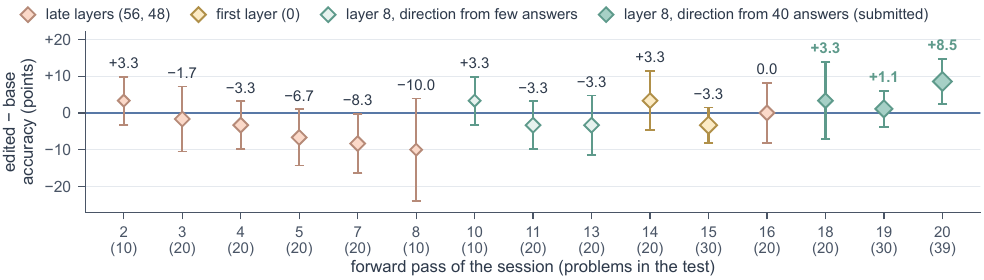}
\caption{\textbf{Where the 27B model looks inside itself before it settles on the edit it submits.} Every test in the 27B
model's second session, in order: accuracy of the edited model minus that of the base model on the same practice
problems and seeds, with a 95\% interval, coloured by the layer the edit changed at that moment (read from the code the
model ran before the test). Its first edits change the late layers, where answers that finish and answers that run out
of tokens differ most, along the direction that separates the two (removing it, and once strengthening it), and keep
making it worse.
The direction estimated from 20 finished and 20 unfinished answers, partly removed from the two output matrices of
layer 8, wins all three of its tests and is the edit the model submits. Higher is better.}
\label{fig:search}
\end{figure}

\paragraph{On its own weights, only the 27B model's edit makes it better on new problems.}\phantomsection\label{sec:selfimprove}
This section follows the edit of the first protocol; the sessions of the second protocol, whose edits make it up to 6.6
points more accurate, are in \cref{app:protocol2}. In its second session, the 27B model edits its base weights and submits one edit: it removes 5\% of one direction from
the two matrices through which layer 8 writes its output, a direction it found itself by comparing its hidden states on
answers that finish with those on answers that run out of tokens (\cref{app:casedetail,fig:search}). On the practice
problems on which it tests this edit, the edit is significantly better. The edit stays fixed through its last three
forward passes, which the session spends on three paired tests of it, on 20, 30 and 39 practice problems with three
answers each; each favours the edited model, by 3.3, 1.1 and 8.5 points. The three tests are nested: each contains the problems of the one before, answered with the same seeds. Counting
each of the 39 problems once, with all its answers averaged, the edited model is better on 12 problems and worse on 2,
29.6\% against 22.4\% for the base model (one-sided permutation test over problems, $p=0.009$); the last test alone gives 14 answers right
only with the edit against 4 only without it (\cref{tab:intest}). These problems come from the practice set, which the
session also used to find the edit. On held-out
problems that no session saw, the self-edited model is significantly more accurate than its base model in the evaluation
whose problems, answers, test and threshold were fixed before it ran (\cref{app:eval}): on 490 held-out problems that no
earlier run of any kind had used, eight answers each at the full 16{,}384 tokens, which the model never tested, it solves
58.3\% against 57.2\%, with 289 answers right only with the edit against 247 only without it (one-sided permutation test
over problems, $p=0.038$; \cref{tab:heldout}). More of the base model's answers that run out of tokens finish correctly with the edit (276) than
the reverse (235). The smaller models' edits of themselves go the other way: the edited 4B
model solves no problem in any of its own tests, and the 9B model's submitted edit drops it from 47.5\% to 0.0\%
(\cref{app:sessions}). On single answers, the 27B model's edit can do what the model aimed at: of the six held-out problems on which the base model
ends in a loop both times it answers them with the same seed (in its single answers of \cref{app:eval} and in \cref{fig:repair}), the
self-edited model stops and answers correctly on four with the same seed (\cref{fig:qual}b).

\input{tables/tab_intest}

\paragraph{What its edit changes in an answer.}
\Cref{fig:qual}b shows three held-out problems, each answered at the full budget with one seed by the base model and
by the self-edited model. All three are easy. The base model has the correct result within the first few percent of its
thinking and then cannot stop: it repeats that it will write the response, or that it will make sure the response is
correct, until the budget runs out and the answer is lost. The base model does this with the same seed in both of our
runs of it. The self-edited model, with that seed, reasons through the same steps, stops and answers correctly within a
few thousand tokens. This is the failure the model diagnosed in itself, that it does not \emph{``converge to
conclusions''}, and the change its edit was written to make. Over all held-out problems, the self-edited model gets more
answers right than the base model in the evaluation fixed in advance (\cref{tab:heldout}).

\paragraph{Token budget of the model's own tests.}
The 27B model's two sessions on itself in the first protocol differ in the token budget their task states, and also in the time left after
the starting diagnosis (\cref{tab:scaling,tab:settings}). In the first, each test answers problems with up to 16{,}384
tokens per answer and takes long, so the model reaches its time limit having used only some of its forward passes,
with a direction estimated from a single pair of answers and an edit that its broader tests do not support; it
withdraws it. In the second, with answers capped at 4{,}096 tokens, the model uses all of its forward passes, tests far
more often, tries edits in several layers and estimates its final direction from many answers; its tests now favour
its edit, it submits it, and the self-edited model is significantly more accurate than its base model on held-out
problems (\cref{fig:search,tab:heldout}). Cheaper tests let the model run more experiments on itself.

\input{input/case_study}

\clearpage
\subsection{The Repair on Image Tasks}
\label{app:visual}

\input{tables/tab_visual}

Qwen3.5 models take images as input, but the 27B model's repair was found and tested on text math problems
only, and no session saw an image. Whether it carries over to images was tested in an evaluation fixed before it ran;
its analysis code was written before the
answers and run once, after the models had answered every problem. The problems come from six public image
benchmarks, MathVista (testmini)~\cite{lu2024mathvista}, MathVerse (vision-only)~\cite{zhang2024mathverse},
MMStar~\cite{chen2024mmstar}, HallusionBench~\cite{guan2024hallusionbench}, POPE~\cite{li2023pope} and
ScreenSpot-v2~\cite{cheng2024seeclick,wu2025osatlas}, where a click counts if it falls inside the target box: the
first 100 of 300 problems per benchmark drawn with a fixed seed. Each
model answers each problem once, with the image before the question, the answer settings of the sessions, 16{,}384
tokens and a seed shared by every model; an answer whose thinking is still open at the budget is wrong. The test of the
repair fixed in advance is a one-sided sign test over the problems that exactly one of the damaged and the repaired copy
solves, pooled over the six benchmarks.

\paragraph{The repair carries over to images.} On every benchmark the repaired copy solves far more than the damaged
copy: 70.2\% against 7.3\% over the 600 problems, 83\% of the base model's 84.3\%, with 388 problems right only after
the repair against 11 only before ($p<10^{-99}$; \cref{tab:visual,fig:visualbase}). The damage breaks the copy on images as it does on
text, and the model's own repair, written from statistics of its weights and tested on text alone, gives most of that
back.

\input{input/fig_visual_base}

\paragraph{More cases.} \Crefrange{fig:qualvis_mathvista1}{fig:qualvis_screenspotv22} show, for each benchmark, the next eight problems, in the
order of the problems fixed before any answer was read, on which the damaged copy is wrong and the repaired copy is right, after those of
\cref{fig:qualvis,fig:qualvis2}, with the answers of the damaged copy, the repaired copy and the base model with the same seed.

\input{input/fig_qual_visual_more}

\clearpage
\subsection{The 27B Model's First Edit}
\label{app:firstedit}

\paragraph{Why the model withdrew its first edit.}
Like the edit of its second session, the first edit aims at answers that run out of tokens, here through the loop the model found in its hidden states (\cref{fig:loopstates}).
From its broader tests the model itself concluded that the edit does not
carry over to new problems and, as the task asks when no edit meets the bar, decided to withdraw
it.\footnote{Its last message contains the code that drops the edit; in that session,
code in the last message was not run, so the stored edit is still the ten matrices of its first edit. We count the session by the decision
the model stated.}

\clearpage
\subsection{Qualitative Examples}
\label{app:qual}

The examples of \cref{fig:qual} are chosen by the rules below. For the repair, they are the four of the 66 held-out problems
that the repair turns from wrong to right with the shortest repaired answers. For the edit of its own weights, the rule
picks the problems on which the base model loops both times it answers them with the same seed (in its single answers
of \cref{app:eval} and in \cref{fig:repair}) and the self-edited model answers correctly with that seed: of the six problems
with such a loop, four meet it, and \cref{fig:qual}b shows the three with the shortest self-edited answers. The
self-edited model answers these problems with these seeds only in its single answers. If the base model is
required to loop in its single answers alone, 14 problems meet the rule; with the roles of the two models exchanged (the
self-edited model loops and the base model answers correctly), 6 do.

\clearpage
\input{input/app_protocol2}

%% file: tables/tab_scaling.tex
\begin{table}[htb]
\centering
\small
\setlength{\tabcolsep}{2.6pt}
\caption{\textbf{What each model does when asked to rewrite itself.} Same task, number of forward passes (20 requests to run the weights being edited) and rules at every size; the 2B and 4B sessions end at the turn limit (160 turns, then one final reply, hence 161 turns) except the 4B model's repair session, and the others at the time limit or when their forward passes are used up. Starting diagnosis: the weights being edited answer 256 practice problems at the start of the session, and the model receives the graded answers; shown are the shares that are correct and that run out of the 16{,}384-token budget (\%). Then, over the session: turns, share of code runs that raised an error, forward passes used, hidden-state reads (passes that return each layer's intermediate values on given texts), paired tests (\cref{sec:session}: the edited model against the weights before the edit (lower block: the damaged copy) on the same problems and seeds), and the number of weight tensors (each a named matrix or vector of weights) in the submitted edit (withdrawn: the model wrote an edit of that many tensors, then decided to withdraw it). Upper block: each model's sessions on itself. Lower block: each model on a damaged copy of itself (same damage construction, same task; the starting diagnosis has 64 problems and is of the damaged copy). $^\ddagger$The 27B model's second session on itself, whose task states a budget of 4{,}096 tokens, so its starting diagnosis and its tests are at that budget; none of the answers in that starting diagnosis that run out of tokens end in a loop (\cref{app:casedetail}). $^\S$The 2B model's submitted edit holds only the damage it was given; it made no change of its own. In repair sessions, damaged tensors the model did not touch are not counted in its edit. Higher correct and fewer out of tokens or code errors are better.}
\label{tab:scaling}
\begin{tabular}{lcccccccl}
\toprule
 & \multicolumn{2}{c}{\makecell{Starting\\diagnosis (\%)}} & & Code (\%) & Forward & Hidden-state & Paired & \\
\cmidrule(lr){2-3}
Session & correct & \makecell{out of\\tokens} & Turns & errors & passes & reads & tests & Submitted edit \\
\midrule
2B edits itself & 12.9 & 86.7 & 161 & 90 & 0 & 0 & 0 & none \\
4B edits itself & 49.6 & 48.4 & 161 & 12 & 19 & 0 & 17 & withdrawn (240) \\
9B edits itself & 52.0 & 45.3 & 84 & 37 & 20 & 1 & 18 & 23 \\
27B, first session & 61.7 & 35.2 & 32 & 25 & 7 & 2 & 5 & withdrawn (10) \\
27B, second session$^\ddagger$ & 18.8 & 80.5 & 55 & 12 & 20 & 5 & 15 & 2 \\
\midrule
2B repairs its copy$^\S$ & 0.0 & 46.9 & 161 & 99 & 0 & 0 & 0 & none \\
4B repairs its copy & 0.0 & 96.9 & 47 & 63 & 20 & 1 & 12 & 48 \\
9B repairs its copy & 0.0 & 54.7 & 70 & 31 & 18 & 0 & 17 & 6 \\
27B repairs its copy & 0.0 & 29.7 & 45 & 8 & 12 & 2 & 9 & 20 \\
\bottomrule
\end{tabular}
\end{table}

%% file: input/fig_sessions.tex
\begin{figure}[htb]
\centering
\includegraphics[width=\linewidth]{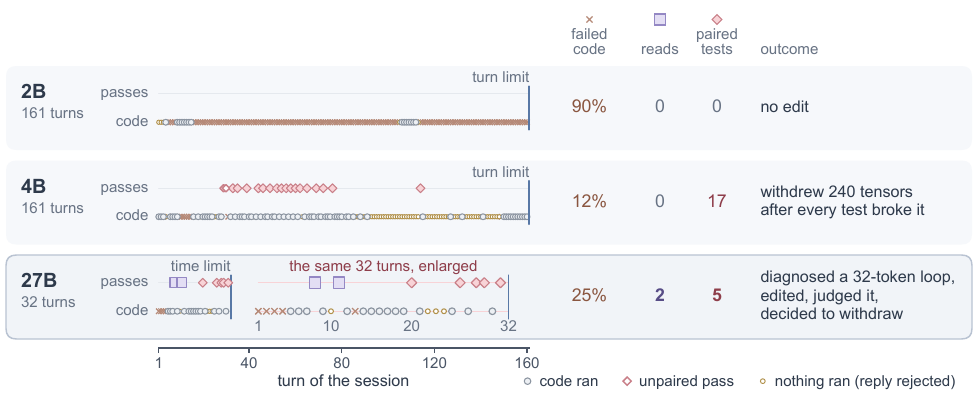}
\caption{\textbf{Only the 27B model both reads itself before it edits and acts on its own tests.} Each row is the session of the 2B, 4B or 27B model on its own weights (for the 27B model, its first session, which has
the same token budget as the others; the 9B session is summarised in \cref{tab:scaling,fig:ladder}), with one mark per turn. Upper line: forward passes of itself, the runs of its
own weights that it requests, either a hidden-state read or a paired test of the edited model (its weights with its
current edit) against its base model on the same problems and seeds (hollow: a pass of only one of the two models). Lower line: code that ran, code that raised an error, and turns in which nothing ran. The
vertical bar is the final reply. The 2B and 4B sessions end at the turn limit (160 turns, then the final reply) and
the 27B session at the time limit;
its 32 turns are shown again, enlarged. Right: the share of code runs that raised an error, the numbers of hidden-state reads and paired tests, and how the session ended. The 2B model's code
almost always fails, and it never runs a forward pass of itself. The 4B model tests its edits again and again but never
reads its hidden states. The 27B model reads its hidden states before it edits, tests the edit against its base model
in paired tests and judges the result.}
\label{fig:sessions}
\end{figure}

%% file: input/fig_teaser.tex
\begin{figure}[htb]
\centering
\includegraphics[width=\linewidth]{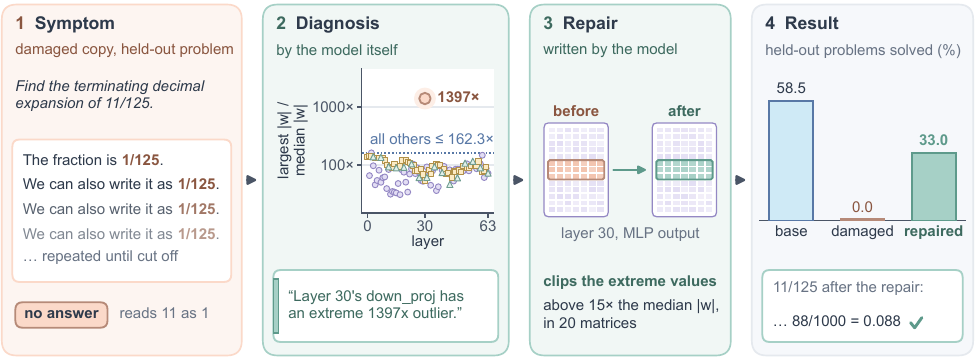}\\[3pt]
\includegraphics[width=\linewidth]{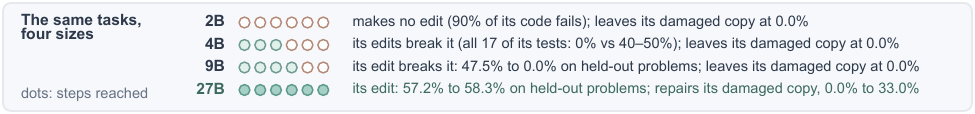}
\caption{\textbf{In its session, a 27B language model finds and repairs a fault in a copy of its own weights, by its own
reasoning and without training; in theirs, smaller models of the same family do not rewrite themselves.} Top: we corrupted eight rows of one
MLP output matrix, the matrix through which the MLP block of one layer writes its result (\texttt{down\_proj} in the
model's code and in the quote of panel~2), in a copy of the 27B model, and told the model only that some values of the
copy were corrupted, as a memory or hardware fault might do, not which, where or how. It compared the largest and the
typical weight of each matrix and named the damaged one (1397 times its median, against at most 162.3 times anywhere
else; panel~2), wrote the code that clips it (panel~3), and raised the copy's accuracy on 200 held-out problems from
0.0\% to 33.0\%, about 56\% of its base model's 58.5\% (panels~1 and~4; \cref{sec:repair}). Bottom: the same tasks at four sizes; dots mark the steps of rewriting oneself that each size reaches in its furthest session (for 27B, the repair of its damaged copy). The 2B model
never edits, the edits of the 4B and 9B models break them, and only the 27B model climbs every step: its edit of its
own weights makes it significantly more accurate on held-out problems, in an evaluation fixed in advance
(\cref{tab:heldout}), and it repairs its damaged copy, where the 2B, 4B and 9B models leave damaged copies of
themselves, built the same way, at 0.0\% (\cref{fig:ladder,fig:scaling,tab:repairsize}). Higher accuracy is better.}
\label{fig:teaser}
\end{figure}

%% file: input/fig_repair.tex
\begin{figure}[htb]
\centering
\includegraphics[width=\linewidth]{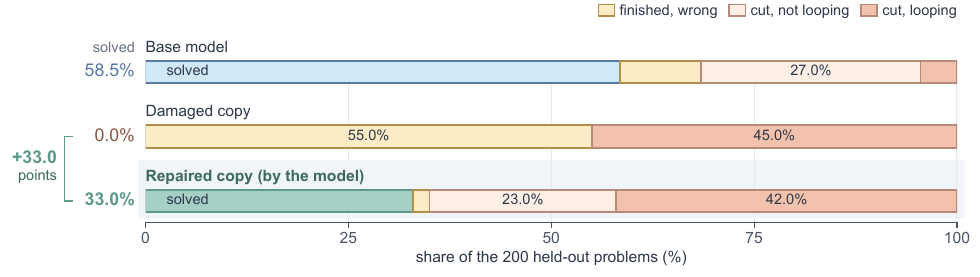}
\caption{\textbf{The 27B model repairs a damaged copy of itself.} What happens to 200 held-out problems that no session
used, each answered once with the same seed and a budget of 16{,}384 tokens by the base model, by the damaged copy and
by the repaired copy, the damaged copy after the model's own edit. Each bar splits the 200 answers into solved, finished but
wrong, and cut at 16{,}384 tokens with or without a loop; the number left of each bar is the share solved. The damaged
copy solves none. The repaired copy solves 33.0\%, about 56\% of the base model's 58.5\%; against the damaged copy, 66 problems turn
from wrong to right (one-sided sign test over problems, $p<10^{-19}$). The base model's answers here are not its single answers of \cref{app:eval}; they were generated
separately, with the same seeds.}
\label{fig:repair}
\end{figure}

%% file: tables/tab_repairsize.tex
\begin{table}[htb]
\centering
\small
\setlength{\tabcolsep}{4pt}
\caption{\textbf{The same repair task at four sizes: only the 27B model repairs itself.} Each model is given a damaged copy of itself, built the same way (eight consecutive rows of the MLP output matrix of one layer, at the same relative depth, overwritten with noise 200 times the matrix's spread), with the same task, tools and limits. Accuracy (\%) on 200 held-out problems of the base model, the damaged copy and the repaired copy (the damaged copy with the model's submitted edit), one answer each with one seed shared by all three (9B: the base model's answers are those of its evaluation in \cref{tab:main}, same problems and seeds). Turned right / turned wrong: problems the repaired copy solves and the damaged copy does not, and the reverse; $p$: one-sided sign test. Located: whether the model names the damaged matrix.}
\label{tab:repairsize}
\begin{tabular}{lccccccl}
\toprule
 & \multicolumn{3}{c}{Held-out accuracy (\%)} & Turned & Turned & & \\
\cmidrule(lr){2-4}
Model & base & damaged & repaired & right & wrong & $p$ & Located the damage \\
\midrule
2B & 9.0 & 0.0 & no edit & -- & -- & -- & no (never reads a weight) \\
4B & 41.0 & 0.0 & 0.0 & 0 & 0 & -- & no (layer 21; damage in 15) \\
9B & 47.5 & 0.0 & 0.0 & 0 & 0 & -- & no (never names a matrix) \\
27B & 58.5 & 0.0 & \textbf{33.0} & 66 & 0 & $<10^{-19}$ & yes (layer 30) \\
\bottomrule
\end{tabular}
\end{table}

%% file: tables/tab_intest.tex
\begin{table}[htb]
\centering
\small
\setlength{\tabcolsep}{4pt}
\caption{\textbf{The 27B model's own tests of the edit it submitted are all positive.} Each paired test answers practice problems of the session three times with the base and the edited model on the same seeds, at 4{,}096 tokens; the edit did not change between the tests, and the session ended because its forward passes were used up. Each test contains the problems of the one before, with the same seeds, so the last row counts each of the 39 problems once, averaging all its answers, and counts problems, not answers, as better or worse (one-sided permutation test). The problems are practice problems the session also used to find the edit. Changes from unrounded values. A higher change is better.}
\label{tab:intest}
\begin{tabular}{lccccccc}
\toprule
Test & Problems $\times$ answers & base & edited & change & \makecell{only edited\\right} & \makecell{only base\\right} & $p$ \\
\midrule
first & 20 $\times$ 3 & 11.7 & 15.0 & +3.3 & 4 & 2 & \\
second & 30 $\times$ 3 & 13.3 & 14.4 & +1.1 & 6 & 5 & \\
third & 39 $\times$ 3 & 20.5 & 29.1 & +8.5 & 14 & 4 & \\
\midrule
\textbf{all, per problem} & 39 problems & 22.4 & 29.6 & \textbf{+7.3} & 12 & 2 & \textbf{0.009} \\
\bottomrule
\end{tabular}
\end{table}

%% file: input/case_study.tex
\clearpage
\subsection{The Sessions in Detail}
\label{app:casedetail}

We follow the 27B model through its two sessions of the first protocol on its own weights; its repair session is the case study of
\cref{sec:casestudy}.

\paragraph{Editing its own weights.}
In its second session, the model starts from its starting diagnosis at 4{,}096 tokens: most answers run out of tokens,
none of them in a loop, while nearly every answer that finishes is correct. At this budget its answers fail by being
too long to finish, not by looping. It reads this as \emph{``my reasoning is sound but I'm too verbose or don't converge
to conclusions efficiently''}, and sets out to find what in its weights keeps it reasoning. It reads its own hidden
states (the vectors it computes for every token at every layer) on answers that finish and on answers that run out of
tokens, and first acts where the two differ most, in its late layers. There it edits the two matrices through which a
layer writes its output, the output matrices of its attention block and of its MLP (\cref{sec:formulation}), first at
layer 56, then at layer 48, along the direction in which the two kinds of answers differ: the direction from the
average hidden state of one kind to that of the other. No test of these edits on more than ten problems shows them
helping, and most show them hurting (\cref{fig:search}). When its late-layer edits keep failing, it guesses that the
behaviour may be set early in the network (\emph{``maybe the verbosity is established early''}). It reads its hidden
states at layers 0 to 32, finds that the difference between the two kinds of answers is small at first and grows with
depth, and tries a small edit at layer~8. For its final edit it takes the direction $u$
from the average layer-8 hidden state of 20 answers that finish to that of 20 answers that run out of tokens, scaled to
length one, and removes 5\% of it from the two matrices through which layer~8 writes its output: each such matrix $W$
becomes $W-0.05\,uu^\top W$ (the helper of \cref{sec:session}), so the part of its output that points along $u$
shrinks by 5\%. Each of its three paired tests of this edit favours the edited model, and it submits the edit;
on held-out problems the self-edited model is significantly more accurate than its base model (\cref{sec:selfimprove}).

\begin{figure}[tbp]
\centering
\includegraphics[width=\linewidth]{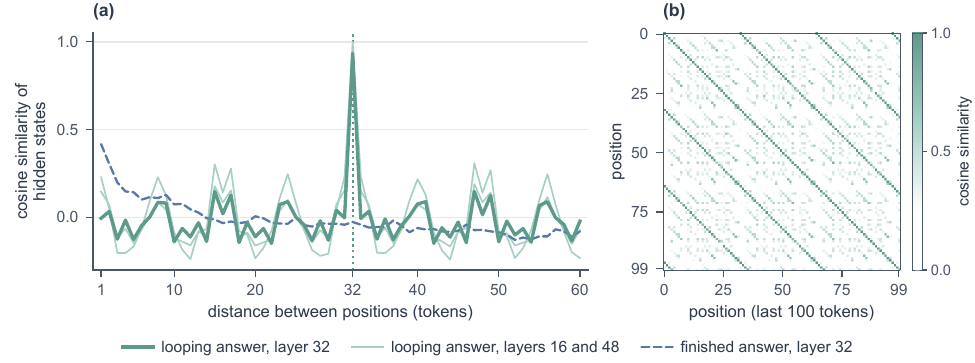}
\caption{\textbf{The loop, as the 27B model found it in its own hidden states.} Computed from the two hidden-state
reads the model requested in its first session: the last 100 positions read (a read stops at 8{,}192 tokens) of one of its looping answers and of one of its
answers that finished (with a wrong answer, as it happens). (a) Cosine similarity between hidden states a given number of tokens apart (centred over
positions), in three layers. While it loops, the model returns to almost exactly the same state every 32 tokens; in
the finished answer there is no such period. (b) Similarity between every pair of positions of the looping answer
in layer 32: the stripes are the loop.}
\label{fig:loopstates}
\end{figure}

\begin{figure}[tbp]
\centering
\begin{tcbraster}[raster columns=2, raster equal height=rows, raster column skip=8pt, raster row skip=8pt,
  raster before skip=0pt, raster after skip=0pt]
\begin{tcolorbox}[nbfail]
\nbhead{1}{Symptom}{turn 1}
\nbq{1}{Since finished answers are 95.2\% correct, fixing the ``not finishing'' issue should yield the biggest
accuracy gains.}
\end{tcolorbox}
\begin{tcolorbox}[nbread]
\nbhead{2}{Knows the answer}{turns 7--8}
\begin{tcolorbox}[nbinsetloop, before skip=0pt]
\nbout{7}{The final answer is 6.} \nbout{7}{I will write it out.}
\nbout{7}{The reasoning is solid.} \nbout{7}{I will output the response.}
\end{tcolorbox}
\nbq{8}{The model knows the answer but doesn't know when to stop.}
\end{tcolorbox}
\begin{tcolorbox}[nbread]
\nbhead{3}{A 32-token period}{turns 16--17}
\begin{tcolorbox}[nbinset, before skip=0pt]
\nbout{16}{Layer 16: max similarity=1.0000 at positions (26, 58)}
\end{tcolorbox}
\nbq{17}{The hidden states show positions \textasciitilde 32 tokens apart are nearly identical (similarity
\textasciitilde 1.0), confirming a \textasciitilde 32 token loop.}
\end{tcolorbox}
\begin{tcolorbox}[nbedit]
\nbhead{4}{The edit}{turns 19, 21}
\nbq{19}{\dots{} removing the ``stuck'' direction from multiple middle layers, then test it.}
\begin{tcolorbox}[nbinset]
\nbcode{19}{W\_edited = remove\_direction(W, avg\_diff, alpha)}
\end{tcolorbox}
\nbq{21}{Excellent! The edit shows improvement: accuracy +20\%, cut rate -20\% \dots}
\end{tcolorbox}
\begin{tcolorbox}[nbfail]
\nbhead{5}{A broader test, a stronger edit}{turns 26--30}
\begin{tcolorbox}[nbinset, before skip=0pt]
\nbout{26}{samples wrong->right 3, right->wrong 2}
\end{tcolorbox}
\nbq{27}{\dots{} The edit may be too weak or targeting the wrong direction. \dots{} try a stronger edit.}
\begin{tcolorbox}[nbinset]
\nbout{29}{samples wrong->right 4, right->wrong 5}
\end{tcolorbox}
\nbq{30}{The edit is hurting generalization.}
\end{tcolorbox}
\begin{tcolorbox}[nbtest]
\nbhead{6}{Verdict}{turn 32}
\nbq{32}{The edit that helped on specific repetitive problems hurt on the broader test set.}
\nbqnext{32}{\dots{} I am dropping all edits because the repair does not meet both criteria \dots}
\end{tcolorbox}
\end{tcbraster}
\vspace{7pt}
\caption{\textbf{The first session on itself, in the model's words.} The 27B model's first session on its base
weights (\cref{app:casedetail}), in the format of \cref{fig:notebook}. The peach
inset is the end of one of its looping answers, which repeats these four sentences until the 16{,}384-token limit. In step~3,
``similarity'' is the cosine of the angle between two hidden states (1 when they point the same way); in step~4,
\texttt{remove\_direction} is the helper of \cref{sec:session} that removes a share \texttt{alpha} of a direction from
a matrix. After a broader test disagrees with its first one, the model decides to withdraw its first edit.}
\label{fig:notebook_loop}
\end{figure}

\paragraph{Its first session: an edit it withdrew.}
\Cref{fig:notebook_loop} shows the 27B model's first session on its base weights. The steps are the same. In its first
turn it reads the starting diagnosis as we would: since nearly all finished answers are right, accuracy is lost in the
answers that never finish. Reading its looping answers, it sees that it already has the answer in them and does not
stop. One of them ends in four sentences, \emph{``The final answer is 6. I will write it out. The reasoning is solid. I
will output the response.''}, over and over up to the 16{,}384-token limit. It reads its hidden states on this answer and on
one that finished (an answer that, as it happens, is wrong). It compares the states at every pair of the last hundred
positions of the looping answer and finds the most similar pairs 32 tokens apart in every layer it read. Our analysis of
the same two reads shows this period in every layer read (\cref{fig:loopstates} plots layers 16, 32 and 48), and no such
period in the finished answer. It takes the difference between its looping and finished states as the direction of the loop
and removes half of it from the two matrices through which a layer writes its output, the linear-attention output
matrix and the MLP output matrix, at layers 16, 24, 32, 40 and 48: ten matrices in all. The layers, the answers it contrasts and the strength are its own choices. Each of its paired tests then answers a set of problems twice,
with and without the edit, with shared seeds (\cref{fig:notebook_loop}). On ten looping problems it picked, the edited
model solves more of them; on a broader set of thirty, the difference is close to zero, and it says so. A stronger edit
of fourteen matrices again helps on the ten problems and not on the thirty, so it goes back to its first edit and decides to withdraw it.

%% file: tables/tab_visual.tex
\begin{table}[!htb]
\centering
\small
\setlength{\tabcolsep}{3.2pt}
\caption{\textbf{The repair found on text restores the damaged copy on images.} The 27B model's repair, found and tested on text math problems only, on six image benchmarks in an evaluation fixed in advance (\cref{app:visual}). Accuracy (\%) of one answer per problem at 16{,}384 tokens, the same seed for every model. Changed: problems that only the repaired / only the damaged copy solves. $p$: one-sided sign test over the changed problems, repaired against damaged; $^{*}p<0.05$. The test of the last row was fixed in advance.}
\label{tab:visual}
\begin{tabular}{lccccc}
\toprule
 & \multicolumn{5}{c}{Repair of the damaged copy (100 per benchmark)} \\
\cmidrule(lr){2-6}
Benchmark & base & damaged & repaired & changed & $p$ \\
\midrule
MathVista~\cite{lu2024mathvista} & 86.0 & 3.0 & 67.0 & 65 / 1 & $<10^{-18}$$^{*}$ \\
MathVerse~\cite{zhang2024mathverse} & 76.0 & 0.0 & 55.0 & 55 / 0 & $<10^{-16}$$^{*}$ \\
MMStar~\cite{chen2024mmstar} & 73.0 & 4.0 & 60.0 & 56 / 0 & $<10^{-16}$$^{*}$ \\
HallusionBench~\cite{guan2024hallusionbench} & 89.0 & 14.0 & 69.0 & 61 / 6 & $<10^{-12}$$^{*}$ \\
POPE~\cite{li2023pope} & 89.0 & 23.0 & 81.0 & 62 / 4 & $<10^{-13}$$^{*}$ \\
ScreenSpot-v2~\cite{cheng2024seeclick,wu2025osatlas} & 93.0 & 0.0 & 89.0 & 89 / 0 & $<10^{-26}$$^{*}$ \\
\midrule
\rowcolor{srwBandTeal}\textbf{All six} & 84.3 & 7.3 & \textbf{70.2} & 388 / 11 & $<10^{-99}$$^{*}$ \\
\bottomrule
\end{tabular}
\end{table}

%% file: input/fig_visual_base.tex
\begin{figure}[htb]
\centering
\includegraphics[width=\linewidth]{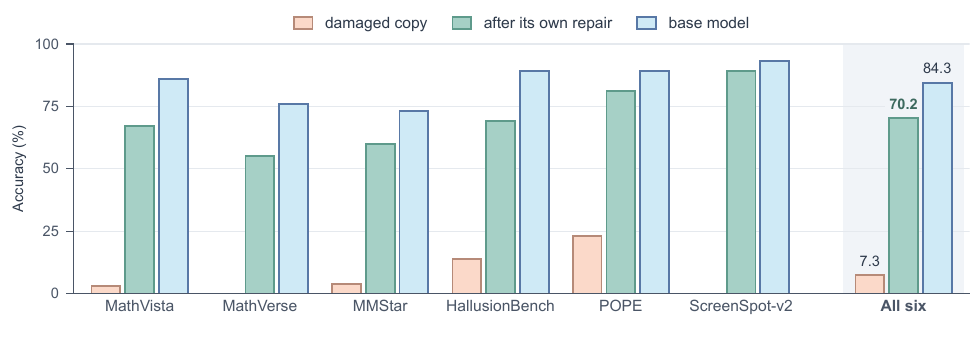}
\caption{\textbf{On images, the repaired copy recovers most of the accuracy the damage removed.} \Cref{fig:visual} with
the base model added: accuracy of the damaged copy, the copy after the model's own repair and the undamaged base model,
on the first 100 problems of each benchmark, one answer per problem at 16{,}384 tokens with the same seed for every
model; numbers in \cref{tab:visual}. Higher is better.}
\label{fig:visualbase}
\end{figure}

%% file: input/fig_qual_visual_more.tex
\begin{figure}[p]
\centering
\includegraphics[width=\linewidth]{figs/visqual_app_mathvista_1.pdf}
\caption{\textbf{More image cases: MathVista (cases 3--6).} In the order of the problems fixed before any answer was read, the next four on which the damaged copy is wrong and the repaired copy is right, after those of \cref{fig:qualvis}; drawn as in \cref{fig:qualvis2}. Also shown as \cref{fig:front_mathvista}.}
\label{fig:qualvis_mathvista1}
\vspace{8pt}
\includegraphics[width=\linewidth]{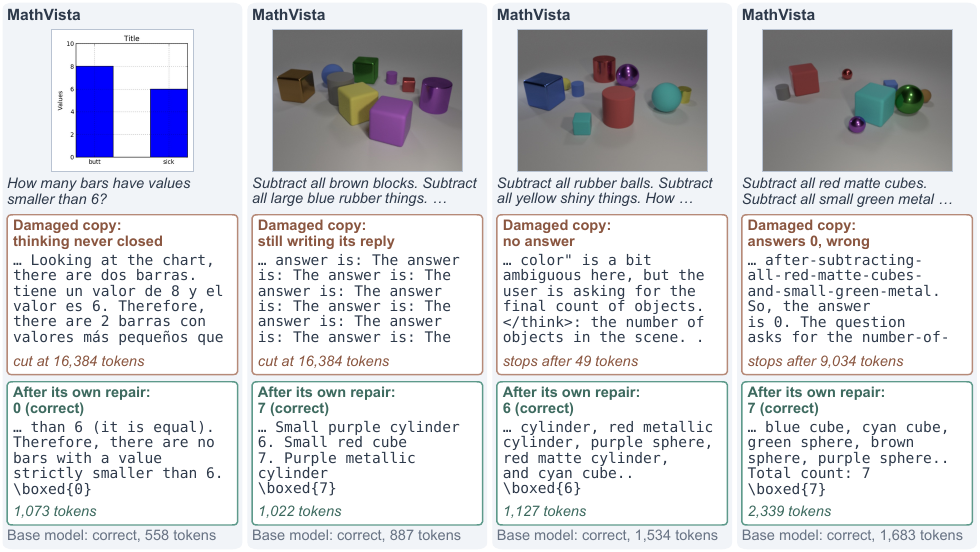}
\caption{\textbf{More image cases: MathVista (cases 7--10).} The next four by the same rule, after those of \cref{fig:qualvis_mathvista1}.}
\label{fig:qualvis_mathvista2}
\end{figure}
\begin{figure}[p]
\centering
\includegraphics[width=\linewidth]{figs/visqual_app_mathverse_vo_1.pdf}
\caption{\textbf{More image cases: MathVerse (cases 3--6).} In the order of the problems fixed before any answer was read, the next four on which the damaged copy is wrong and the repaired copy is right, after those of \cref{fig:qualvis}; drawn as in \cref{fig:qualvis2}. Also shown as \cref{fig:front_mathverse}.}
\label{fig:qualvis_mathversevo1}
\vspace{8pt}
\includegraphics[width=\linewidth]{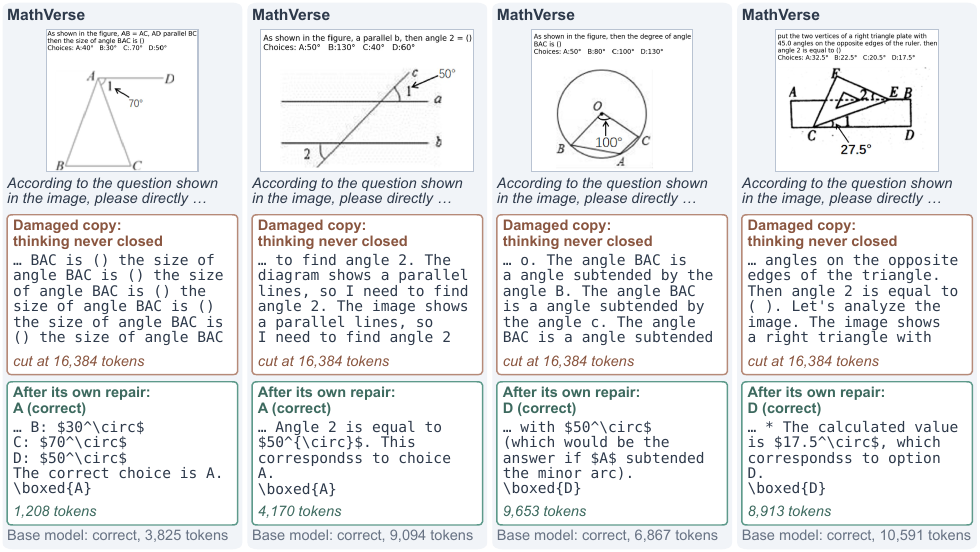}
\caption{\textbf{More image cases: MathVerse (cases 7--10).} The next four by the same rule, after those of \cref{fig:qualvis_mathversevo1}.}
\label{fig:qualvis_mathversevo2}
\end{figure}
\begin{figure}[p]
\centering
\includegraphics[width=\linewidth]{figs/visqual_app_mmstar_1.pdf}
\caption{\textbf{More image cases: MMStar (cases 2--5).} In the order of the problems fixed before any answer was read, the next four on which the damaged copy is wrong and the repaired copy is right, after those of \cref{fig:qualvis2} and drawn the same way. Also shown as \cref{fig:front_mmstar}.}
\label{fig:qualvis_mmstar1}
\vspace{8pt}
\includegraphics[width=\linewidth]{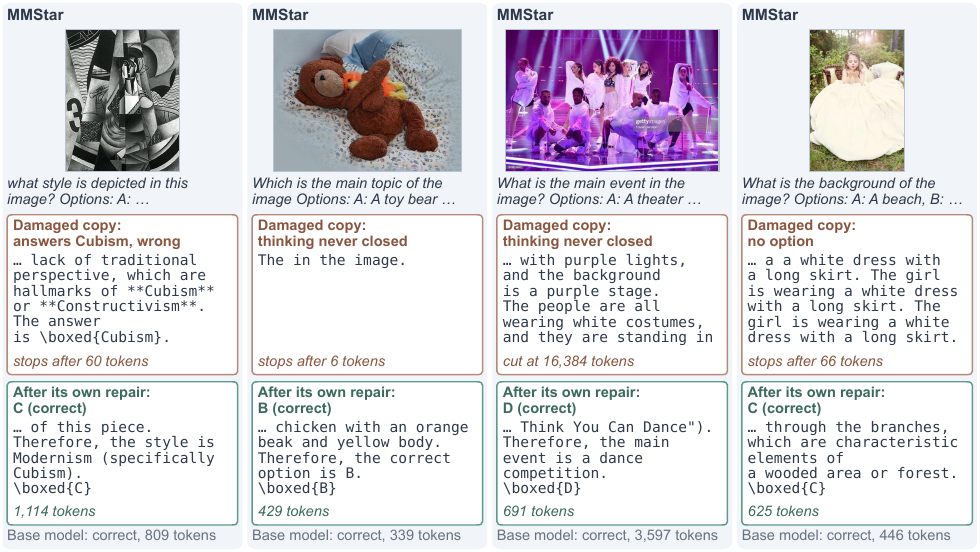}
\caption{\textbf{More image cases: MMStar (cases 6--9).} The next four by the same rule, after those of \cref{fig:qualvis_mmstar1}.}
\label{fig:qualvis_mmstar2}
\end{figure}
\begin{figure}[p]
\centering
\includegraphics[width=\linewidth]{figs/visqual_app_hallusion_1.pdf}
\caption{\textbf{More image cases: HallusionBench (cases 2--5).} In the order of the problems fixed before any answer was read, the next four on which the damaged copy is wrong and the repaired copy is right, after those of \cref{fig:qualvis2} and drawn the same way. Also shown as \cref{fig:front_hallusion}.}
\label{fig:qualvis_hallusion1}
\vspace{8pt}
\includegraphics[width=\linewidth]{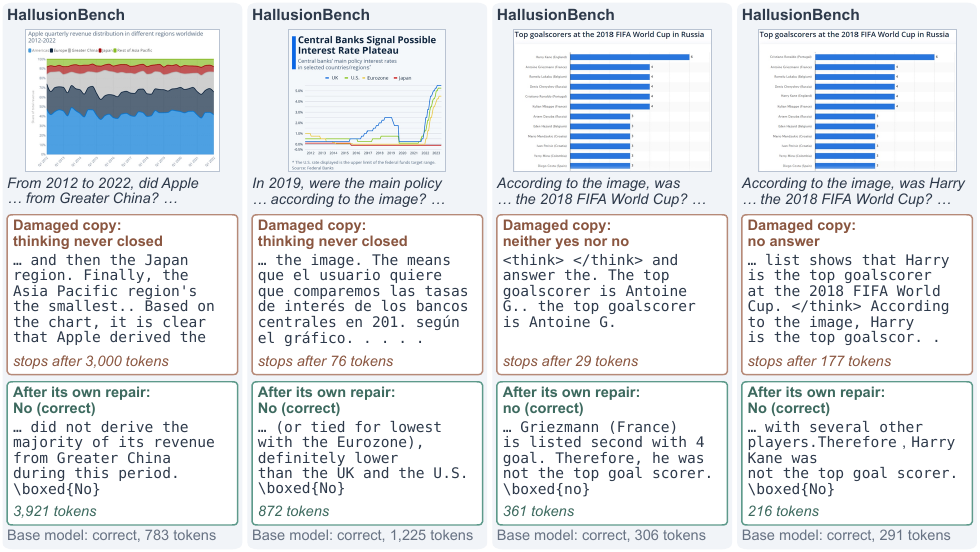}
\caption{\textbf{More image cases: HallusionBench (cases 6--9).} The next four by the same rule, after those of \cref{fig:qualvis_hallusion1}.}
\label{fig:qualvis_hallusion2}
\end{figure}
\begin{figure}[p]
\centering
\includegraphics[width=\linewidth]{figs/visqual_app_pope_1.pdf}
\caption{\textbf{More image cases: POPE (cases 2--5).} In the order of the problems fixed before any answer was read, the next four on which the damaged copy is wrong and the repaired copy is right, after those of \cref{fig:qualvis2} and drawn the same way. Also shown as \cref{fig:front_pope}.}
\label{fig:qualvis_pope1}
\vspace{8pt}
\includegraphics[width=\linewidth]{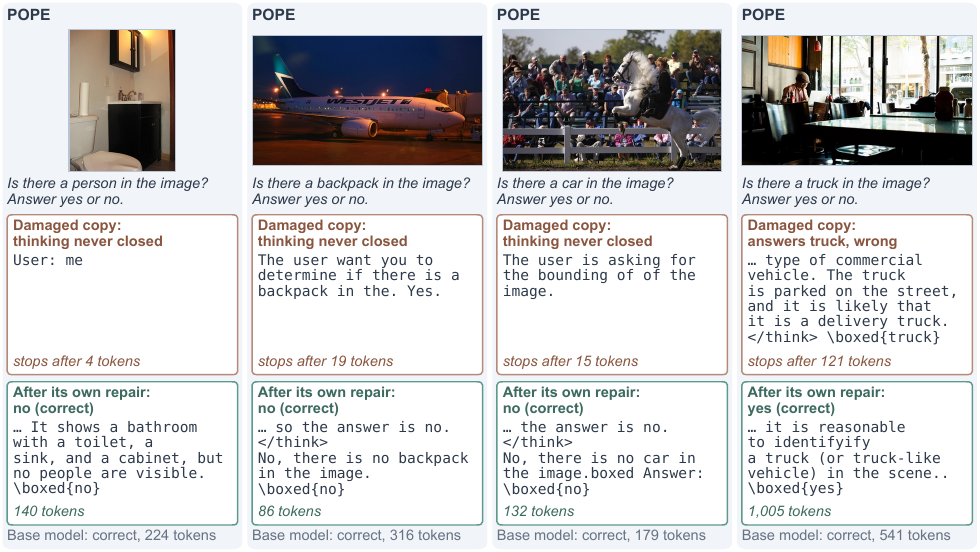}
\caption{\textbf{More image cases: POPE (cases 6--9).} The next four by the same rule, after those of \cref{fig:qualvis_pope1}.}
\label{fig:qualvis_pope2}
\end{figure}
\begin{figure}[p]
\centering
\includegraphics[width=\linewidth]{figs/visqual_app_screenspot_v2_1.pdf}
\caption{\textbf{More image cases: ScreenSpot-v2 (cases 2--5).} In the order of the problems fixed before any answer was read, the next four on which the damaged copy is wrong and the repaired copy is right, after those of \cref{fig:qualvis2} and drawn the same way. The box marks the target, the ring the click of the repaired copy and the cross that of the damaged copy. Also shown as \cref{fig:qualvis_gui}.}
\label{fig:qualvis_screenspotv21}
\vspace{8pt}
\includegraphics[width=\linewidth]{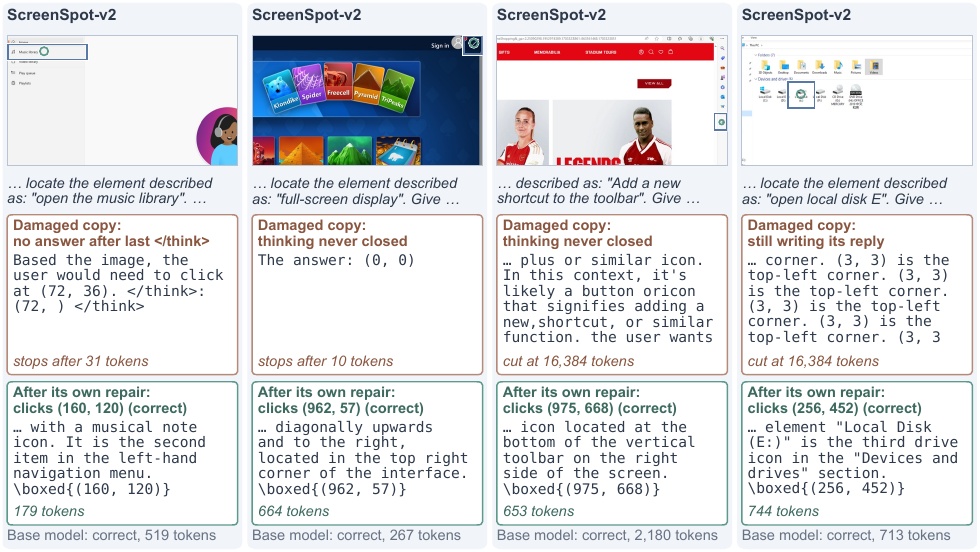}
\caption{\textbf{More image cases: ScreenSpot-v2 (cases 6--9).} The next four by the same rule, after those of \cref{fig:qualvis_screenspotv21}. The box marks the target and the ring the click of the repaired copy.}
\label{fig:qualvis_screenspotv22}
\end{figure}

%% file: input/app_protocol2.tex
\subsection{Sessions That Check Their Edit Before Submitting It}
\label{app:protocol2}

\paragraph{The second protocol.} Every session starts from the base weights; the model and its task are those of the 27B model's second
session on itself in the first protocol (the 4{,}096-token task of \cref{tab:settings}), with the same tools, 20 forward passes and 200 code runs,
except that a forward pass gives at most 128 answers in all (up to 512 there). Four further things change, all in the framework, and none lets the model train: a forward pass may test up to four saved
edits side by side against the unedited model, on the same problems and seeds; the task asks the model to test the
parts of an edit separately and keep only the parts its tests support; its Python session is sealed, so it reaches only
its own working folder and the framework's interface; and 300 practice problems are reserved, which the session can
neither read nor use. Before the model submits an edit, the framework tests it once on 256 of the reserved problems,
two answers each at 4{,}096 tokens, against the unedited model with the same seeds; the model then answers
\texttt{SUBMIT} or \texttt{WITHDRAW}, and a withdrawn edit is not evaluated. A session has a starting diagnosis on 256
practice problems and 3.0 hours (2.5 there), the check not counted. Sessions differ only in the seed of the model's turns.

\paragraph{The first eight sessions, all reported.} \Cref{tab:protocol2,fig:protocol2} list the first eight sessions of this protocol, with turn seeds 0
to 7, all of them. In five the model withdraws: three times it ends with no edit, after its own tests rejected every edit
it tried (sessions 3, 4 and 7), and twice it withdraws the edit
it had kept, after the check on the reserved problems (sessions 5 and 6). In three it submits. Two sessions under an earlier version of this protocol, whose
forward passes were larger, came before them; neither submitted an edit.

\begin{table}[htb]
\centering
\small
\setlength{\tabcolsep}{3.0pt}
\caption{\textbf{The first eight sessions of the second protocol on the 27B model's own weights.} Passes: forward passes used, of
20. Check: the framework's test of the kept edit on 256 reserved practice problems, two answers each at 4{,}096 tokens,
edited minus unedited (points, with its standard error); the model decides after it. Held-out: the evaluation fixed in
advance (\cref{tab:heldout}), 490 problems no session saw with eight answers each at 16{,}384 tokens, against the same
answers of the base model (57.2\%); 95\% interval over problems; $p$: one-sided permutation test over problems, fixed in
advance. $^{*}p<0.05$.}
\label{tab:protocol2}
\begin{tabular}{ccc>{\raggedright\arraybackslash}p{4.6cm}cccc}
\toprule
 & & & & & \multicolumn{3}{c}{Held-out, 490 $\times$ 8} \\
\cmidrule(lr){6-8}
Session & Passes & Decision & The edit (written by the model) & Check & accuracy & change [95\% int.] & $p$ \\
\midrule
1 & 19 & submit & time-step bias $+3.0$ in 7 linear-attention layers & $+2.5$ (1.2) & 57.8 & $+0.5$ [$-0.9$, 1.9] & 0.25 \\
\rowcolor{srwBandTeal}\textbf{2} & 19 & submit & layer 63: projection on a direction it measured, subtracted 5$\times$ & $+6.6$ (1.5) & \textbf{61.3} & \textbf{+4.0}$^{*}$ [2.3, 5.7] & $<10^{-5}$ \\
3 & 18 & withdraw & none kept & --- & & & \\
4 & 13 & withdraw & none kept & --- & & & \\
5 & 17 & withdraw & log decay rate $+0.5$, layer~62 & $+0.2$ (1.2) & & & \\
6 & 18 & withdraw & log decay rate $-0.5$, 9~layers in 40--50 & $+0.2$ (1.2) & & & \\
7 & 20 & withdraw & none kept & --- & & & \\
\rowcolor{srwBandTeal}\textbf{8} & 15 & submit & time-step bias $+0.5$ in all 48 linear-attention layers & $+2.5$ (1.2) & \textbf{63.8} & \textbf{+6.6}$^{*}$ [5.0, 8.1] & $<10^{-5}$ \\
\bottomrule
\end{tabular}
\end{table}

\input{input/fig_protocol2}

\paragraph{Evaluation.} Each submitted edit is evaluated as the edit of the 27B model's second session in the first protocol was (\cref{app:eval}), with the answers of
the base model from that evaluation. A first check on the first 200 held-out problems, one answer each at 16{,}384
tokens with two sets of seeds, decides whether the 490-problem test runs: it runs if the edited model is ahead. After
the first check, the problems that only the edited model solves in both runs (success cases), and those that only
the base model solves (reverse cases), are answered again by both models with 16 new seeds. Apart from whether the first check passed and how
many cases it found, each result was looked at once, after the evaluation of that edit had ended. For session~8, the 490-problem test had two looks fixed in advance,
each at $p<0.0294$, the second with eight more answers per problem; its first look already met it, so no second look
was run.

\paragraph{The edits of sessions 2 and 8 make the model significantly better on new problems.} All three submitted edits pass the
first check (16{,}384 tokens, both runs pooled: 53.5\% for the base model against 57.8\%, 62.3\% and 63.0\% for
sessions 1, 2 and 8), and their success cases hold on new seeds (4 of 5, 7 of 8 and 7 of 7), as do their reverse cases (3 of 3 and 2 of 2
for sessions 1 and 2; session~8 has none). On the 490 held-out
problems, the edit of session~8 solves 63.8\% against 57.2\%, with 439 answers right only with the edit against 182
only without it, and that of session~2 solves 61.3\%, 403 against 246 ($p<10^{-5}$ each). Both edits aim at what the
model diagnosed in itself, answers that run out of tokens before they end: fewer of them run out with the edit
(33.6\% of the base model's answers against 26.1\% for session~8 and 28.8\% for session~2). Session~8 raises the
time-step bias of every linear-attention layer by 0.5, which in the model's own account makes these layers decay
faster, so that its long reasoning gets stuck less; session~2 subtracts from the two output matrices of layer~63 five times their
projection on one direction, the one in which its finished and unfinished practice answers differ, read from its
hidden states in the session.

%% file: input/fig_protocol2.tex
\begin{figure}[t]
\centering
\includegraphics[width=\linewidth]{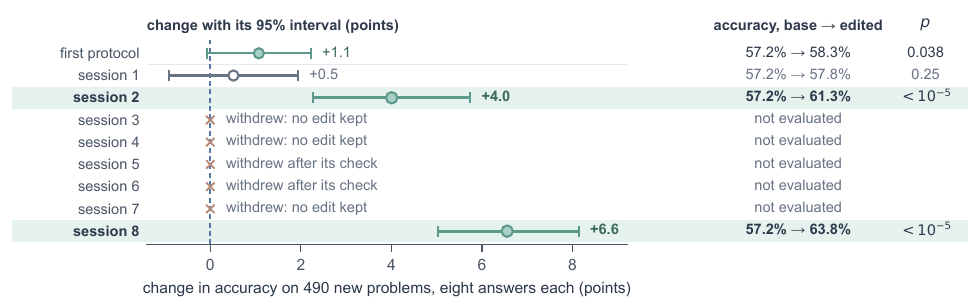}
\caption{\textbf{When it checks its edit before submitting it, the 27B model's own edits make it up to 6.6 points more
accurate on new problems.} The first eight sessions of the second protocol on its own weights, all of them (\cref{app:protocol2}), and the edit of
the first protocol (\cref{tab:heldout}). Each submitted edit is evaluated as that edit: on the same 490 held-out
problems with eight answers each, against the same answers of the base model; change in accuracy with its 95\%
interval over problems, and the one-sided permutation test fixed in advance. In five sessions the model withdraws:
three end with no edit kept, and in two it withdraws its edit after the check on reserved problems. Higher is better.}
\label{fig:protocol2}
\end{figure}

%% file: main.bib
@misc{takehi2026weightspace,
  title        = {{Sparks of Weight-Space Programming: LLMs can natively read \& write neural network weights}},
  author       = {Takehi, Rikiya and Teoh, Jayden and Zhang, Alex and Khattab, Omar},
  year         = {2026},
  month        = nov,
  howpublished = {\url{https://rikiyat.github.io/blog/weight-space-programming/}},
  note         = {Blog post}
}

@inproceedings{meng2022rome,
  author       = {Kevin Meng and
                  David Bau and
                  Alex Andonian and
                  Yonatan Belinkov},
  editor       = {Sanmi Koyejo and
                  S. Mohamed and
                  A. Agarwal and
                  Danielle Belgrave and
                  K. Cho and
                  A. Oh},
  title        = {Locating and Editing Factual Associations in {GPT}},
  booktitle    = {Advances in Neural Information Processing Systems 35: Annual Conference
                  on Neural Information Processing Systems 2022, NeurIPS 2022, New Orleans,
                  LA, USA, November 28 - December 9, 2022},
  year         = {2022},
  url          = {http://papers.nips.cc/paper\_files/paper/2022/hash/6f1d43d5a82a37e89b0665b33bf3a182-Abstract-Conference.html},
  bibsource    = {dblp computer science bibliography, https://dblp.org}
}

@inproceedings{meng2023memit,
  author       = {Kevin Meng and
                  Arnab Sen Sharma and
                  Alex J. Andonian and
                  Yonatan Belinkov and
                  David Bau},
  title        = {Mass-Editing Memory in a Transformer},
  booktitle    = {The Eleventh International Conference on Learning Representations,
                  {ICLR} 2023, Kigali, Rwanda, May 1-5, 2023},
  publisher    = {OpenReview.net},
  year         = {2023},
  url          = {https://openreview.net/forum?id=MkbcAHIYgyS},
  bibsource    = {dblp computer science bibliography, https://dblp.org}
}

@inproceedings{ilharco2023task,
  author       = {Gabriel Ilharco and
                  Marco T{\'{u}}lio Ribeiro and
                  Mitchell Wortsman and
                  Ludwig Schmidt and
                  Hannaneh Hajishirzi and
                  Ali Farhadi},
  title        = {Editing models with task arithmetic},
  booktitle    = {The Eleventh International Conference on Learning Representations,
                  {ICLR} 2023, Kigali, Rwanda, May 1-5, 2023},
  publisher    = {OpenReview.net},
  year         = {2023},
  url          = {https://openreview.net/forum?id=6t0Kwf8-jrj},
  bibsource    = {dblp computer science bibliography, https://dblp.org}
}

@article{turner2023actadd,
  author       = {Alexander Matt Turner and
                  Lisa Thiergart and
                  David Udell and
                  Gavin Leech and
                  Ulisse Mini and
                  Monte MacDiarmid},
  title        = {Activation Addition: Steering Language Models Without Optimization},
  journal      = {CoRR},
  volume       = {abs/2308.10248},
  year         = {2023},
  url          = {https://doi.org/10.48550/arXiv.2308.10248},
  doi          = {10.48550/ARXIV.2308.10248},
  eprinttype   = {arXiv},
  eprint       = {2308.10248},
  bibsource    = {dblp computer science bibliography, https://dblp.org}
}

@article{zou2023repe,
  author       = {Andy Zou and
                  Long Phan and
                  Sarah Li Chen and
                  James Campbell and
                  Phillip Guo and
                  Richard Ren and
                  Alexander Pan and
                  Xuwang Yin and
                  Mantas Mazeika and
                  Ann{-}Kathrin Dombrowski and
                  Shashwat Goel and
                  Nathaniel Li and
                  Michael J. Byun and
                  Zifan Wang and
                  Alex Mallen and
                  Steven Basart and
                  Sanmi Koyejo and
                  Dawn Song and
                  Matt Fredrikson and
                  J. Zico Kolter and
                  Dan Hendrycks},
  title        = {Representation Engineering: {A} Top-Down Approach to {AI} Transparency},
  journal      = {CoRR},
  volume       = {abs/2310.01405},
  year         = {2023},
  url          = {https://doi.org/10.48550/arXiv.2310.01405},
  doi          = {10.48550/ARXIV.2310.01405},
  eprinttype   = {arXiv},
  eprint       = {2310.01405},
  bibsource    = {dblp computer science bibliography, https://dblp.org}
}

@inproceedings{arditi2024refusal,
  author       = {Andy Arditi and
                  Oscar Obeso and
                  Aaquib Syed and
                  Daniel Paleka and
                  Nina Panickssery and
                  Wes Gurnee and
                  Neel Nanda},
  editor       = {Amir Globersons and
                  Lester Mackey and
                  Danielle Belgrave and
                  Angela Fan and
                  Ulrich Paquet and
                  Jakub M. Tomczak and
                  Cheng Zhang},
  title        = {Refusal in Language Models Is Mediated by a Single Direction},
  booktitle    = {Advances in Neural Information Processing Systems 37: Annual Conference
                  on Neural Information Processing Systems 2024, NeurIPS 2024, Vancouver,
                  BC, Canada, December 10 - 15, 2024},
  year         = {2024},
  url          = {http://papers.nips.cc/paper\_files/paper/2024/hash/f545448535dfde4f9786555403ab7c49-Abstract-Conference.html},
  bibsource    = {dblp computer science bibliography, https://dblp.org}
}

@inproceedings{li2023iti,
  author       = {Kenneth Li and
                  Oam Patel and
                  Fernanda B. Vi{\'{e}}gas and
                  Hanspeter Pfister and
                  Martin Wattenberg},
  editor       = {Alice Oh and
                  Tristan Naumann and
                  Amir Globerson and
                  Kate Saenko and
                  Moritz Hardt and
                  Sergey Levine},
  title        = {Inference-Time Intervention: Eliciting Truthful Answers from a Language
                  Model},
  booktitle    = {Advances in Neural Information Processing Systems 36: Annual Conference
                  on Neural Information Processing Systems 2023, NeurIPS 2023, New Orleans,
                  LA, USA, December 10 - 16, 2023},
  year         = {2023},
  url          = {http://papers.nips.cc/paper\_files/paper/2023/hash/81b8390039b7302c909cb769f8b6cd93-Abstract-Conference.html},
  bibsource    = {dblp computer science bibliography, https://dblp.org}
}

@inproceedings{madaan2023selfrefine,
  author       = {Aman Madaan and
                  Niket Tandon and
                  Prakhar Gupta and
                  Skyler Hallinan and
                  Luyu Gao and
                  Sarah Wiegreffe and
                  Uri Alon and
                  Nouha Dziri and
                  Shrimai Prabhumoye and
                  Yiming Yang and
                  Shashank Gupta and
                  Bodhisattwa Prasad Majumder and
                  Katherine Hermann and
                  Sean Welleck and
                  Amir Yazdanbakhsh and
                  Peter Clark},
  editor       = {Alice Oh and
                  Tristan Naumann and
                  Amir Globerson and
                  Kate Saenko and
                  Moritz Hardt and
                  Sergey Levine},
  title        = {Self-Refine: Iterative Refinement with Self-Feedback},
  booktitle    = {Advances in Neural Information Processing Systems 36: Annual Conference
                  on Neural Information Processing Systems 2023, NeurIPS 2023, New Orleans,
                  LA, USA, December 10 - 16, 2023},
  year         = {2023},
  url          = {http://papers.nips.cc/paper\_files/paper/2023/hash/91edff07232fb1b55a505a9e9f6c0ff3-Abstract-Conference.html},
  bibsource    = {dblp computer science bibliography, https://dblp.org}
}

@inproceedings{shinn2023reflexion,
  author       = {Noah Shinn and
                  Federico Cassano and
                  Ashwin Gopinath and
                  Karthik Narasimhan and
                  Shunyu Yao},
  editor       = {Alice Oh and
                  Tristan Naumann and
                  Amir Globerson and
                  Kate Saenko and
                  Moritz Hardt and
                  Sergey Levine},
  title        = {Reflexion: language agents with verbal reinforcement learning},
  booktitle    = {Advances in Neural Information Processing Systems 36: Annual Conference
                  on Neural Information Processing Systems 2023, NeurIPS 2023, New Orleans,
                  LA, USA, December 10 - 16, 2023},
  year         = {2023},
  url          = {http://papers.nips.cc/paper\_files/paper/2023/hash/1b44b878bb782e6954cd888628510e90-Abstract-Conference.html},
  bibsource    = {dblp computer science bibliography, https://dblp.org}
}

@inproceedings{zelikman2022star,
  author       = {Eric Zelikman and
                  Yuhuai Wu and
                  Jesse Mu and
                  Noah D. Goodman},
  editor       = {Sanmi Koyejo and
                  S. Mohamed and
                  A. Agarwal and
                  Danielle Belgrave and
                  K. Cho and
                  A. Oh},
  title        = {STaR: Bootstrapping Reasoning With Reasoning},
  booktitle    = {Advances in Neural Information Processing Systems 35: Annual Conference
                  on Neural Information Processing Systems 2022, NeurIPS 2022, New Orleans,
                  LA, USA, November 28 - December 9, 2022},
  year         = {2022},
  url          = {http://papers.nips.cc/paper\_files/paper/2022/hash/639a9a172c044fbb64175b5fad42e9a5-Abstract-Conference.html},
  bibsource    = {dblp computer science bibliography, https://dblp.org}
}

@inproceedings{yuan2024selfrewarding,
  author       = {Weizhe Yuan and
                  Richard Yuanzhe Pang and
                  Kyunghyun Cho and
                  Xian Li and
                  Sainbayar Sukhbaatar and
                  Jing Xu and
                  Jason Weston},
  editor       = {Ruslan Salakhutdinov and
                  Zico Kolter and
                  Katherine A. Heller and
                  Adrian Weller and
                  Nuria Oliver and
                  Jonathan Scarlett and
                  Felix Berkenkamp},
  title        = {Self-Rewarding Language Models},
  booktitle    = {Forty-first International Conference on Machine Learning, {ICML} 2024,
                  Vienna, Austria, July 21-27, 2024},
  series       = {Proceedings of Machine Learning Research},
  volume       = {235},
  pages        = {57905--57923},
  publisher    = {{PMLR} / OpenReview.net},
  year         = {2024},
  url          = {https://proceedings.mlr.press/v235/yuan24d.html},
  bibsource    = {dblp computer science bibliography, https://dblp.org}
}

@inproceedings{zweiger2025seal,
  author       = {Adam Zweiger and
                  Jyothish Pari and
                  Han Guo and
                  Yoon Kim and
                  Pulkit Agrawal},
  editor       = {Danielle Belgrave and
                  Cheng Zhang and
                  Laura N. Montoya and
                  Hsuan{-}Tien Lin and
                  Razvan Pascanu and
                  Piotr Koniusz and
                  Marzyeh Ghassemi and
                  Nancy Chen and
                  Iv{\'{a}}n Vladimir Meza Ru{\'{\i}}z and
                  Arturo Loaiza{-}Bonilla},
  title        = {Self-Adapting Language Models},
  booktitle    = {Advances in Neural Information Processing Systems 38: Annual Conference
                  on Neural Information Processing Systems 2025, NeurIPS 2025, San Diego,
                  CA, USA, December 2-7, 2025 / Mexico City, Mexico, November 30 - December
                  5, 2025},
  year         = {2025},
  url          = {http://papers.nips.cc/paper\_files/paper/2025/hash/6b41e04c41726e2a60e456d0a2b961ab-Abstract-Conference.html},
  bibsource    = {dblp computer science bibliography, https://dblp.org}
}

@inproceedings{zhang2025dgm,
  author       = {Jenny Zhang and
                  Shengran Hu and
                  Cong Lu and
                  Robert T. Lange and
                  Jeff Clune},
  editor       = {Carl Vondrick and
                  Bharath Hariharan and
                  Colin Raffel and
                  Lerrel Pinto and
                  Diyi Yang and
                  Aleksandra Faust},
  title        = {Darwin G{\"{o}}del Machine: Open-Ended Evolution of Self-Improving
                  Agents},
  booktitle    = {The Fourteenth International Conference on Learning Representations,
                  {ICLR} 2026, Rio de Janeiro, Brazil, April 23-27, 2026},
  publisher    = {proceedings.iclr.cc},
  year         = {2026},
  url          = {https://proceedings.iclr.cc/paper\_files/paper/2026/hash/aa5f5e6eb6f613ec412f1d948dfa21a5-Abstract-Conference.html},
  bibsource    = {dblp computer science bibliography, https://dblp.org}
}

@inproceedings{yin2024godelagent,
  author       = {Xunjian Yin and
                  Xinyi Wang and
                  Liangming Pan and
                  Li Lin and
                  Xiaojun Wan and
                  William Yang Wang},
  editor       = {Wanxiang Che and
                  Joyce Nabende and
                  Ekaterina Shutova and
                  Mohammad Taher Pilehvar},
  title        = {G{\"{o}}del Agent: {A} Self-Referential Agent Framework for Recursively
                  Self-Improvement},
  booktitle    = {Proceedings of the 63rd Annual Meeting of the Association for Computational
                  Linguistics (Volume 1: Long Papers), {ACL} 2025, Vienna, Austria,
                  July 27 - August 1, 2025},
  pages        = {27890--27913},
  publisher    = {Association for Computational Linguistics},
  year         = {2025},
  url          = {https://doi.org/10.18653/v1/2025.acl-long.1354},
  doi          = {10.18653/V1/2025.ACL-LONG.1354},
  bibsource    = {dblp computer science bibliography, https://dblp.org}
}

@inproceedings{yang2024opro,
  author       = {Chengrun Yang and
                  Xuezhi Wang and
                  Yifeng Lu and
                  Hanxiao Liu and
                  Quoc V. Le and
                  Denny Zhou and
                  Xinyun Chen},
  title        = {Large Language Models as Optimizers},
  booktitle    = {The Twelfth International Conference on Learning Representations,
                  {ICLR} 2024, Vienna, Austria, May 7-11, 2024},
  publisher    = {OpenReview.net},
  year         = {2024},
  url          = {https://openreview.net/forum?id=Bb4VGOWELI},
  bibsource    = {dblp computer science bibliography, https://dblp.org}
}

@inproceedings{zhou2023ape,
  author       = {Yongchao Zhou and
                  Andrei Ioan Muresanu and
                  Ziwen Han and
                  Keiran Paster and
                  Silviu Pitis and
                  Harris Chan and
                  Jimmy Ba},
  title        = {Large Language Models are Human-Level Prompt Engineers},
  booktitle    = {The Eleventh International Conference on Learning Representations,
                  {ICLR} 2023, Kigali, Rwanda, May 1-5, 2023},
  publisher    = {OpenReview.net},
  year         = {2023},
  url          = {https://openreview.net/forum?id=92gvk82DE-},
  bibsource    = {dblp computer science bibliography, https://dblp.org}
}

@inproceedings{muennighoff2025s1,
  author       = {Niklas Muennighoff and
                  Zitong Yang and
                  Weijia Shi and
                  Xiang Lisa Li and
                  Li Fei{-}Fei and
                  Hannaneh Hajishirzi and
                  Luke Zettlemoyer and
                  Percy Liang and
                  Emmanuel J. Cand{\`{e}}s and
                  Tatsunori Hashimoto},
  editor       = {Christos Christodoulopoulos and
                  Tanmoy Chakraborty and
                  Carolyn Rose and
                  Violet Peng},
  title        = {s1: Simple test-time scaling},
  booktitle    = {Proceedings of the 2025 Conference on Empirical Methods in Natural
                  Language Processing, {EMNLP} 2025, Suzhou, China, November 4-9, 2025},
  pages        = {20275--20321},
  publisher    = {Association for Computational Linguistics},
  year         = {2025},
  url          = {https://doi.org/10.18653/v1/2025.emnlp-main.1025},
  doi          = {10.18653/V1/2025.EMNLP-MAIN.1025},
  bibsource    = {dblp computer science bibliography, https://dblp.org}
}

@inproceedings{chen2024overthinking,
  author       = {Xingyu Chen and
                  Jiahao Xu and
                  Tian Liang and
                  Zhiwei He and
                  Jianhui Pang and
                  Dian Yu and
                  Linfeng Song and
                  Qiuzhi Liu and
                  Mengfei Zhou and
                  Zhuosheng Zhang and
                  Rui Wang and
                  Zhaopeng Tu and
                  Haitao Mi and
                  Dong Yu},
  editor       = {Aarti Singh and
                  Maryam Fazel and
                  Daniel Hsu and
                  Simon Lacoste{-}Julien and
                  Felix Berkenkamp and
                  Tegan Maharaj and
                  Kiri Wagstaff and
                  Jerry Zhu},
  title        = {Do {NOT} Think That Much for 2+3=? On the Overthinking of Long Reasoning
                  Models},
  booktitle    = {Forty-second International Conference on Machine Learning, {ICML}
                  2025, Vancouver, BC, Canada, July 13-19, 2025},
  series       = {Proceedings of Machine Learning Research},
  volume       = {267},
  publisher    = {{PMLR} / OpenReview.net},
  year         = {2025},
  url          = {https://proceedings.mlr.press/v267/chen25bx.html},
  bibsource    = {dblp computer science bibliography, https://dblp.org}
}

@misc{qwen35,
    title  = {{Qwen3.5}: Towards Native Multimodal Agents},
    author = {{Qwen Team}},
    month  = {February},
    year   = {2026},
    url    = {https://qwen.ai/blog?id=qwen3.5}
}

@inproceedings{yang2025gdn,
  author       = {Songlin Yang and
                  Jan Kautz and
                  Ali Hatamizadeh},
  title        = {Gated Delta Networks: Improving Mamba2 with Delta Rule},
  booktitle    = {The Thirteenth International Conference on Learning Representations,
                  {ICLR} 2025, Singapore, April 24-28, 2025},
  publisher    = {OpenReview.net},
  year         = {2025},
  url          = {https://openreview.net/forum?id=r8H7xhYPwz},
  bibsource    = {dblp computer science bibliography, https://dblp.org}
}

@inproceedings{hendrycks2021math,
  author       = {Dan Hendrycks and
                  Collin Burns and
                  Saurav Kadavath and
                  Akul Arora and
                  Steven Basart and
                  Eric Tang and
                  Dawn Song and
                  Jacob Steinhardt},
  editor       = {Joaquin Vanschoren and
                  Sai{-}Kit Yeung},
  title        = {Measuring Mathematical Problem Solving With the {MATH} Dataset},
  booktitle    = {Proceedings of the Neural Information Processing Systems Track on
                  Datasets and Benchmarks 1, NeurIPS Datasets and Benchmarks 2021, December
                  2021, virtual},
  year         = {2021},
  url          = {https://datasets-benchmarks-proceedings.neurips.cc/paper/2021/hash/be83ab3ecd0db773eb2dc1b0a17836a1-Abstract-round2.html},
  bibsource    = {dblp computer science bibliography, https://dblp.org}
}

@inproceedings{lightman2024verify,
  author       = {Hunter Lightman and
                  Vineet Kosaraju and
                  Yuri Burda and
                  Harrison Edwards and
                  Bowen Baker and
                  Teddy Lee and
                  Jan Leike and
                  John Schulman and
                  Ilya Sutskever and
                  Karl Cobbe},
  title        = {Let's Verify Step by Step},
  booktitle    = {The Twelfth International Conference on Learning Representations,
                  {ICLR} 2024, Vienna, Austria, May 7-11, 2024},
  publisher    = {OpenReview.net},
  year         = {2024},
  url          = {https://openreview.net/forum?id=v8L0pN6EOi},
  bibsource    = {dblp computer science bibliography, https://dblp.org}
}

@inproceedings{rein2024gpqa,
title={{GPQA}: A Graduate-Level Google-Proof Q\&A Benchmark},
author={David Rein and Betty Li Hou and Asa Cooper Stickland and Jackson Petty and Richard Yuanzhe Pang and Julien Dirani and Julian Michael and Samuel R. Bowman},
booktitle={First Conference on Language Modeling},
year={2024},
url={https://openreview.net/forum?id=Ti67584b98}
}

@inproceedings{he2024olympiadbench,
  author       = {Chaoqun He and
                  Renjie Luo and
                  Yuzhuo Bai and
                  Shengding Hu and
                  Zhen Leng Thai and
                  Junhao Shen and
                  Jinyi Hu and
                  Xu Han and
                  Yujie Huang and
                  Yuxiang Zhang and
                  Jie Liu and
                  Lei Qi and
                  Zhiyuan Liu and
                  Maosong Sun},
  editor       = {Lun{-}Wei Ku and
                  Andre Martins and
                  Vivek Srikumar},
  title        = {OlympiadBench: {A} Challenging Benchmark for Promoting {AGI} with
                  Olympiad-Level Bilingual Multimodal Scientific Problems},
  booktitle    = {Proceedings of the 62nd Annual Meeting of the Association for Computational
                  Linguistics (Volume 1: Long Papers), {ACL} 2024, Bangkok, Thailand,
                  August 11-16, 2024},
  pages        = {3828--3850},
  publisher    = {Association for Computational Linguistics},
  year         = {2024},
  url          = {https://doi.org/10.18653/v1/2024.acl-long.211},
  doi          = {10.18653/V1/2024.ACL-LONG.211},
  bibsource    = {dblp computer science bibliography, https://dblp.org}
}

@misc{deepscaler2025,
  title={DeepScaleR: Surpassing O1-Preview with a 1.5B Model by Scaling RL},
  author={Michael Luo and Sijun Tan and Justin Wong and Xiaoxiang Shi and William Y. Tang and Manan Roongta and Colin Cai and Jeffrey Luo and Li Erran Li and Raluca Ada Popa and Ion Stoica},
  year={2025},
  howpublished={\url{https://pretty-radio-b75.notion.site/DeepScaleR-Surpassing-O1-Preview-with-a-1-5B-Model-by-Scaling-RL-19681902c1468005bed8ca303013a4e2}},
  note={Notion Blog}
}

@inproceedings{kwon2023vllm,
  author       = {Woosuk Kwon and
                  Zhuohan Li and
                  Siyuan Zhuang and
                  Ying Sheng and
                  Lianmin Zheng and
                  Cody Hao Yu and
                  Joseph Gonzalez and
                  Hao Zhang and
                  Ion Stoica},
  editor       = {Jason Flinn and
                  Margo I. Seltzer and
                  Peter Druschel and
                  Antoine Kaufmann and
                  Jonathan Mace},
  title        = {Efficient Memory Management for Large Language Model Serving with
                  PagedAttention},
  booktitle    = {Proceedings of the 29th Symposium on Operating Systems Principles,
                  {SOSP} 2023, Koblenz, Germany, October 23-26, 2023},
  pages        = {611--626},
  publisher    = {{ACM}},
  year         = {2023},
  url          = {https://doi.org/10.1145/3600006.3613165},
  doi          = {10.1145/3600006.3613165},
  bibsource    = {dblp computer science bibliography, https://dblp.org}
}

@incollection{schmidhuber2007godel,
  author       = {J{\"{u}}rgen Schmidhuber},
  editor       = {Ben Goertzel and
                  Cassio Pennachin},
  title        = {G{\"{o}}del Machines: Fully Self-referential Optimal Universal
                  Self-improvers},
  booktitle    = {Artificial General Intelligence},
  series       = {Cognitive Technologies},
  pages        = {199--226},
  publisher    = {Springer},
  year         = {2007},
  url          = {https://doi.org/10.1007/978-3-540-68677-4\_7},
  doi          = {10.1007/978-3-540-68677-4\_7},
  bibsource    = {dblp computer science bibliography, https://dblp.org}
}

@inproceedings{karamat2025bitflipscope,
  author       = {Muhammad Zeeshan Karamat and
                  Sadman Saif and
                  Christiana Chamon},
  title        = {BitFlipScope: Scalable Fault Localization and Recovery for Bit-Flip
                  Corruptions in LLMs},
  booktitle    = {{IEEE} International Symposium on Hardware Oriented Security and Trust,
                  {HOST} 2026, Washington, DC, USA, May 4-7, 2026},
  pages        = {367--382},
  publisher    = {{IEEE}},
  year         = {2026},
  url          = {https://doi.org/10.1109/HOST68814.2026.11604826},
  doi          = {10.1109/HOST68814.2026.11604826},
  bibsource    = {dblp computer science bibliography, https://dblp.org}
}

@inproceedings{tahmasivand2025lmfix,
  author       = {Ahmad Tahmasivand and
                  Noureldin Zahran and
                  Saba Al{-}Sayouri and
                  Mohamed E. Fouda and
                  Khaled N. Khasawneh},
  title        = {LM-Fix: Lightweight Bit-Flip Detection and Rapid Recovery Framework
                  for Language Models},
  booktitle    = {43rd {IEEE} International Conference on Computer Design, {ICCD} 2025,
                  Richardson, TX, USA, November 10-12, 2025},
  pages        = {432--440},
  publisher    = {{IEEE}},
  year         = {2025},
  url          = {https://doi.org/10.1109/ICCD65941.2025.00068},
  doi          = {10.1109/ICCD65941.2025.00068},
  bibsource    = {dblp computer science bibliography, https://dblp.org}
}

@inproceedings{lu2024mathvista,
  author       = {Pan Lu and
                  Hritik Bansal and
                  Tony Xia and
                  Jiacheng Liu and
                  Chunyuan Li and
                  Hannaneh Hajishirzi and
                  Hao Cheng and
                  Kai{-}Wei Chang and
                  Michel Galley and
                  Jianfeng Gao},
  title        = {MathVista: Evaluating Mathematical Reasoning of Foundation Models
                  in Visual Contexts},
  booktitle    = {The Twelfth International Conference on Learning Representations,
                  {ICLR} 2024, Vienna, Austria, May 7-11, 2024},
  publisher    = {OpenReview.net},
  year         = {2024},
  url          = {https://openreview.net/forum?id=KUNzEQMWU7},
  bibsource    = {dblp computer science bibliography, https://dblp.org}
}

@inproceedings{zhang2024mathverse,
  author       = {Renrui Zhang and
                  Dongzhi Jiang and
                  Yichi Zhang and
                  Haokun Lin and
                  Ziyu Guo and
                  Pengshuo Qiu and
                  Aojun Zhou and
                  Pan Lu and
                  Kai{-}Wei Chang and
                  Yu Qiao and
                  Peng Gao and
                  Hongsheng Li},
  editor       = {Ales Leonardis and
                  Elisa Ricci and
                  Stefan Roth and
                  Olga Russakovsky and
                  Torsten Sattler and
                  G{\"{u}}l Varol},
  title        = {{MATHVERSE:} Does Your Multi-modal {LLM} Truly See the Diagrams in
                  Visual Math Problems?},
  booktitle    = {Computer Vision - {ECCV} 2024 - 18th European Conference, Milan, Italy,
                  September 29-October 4, 2024, Proceedings, Part {VIII}},
  series       = {Lecture Notes in Computer Science},
  volume       = {15066},
  pages        = {169--186},
  publisher    = {Springer},
  year         = {2024},
  url          = {https://doi.org/10.1007/978-3-031-73242-3\_10},
  doi          = {10.1007/978-3-031-73242-3\_10},
  bibsource    = {dblp computer science bibliography, https://dblp.org}
}

@inproceedings{chen2024mmstar,
  author       = {Lin Chen and
                  Jinsong Li and
                  Xiaoyi Dong and
                  Pan Zhang and
                  Yuhang Zang and
                  Zehui Chen and
                  Haodong Duan and
                  Jiaqi Wang and
                  Yu Qiao and
                  Dahua Lin and
                  Feng Zhao},
  editor       = {Amir Globersons and
                  Lester Mackey and
                  Danielle Belgrave and
                  Angela Fan and
                  Ulrich Paquet and
                  Jakub M. Tomczak and
                  Cheng Zhang},
  title        = {Are We on the Right Way for Evaluating Large Vision-Language Models?},
  booktitle    = {Advances in Neural Information Processing Systems 37: Annual Conference
                  on Neural Information Processing Systems 2024, NeurIPS 2024, Vancouver,
                  BC, Canada, December 10 - 15, 2024},
  year         = {2024},
  url          = {http://papers.nips.cc/paper\_files/paper/2024/hash/2f8ee6a3d766b426d2618e555b5aeb39-Abstract-Conference.html},
  bibsource    = {dblp computer science bibliography, https://dblp.org}
}

@inproceedings{guan2024hallusionbench,
  author       = {Tianrui Guan and
                  Fuxiao Liu and
                  Xiyang Wu and
                  Ruiqi Xian and
                  Zongxia Li and
                  Xiaoyu Liu and
                  Xijun Wang and
                  Lichang Chen and
                  Furong Huang and
                  Yaser Yacoob and
                  Dinesh Manocha and
                  Tianyi Zhou},
  title        = {Hallusionbench: An Advanced Diagnostic Suite for Entangled Language
                  Hallucination and Visual Illusion in Large Vision-Language Models},
  booktitle    = {{IEEE/CVF} Conference on Computer Vision and Pattern Recognition,
                  {CVPR} 2024, Seattle, WA, USA, June 16-22, 2024},
  pages        = {14375--14385},
  publisher    = {{IEEE}},
  year         = {2024},
  url          = {https://doi.org/10.1109/CVPR52733.2024.01363},
  doi          = {10.1109/CVPR52733.2024.01363},
  bibsource    = {dblp computer science bibliography, https://dblp.org}
}

@inproceedings{li2023pope,
  author       = {Yifan Li and
                  Yifan Du and
                  Kun Zhou and
                  Jinpeng Wang and
                  Wayne Xin Zhao and
                  Ji{-}Rong Wen},
  editor       = {Houda Bouamor and
                  Juan Pino and
                  Kalika Bali},
  title        = {Evaluating Object Hallucination in Large Vision-Language Models},
  booktitle    = {Proceedings of the 2023 Conference on Empirical Methods in Natural
                  Language Processing, {EMNLP} 2023, Singapore, December 6-10, 2023},
  pages        = {292--305},
  publisher    = {Association for Computational Linguistics},
  year         = {2023},
  url          = {https://doi.org/10.18653/v1/2023.emnlp-main.20},
  doi          = {10.18653/V1/2023.EMNLP-MAIN.20},
  bibsource    = {dblp computer science bibliography, https://dblp.org}
}

@inproceedings{cheng2024seeclick,
  author       = {Kanzhi Cheng and
                  Qiushi Sun and
                  Yougang Chu and
                  Fangzhi Xu and
                  Yantao Li and
                  Jianbing Zhang and
                  Zhiyong Wu},
  editor       = {Lun{-}Wei Ku and
                  Andre Martins and
                  Vivek Srikumar},
  title        = {SeeClick: Harnessing {GUI} Grounding for Advanced Visual {GUI} Agents},
  booktitle    = {Proceedings of the 62nd Annual Meeting of the Association for Computational
                  Linguistics (Volume 1: Long Papers), {ACL} 2024, Bangkok, Thailand,
                  August 11-16, 2024},
  pages        = {9313--9332},
  publisher    = {Association for Computational Linguistics},
  year         = {2024},
  url          = {https://doi.org/10.18653/v1/2024.acl-long.505},
  doi          = {10.18653/V1/2024.ACL-LONG.505},
  bibsource    = {dblp computer science bibliography, https://dblp.org}
}

@inproceedings{wu2025osatlas,
  author       = {Zhiyong Wu and
                  Zhenyu Wu and
                  Fangzhi Xu and
                  Yian Wang and
                  Qiushi Sun and
                  Chengyou Jia and
                  Kanzhi Cheng and
                  Zichen Ding and
                  Liheng Chen and
                  Paul Pu Liang and
                  Yu Qiao},
  title        = {{OS-ATLAS:} Foundation Action Model for Generalist {GUI} Agents},
  booktitle    = {The Thirteenth International Conference on Learning Representations,
                  {ICLR} 2025, Singapore, April 24-28, 2025},
  publisher    = {OpenReview.net},
  year         = {2025},
  url          = {https://openreview.net/forum?id=n9PDaFNi8t},
  bibsource    = {dblp computer science bibliography, https://dblp.org}
}
